\documentclass[preprint,12pt,authoryear]{elsarticle}

\usepackage[noend]{algpseudocode}
\usepackage{algorithmicx,algorithm}

\algnewcommand{\algorithmicforeach}{\textbf{for each}}
\algdef{SE}[FOR]{ForEach}{EndForEach}[1]
  {\algorithmicforeach\ #1\ \algorithmicdo}
  {\algorithmicend\ \algorithmicforeach}

\usepackage[final]{changes}

\usepackage{amssymb}
\usepackage{makecell}
\usepackage{booktabs}
\usepackage{graphicx}
\usepackage{multirow}
\usepackage{float}
\usepackage{subfig}
\usepackage{color}
\usepackage{amsmath,bm}
\usepackage[export]{adjustbox}

\usepackage{lineno}

\usepackage{hyperref}

\usepackage{color, soul}
\setulcolor{red}
\setstcolor{blue}
\sethlcolor{yellow}

\begin{document}

\begin{frontmatter}



\title{Robust structure from motion for aerial-ground images via detector-free feature matching and multi-view track refinement}


\author[label1,label2,label3,label4]{San Jiang\corref{cor1}}
\author[label1]{Hui Wang}
\author[label1,label2,label3,label4]{Xing Zhang}
\author[label1,label2,label3,label4]{Zhongwen Hu}
\author[label5]{Zhijun Wang}
\author[label1]{Ruisheng Wang}
\author[label6]{Wanshou Jiang}
\author[label1,label2,label3,label4]{Qingquan Li}

\affiliation[label1]{organization={School of Architecture and Urban Planning, Shenzhen University},
            city={Guangdong Shenzhen},
            postcode={518060}, 
            country={China}}
            
\affiliation[label2]{organization={Guangdong Key Laboratory of Urban Informatics, Shenzhen University},
            city={Guangdong Shenzhen},
            postcode={518060}, 
            country={China}}

\affiliation[label3]{organization={MNR Key Laboratory for Geo-Environmental Monitoring of Great Bay Area, Shenzhen University},
            city={Shenzhen},
            postcode={518060}, 
            country={China}}
            
\affiliation[label4]{organization={Shenzhen Key Laboratory of Spatial Smart Sensing and Services},
            city={Shenzhen},
            postcode={518060}, 
            country={China}}

\affiliation[label5]{organization={Guangdong Laboratory of Artificial Intelligence and Digital Economy (Shenzhen)},
            city={Shenzhen},
            postcode={518060}, 
            country={China}}

\affiliation[label6]{organization={State Key Laboratory of Information Engineering in Surveying, Mapping and Remote Sensing, Wuhan University},
            city={Wuhan},
            postcode={430072}, 
            country={China}}

\cortext[cor1]{Corresponding Author: jiangsan@szu.edu.cn}

\begin{abstract}
Integrated 3D reconstruction from aerial-ground images is essential for generating high-precision urban 3D models, yet severe variations in viewpoint, scale, and rotation make robust feature matching highly challenging. To address these limitations, this study introduces a rotation-robust detector-free matching network coupled with multi-view track refinement for incremental Structure from Motion (ISfM). The proposed workflow features four key modules. First, rotation-aware feature extraction replaces traditional convolutions with an Omnidirectional State Space Block (OSS Block) that selectively scans across eight symmetrical directions to model long-range spatial dependencies and synthesize rotation-invariant feature maps. Second, multi-scale attention transformation utilizes quadtree attention to build a hierarchical token pyramid that isolates high-association token regions and discards irrelevant areas, capturing long-range context with linear computational complexity. Third, bi-directional feature matching executes a symmetric coarse-to-fine matching scheme where coarse alignment computes dual-direction Softmax confidence matrices under mutual nearest neighbor constraints, and fine alignment uses a multi-layer perceptron to regress sub-pixel coordinate offsets. Finally, multi-view track refinement employs an integrated indexing structure to evaluate localized spatial proximity and link disjoint sub-tracks to the highest-confidence anchor point, ensuring stable feature repeatability across the ISfM pipeline. By using real aerial-ground datasets, experimental results demonstrate that the proposed method improves AUC at 5$^{\circ}$ pose error by 93.9\% compared with LoFTR and achieves the highest precision in ISfM reconstruction, with the improved accuracy ranging from 27.6\% to 32.7\%. The proposed method provides a reliable solution for integrated 3D reconstruction of aerial-ground images.
\end{abstract}



\begin{keyword}
detector-free feature matching \sep structure from motion \sep deep learning \sep aerial-ground images \sep 3D reconstruction


\end{keyword}

\end{frontmatter}


\section{Introduction}
\label{sec1}

3D reconstruction of complex urban scenes is extensively explored within the domain of photogrammetry and computer vision \citep{ge2023rapid}. Over the past decades, UAV (Unmanned Aerial Vehicle) oblique photogrammetry has emerged as a standard solution for urban 3D reconstruction, which enables the flexible configuration of data acquisition trajectories and the multi-view image capture of building facades \citep{jiang2022unmanned,xiang2019mini}. However, constrained by limited flight altitudes and sensor viewing angles, 3D models derived solely from UAV aerial platforms often struggle to achieve satisfactory accuracy and completeness in dense urban environments because of intricate architectures and severe occlusions. To resolve this issue, ground mobile mapping systems (MMS) have been increasingly cooperating with low-altitude UAV platforms for aerial-ground image acquisitions \citep{li2023fusion}. By capturing high-resolution, ground-level images of building facades, MMS effectively compensates for the observation perspectives of UAV platforms and replenishes missing details at the bottoms of buildings. Consequently, the integration of aerial and ground images has become the definitive framework for achieving high-fidelity, comprehensive urban 3D reconstructions \citep{zhou2024unified, zhu2020leveraging}.

Because of its independence on precise initial values of unknowns and resilience against noise, incremental Structure from Motion (ISfM) has become the golden standard technique for camera orientation and scene reconstruction \citep{jiang2020efficient, xu2021robust}. The pipeline of ISfM typically consists of two main stages, i.e., feature matching and incremental reconstruction. The former aims to establish accurate correspondences between images for two-view geometry estimation; the latter attempts to recover camera poses and 3D points via iterative BA (bundle adjustment) optimization. Owing to severe discrepancies in acquisition perspectives, aerial and ground images exhibit massive variations, such as in viewpoints, scales, and illuminations. Traditional hand-crafted feature methods fail to establish effective feature matches for aerial-ground images, which seriously degenerate direct image matching and lead to the failure of joint aerial-ground ISfM reconstruction. Thus, the integration of aerial and ground images mainly depends on their successful registration.

In the literature, existing studies can be classified into two categories \citep{gao2018a}. The first one leverages a two-step procedure that generates sparse models from aerial and ground images separately and registers them into a unified coordinate system by using POS (Positioning and Orientation System) data as the geometric clues. Therefore, the cross-view image integration task can be reframed as a point cloud registration problem between the two sparse models \citep{gao2018b, shan2014accurate}. This alignment is typically executed using robust geometric transformation estimators. Nevertheless, the registration accuracy often degrades severely, primarily because of the sparse models' restricted overlapping regions and the presence of structural outliers. On the contrary, the second one aims to establish feature matches between aerial and ground images, which are used as tie-points for the BA (bundle adjustment) optimization in ISfM. The perspective distortions and scale differences are the main issues that cause the failure of feature matching between aerial and ground images. To cope with these issues, creative solutions have been reported, e.g., image rectification-based methods \citep{li2023fusion, liu2023tie, wu2018integration} and view rendering-based methods \citep{yu2026aerial, zhu2020leveraging}. However, their performance either relies on auxiliary data and scene structures or consumes high computational costs.

In recent years, deep learning-based feature matching networks have demonstrated remarkable potential in complex scenes and outperformed traditional handcrafted methods on public challenging datasets \citep{jiang2021learned}. According to the usage of feature detectors, existing networks are mainly divided into two groups, i.e., detector-based and detector-free methods. Detector-based methods first extract local features and then match them across images, e.g., SuperPoint \citep{detone2018superpoint} and ASLFeat \citep{luo2020aslfeat}. Detector-based methods show good compatibility with the downstream ISfM workflow. However, they often fail to extract effective features in weak and repetitive texture regions and have low repeatability in feature detection between aerial and ground images. In contrast, detector-free methods directly form dense pixel-level correspondences on feature maps without explicit keypoint detection, which deliver an order of magnitude more matches and show superior performance in texture-poor and cross-view scenes \citep{wang2025aerial}. The typical methods are LoFTR \citep{sun2021loftr} and its subsequent optimized versions \citep{chen2022aspanformer, tang2026robust}. Nevertheless, detector-free methods still face two critical bottlenecks when applied to ISfM reconstruction. First, they lack sufficient rotational robustness, which is critical for processing aerial and ground images. Existing detector-free networks use CNN-based feature extraction modules, whose square convolution kernels and translation-invariant operations cannot effectively capture rotated features. For these issues, existing solutions include image rotation preprocessing \citep{morelli2024deep}, improved convolution kernels \citep{lee2023learning}, and rotation-augmented training \citep{parihar2021rord}. However, they only handle fixed-angle rotations and filter out edge features, as well as incur high computational costs. Second, detector-free methods are poorly compatible with ISfM. The main reason arises from the fact that they generate matches independently for each image pair, leading to low repeatability of feature points across views. This causes fragmented feature tracks among multi-views and disconnected sub-scenes in ISfM reconstruction. Although the widely used grid quantization strategy alleviates this issue, it significantly sacrifices the accuracy of feature matching.

To address the these issues, this study proposes a detector-free feature matching network for aerial and ground images and a multi-view track refinement strategy tailored for ISfM. The main contribution of this study can be summarized as follows: (1) we propose a detector-free feature matching network that integrates an omnidirectional state space block, multi-scale attention, and bidirectional fine matching to address the issues of rotation-sensitive feature extraction and low-accuracy feature matching; (2) we propose a multi-view track refinement strategy that uses an integrated indexing structure to evaluate localized spatial proximity and connect disjoint sub-tracks to the highest-confidence anchor point, which ensures feature repeatability for ISfM reconstruction; and finally (3) we verify the performance of the proposed workflow using aerial-ground images in terms of feature matching and ISfM Reconstruction and compare it with state-of-the-art methods.

This study is organized as follows. Section \ref{sec2} presents the related work in the literature. Section \ref{sec3} gives the details of the proposed solution for aerial-ground image feature matching, which is followed by experimental tests in terms of feature matching and ISfM reconstruction in Section \ref{sec4}. Finally, Section \ref{sec5} presents conclusions and future studies.

\section{Related work}
\label{sec2}

In the literature, existing deep learning-based feature matching networks are mainly grouped into two categories, i.e., detector-based methods and detector-free methods. Thus, this section conducts a review from these two aspects as well as their integration with the ISfM reconstruction workflow.

\subsection{Detector-based feature matching}
\label{sec2.1}

Detector-based feature matching is a classic and mature technical route in the literature. Its workflow follows the detect-then-match mode, which detects repeatable features from each image and then searches for matches from two sets of features. The typical methods consist of the earlier hand-crafted algorithms, e.g., SIFT (Scale Invariant Feature Transform) \citep{lowe2004distinctive}, and the recently learned networks, e.g., L2-Net \citep{tian2017l2} and Geodesc \citep{luo2018geodesc}. These methods have been well studied and used in the classical ISfM workflow \citep{schonberger2016structure}.

Existing research primarily focuses on enhancing image matching performance from two perspectives, i.e., feature extraction and feature matching. For feature extraction optimization, \cite{luo2019contextdesc} proposed the ContextDesc network that explores the powerful representation learning capabilities of CNN (convolutional neural network) to extract deep semantic features from image patches, thereby enriching the discriminative power of descriptors. To overcome the limitations of traditional detect-then-describe frameworks, where variations in factors such as illumination and scale lead to incomplete feature repeatability, \cite{dusmanu2019d2} introduced an end-to-end training mechanism that jointly detects and describes features, designing D2-Net to improve environmental adaptability and feature consistency. For feature matching improvement, \cite{sarlin2020superglue} used GNN (graph neural network) and attention mechanisms to dynamically model global relationships among features, thereby increasing the distinctiveness of feature descriptors. Subsequently, employing the Sinkhorn algorithm, they reframed the feature matching task as an optimal transport problem and developed the SuperGlue network that significantly improves matching accuracy. Inspired by SuperGlue, there are many other revised versions that focus on performance improvement from varying aspects, such as LightGlue for matching efficiency \citep{lindenberger2023lightglue}.

The prominent characteristic of detect-then-match methods is that keypoints extracted from a single image can be repeatedly utilized across all overlapped images, which is highly advantageous for forming multi-view tracks. Consequently, these methods adapt seamlessly to the ISfM workflow. However, the core of this matching workflow relies heavily on the performance of feature detectors. If the density of extracted features is insufficient or the distinctiveness of descriptors is very limited, the downstream matching process is severely bottlenecked, making it exceedingly difficult to yield valid feature matches. As a result, these methods struggle to perform effectively in textureless scenes as well as for images with large perspective deformations and scale differences. This is also the main reason to revise feature matching algorithms for aerial-ground images, such as image geometric rectification \citep{li2023fusion, wu2018integration} and intermediate view rendering \citep{yu2026aerial, zhu2020leveraging}.

\subsection{Detector-free feature matching}
\label{sec2.2}
In contrast to detector-based methods, detector-free feature matching methods abandon the traditional strategy of extracting local features from single images via detectors. Instead, they directly process image pairs in an end-to-end manner. These methods can make full use of all pixel information and mine rich feature correspondences between image pairs, which has shown great potential in high-difficulty matching tasks \citep{xu2024local}.

In recent years, with the development of the Transformer, detector-free feature matching methods have developed rapidly. LoFTR is the most representative method \citep{sun2021loftr}, and its core contribution is to use a Transformer to realize dynamic feature update and adopt a coarse-to-fine matching strategy. By leveraging a feature pyramid network (FPN), LoFTR first extracts coarse and fine-scale feature maps from image pairs. The coarse features are then fused with positional encodings and updated via an attention-based local transformation module to establish initial match pairs through bidirectional normalization and mutual nearest neighbor (MNN) criteria. Finally, local windows are cropped from the fine-grained feature maps around these coarse seeds to optimize the final correspondences to sub-pixel accuracy based on feature correlation. Motivated by LoFTR, detector-free feature matching has gained lots of attention in the literature. \cite{chen2022aspanformer} designed a global-local attention block to adjust attention span adaptively and proposed the AspanFormer network, addressing the issue of lacking local interaction in LoFTR. Mimicking human interaction behavior, \cite{wang2022matchformer} proposed the MatchFormer network that uses a self-attention block for non-local feature detection and a cross-attention block for their similarity induction. These networks have also been applied to aerial-ground image matching \citep{zhu2024vdft}.

Detector-free feature matching methods exhibit two prominent advantages over their detector-based counterparts. First, by bypassing the feature detection stage, they effectively mine dense semantic information in textureless scenes and for perspective-distorted images. Second, their end-to-end processing mode streamlines the matching workflow, mitigating the error propagation typically caused by decoupling feature extraction from matching. Despite these strengths, detector-free feature matching possesses an inherent limitation. Because they process image pairs independently, the generated match points exhibit poor repeatability across different image pairs. This lack of consistency fails to guarantee a reliable multi-view track, making it difficult to adapt to ISfM reconstruction. Thus, recent studies attempt to improve the matching accuracy \citep{edstedt2024roma, zhou2024unified}.

\subsection{Detector-free SfM refinement}
\label{sec2.3}
The localization precision of feature matching significantly affects the precision of ISfM reconstruction, which is verified in both detector-based and detector-free ISfM \citep{he2024detector, lindenberger2021pixel}. Especially, detector-free matching methods have poor repeatability of matching points between different image pairs, which would lead to fragmented feature tracks in track generation. To enhance the adaptability of detector-free matching to ISfM, researchers have proposed detector-free SfM refinement methods, and Detector-free SfM \citep{he2024detector} is the most representative solution. It adopts a coarse-to-fine matching method, which solves the consistency problem of detector-free feature matching in an iterative manner. In the coarse SfM stage, the positions of features are rounded to coarsely gridded values to achieve quantitative matching, and coarse SfM reconstruction is performed to obtain initial camera poses and 3D scenes. In the iterative refinement stage, feature track refinement and geometric refinement in Detector-free SfM are alternately performed to obtain accurate camera poses and dense 3D point clouds. Based on the framework of Detector-free SfM, \cite{lee2025dense} propose the Dense-SfM workflow that achieves track extension via realistic scene rendering and 3D point visibility analysis by incorporating the rendering ability of 3D Gaussian Splatting. Dense-SfM aims to address the issue of consistent track generation in the detector-free feature matching methods.

Although detector-free SfM refinement can mitigate fragmented feature tracks and enhance compatibility in the ISfM workflow, its coarse grid quantization strategy introduces significant localization errors that compromise overall accuracy. Furthermore, its iterative refinement process fails to account for the severe scale and rotation discrepancies inherent to aerial-ground images, ultimately limiting the precision of the final reconstruction results.

\section{Methodology}
\label{sec3}

Detector-free feature matching methods demonstrate strong potential for aerial-ground images. However, existing methods suffer from low matching point localization accuracy and insufficient cross-view feature repeatability, which fail to guarantee consistent multi-view track generation and make these frameworks incompatible with existing ISfM workflows. Thus, this study proposes a rotation-robust detector-free feature matching network combined with a confidence-driven detector-free ISfM framework for high-precision aerial-ground image matching and 3D reconstruction.

\subsection{Rotation-robust and high-precision feature matching network}
\label{sec3.1}

Detector-free methods offer powerful feature matching ability in complex environments. However, they still face significant bottlenecks when directly applied to aerial-ground images. On the one hand, the severe discrepancies in scale, viewpoint, and perspective drastically worsen the repeatability of match points; on the other hand, the most prevalent detector-free networks depend on CNN-based feature extraction backbones. Since standard CNNs exhibit poor rotation robustness, they often struggle to accommodate the pronounced orientation variations. To overcome these challenges, this study designs a detector-free feature matching network that integrates omnidirectional state space blocks (OSS block), multi-scale attention, and bidirectional fine matching to enhance rotation robustness and matching precision.

\begin{figure}[!htp]
    \centering
    \includegraphics[width=0.9\linewidth]{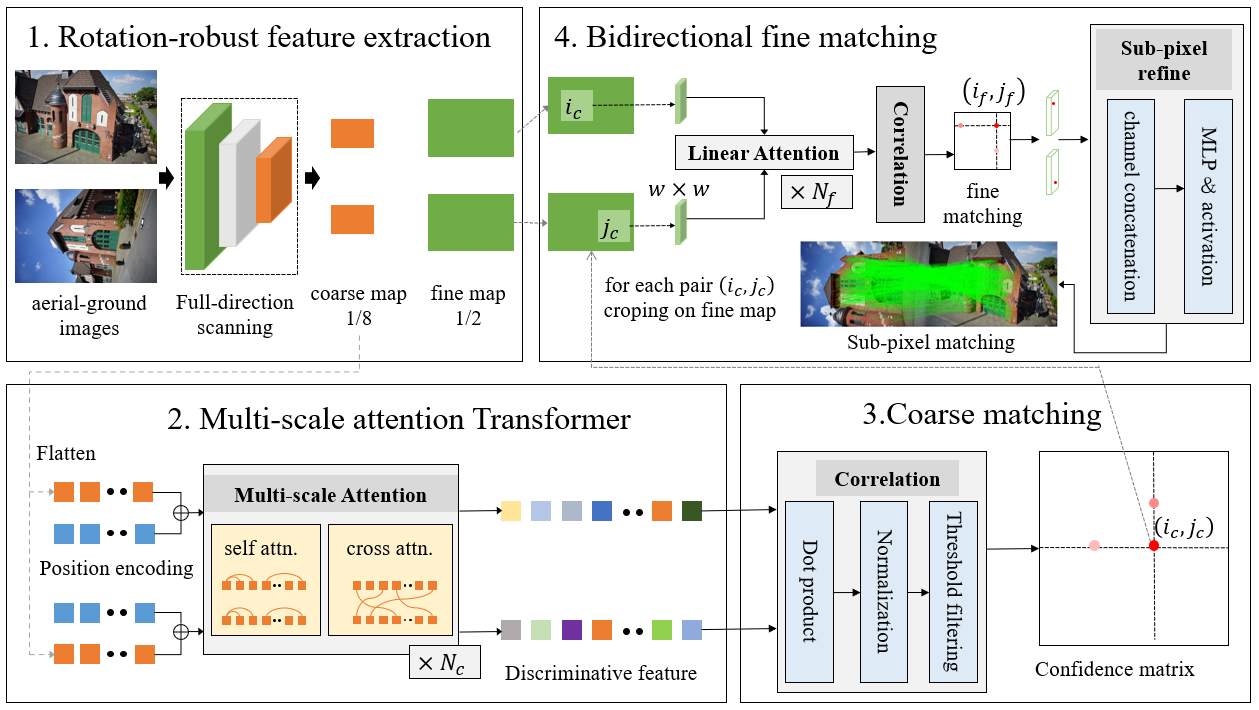}
    \caption{The workflow of the proposed detector-free feature matching network.}
    \label{fig1}
 \end{figure}

Figure \ref{fig1} illustrates the overall architecture of the proposed rotation-robust detector-free feature matching network, which consists of four core modules. First, the rotation-robust feature extraction module uses an omnidirectional scanning mechanism to obtain multi-scale, rotation-invariant feature representations from aerial-ground images. Second, the multi-scale attention transform module applies a hierarchical quadtree attention mechanism to the coarse feature maps, which enables efficient long-range contextual interactions and produces highly discriminative feature sequences. Third, the coarse matching module then calculates scaled dot-product correlations and applies symmetric normalization constraints to establish reliable initial correspondences. Finally, after cropping localized windows from the high-resolution feature map, the bidirectional fine matching module processes them via bidirectional self- and cross-attention, and leverages an offset regression network to refine the initial match points to sub-pixel accuracy.

\subsubsection{Rotation-robust feature extraction}
\label{sec3.1.2}

For aerial-ground images, this module aims to extract the coarse and fine feature maps that would be used in subsequent feature transformation and matching. However, the significant orientation discrepancies between aerial-ground images present a fundamental challenge for CNN-based detector-free matching networks. Since standard CNNs rely on isotropic square kernels and translation-invariant operations, applying identical filters to rotated structures yields inconsistent feature responses that would introduce geometric distortions and degrade matching accuracy. While existing methods attempt to mitigate this by rotating images by 90$^\circ$, 180$^\circ$, or 270$^\circ$ during preprocessing \citep{morelli2024deep}, these discrete adjustments fail to resolve arbitrarily small-angle rotations. Furthermore, because conventional backbones rely on fixed-direction localized convolutions or one-dimensional sequence modeling, they cannot extract globally consistent features under varied orientations, drastically reducing feature robustness across aerial-ground images.

To address the challenge of extracting rotation-invariant features, the Rotation-Robust Feature Extraction (RRFE) module replaces traditional convolutional networks with an OSS block to generate stable, invariant feature maps for aerial-ground image matching. The core component, i.e., the Omnidirectional Selective Scan Module (OSSM), combines the linear computational efficiency of State Space Models (SSMs) with a multi-angle scanning paradigm. To accommodate the multi-directional distribution of spatial features in aerial-ground images, the OSSM executes selective scanning across eight symmetrical directions, i.e., horizontal, vertical, diagonal, anti-diagonal, and their respective inverses. Fusing these directional feature streams enables the network to capture structural dependencies under fine-grained rotational variations, enforcing strict rotation robustness and providing high-quality feature maps for subsequent coarse and fine matching.

\begin{figure}[!t]
    \centering
    \includegraphics[width=0.8\linewidth]{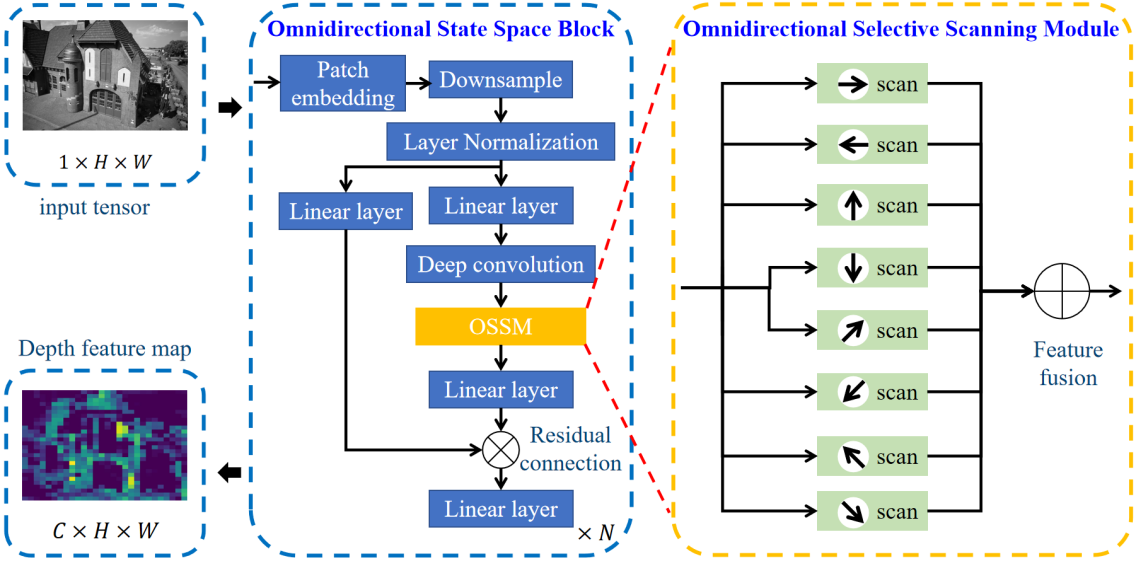}
    \caption{The network principle of the rotational robust feature extraction module.}
    \label{fig2}
 \end{figure}

As shown in Figure \ref{fig2}, the RRFE module proceeds in five steps to extract discriminative, rotation-robust multi-scale feature maps that are used to guide subsequent coarse and fine matching.

\begin{enumerate}[(1)]
    \item Feature initialization transforms the input two-dimensional image into a one-dimensional feature sequence via patch embedding. 
    \item Local feature enhancement downsamples this sequence to expand the receptive field, applies layer normalization to standardize the data distribution, and utilizes depthwise convolution to capture non-linear local representations with minimal parameter overhead. 
    \item Omnidirectional global feature modeling reorganizes the enhanced sequences into eight symmetric scanning paths, processes them through an SSM unit to capture multi-angle dependencies, concatenates the resulting paths, and finally compresses the tensor back to its original shape via a lightweight linear layer.
    \item Residual connection fusion integrates these omnidirectional global contexts with the initial local features in linear time complexity, substantially bolstering feature distinctiveness. 
    \item Multi-scale feature stacking and output chains multiple OSS Block operations alongside progressive downsampling to simultaneously export coarse-scale feature maps $\tilde{F}^{\text{A}}$, $\tilde{F}^{\text{B}}$ at 1/8 resolution for global similarity matching and fine-scale feature maps $\hat{F}^{\text{A}}$, $\hat{F}^{\text{B}}$ at 1/2 resolution to guarantee sub-pixel localization accuracy during fine matching.
\end{enumerate}

\subsubsection{Multi-scale attention Transformer}
\label{sec3.1.2}

The essence of the detector-free matching is to convert the problem of feature matching into semantic association modeling, which originates from the Transformer-based attention mechanism. However, the traditional attention mechanism incurs a quadratic computational complexity that is prohibitive for high-resolution aerial-ground images. To improve efficiency, existing methods often adopt linear attention approximations \citep{yang2023gated}; yet, this mathematical simplification introduces information loss and weakens the network's capacity to capture long-range dependencies. This limitation is severely compounded in aerial-ground images, where drastic variations in observation distances yield striking feature discrepancies. Thus, the information loss inherent to linear attention degrades the accuracy of multi-scale semantic modeling, exacerbating the overall deficiency in feature association.

\begin{figure}[!t]
    \centering
    \includegraphics[width=0.8\linewidth]{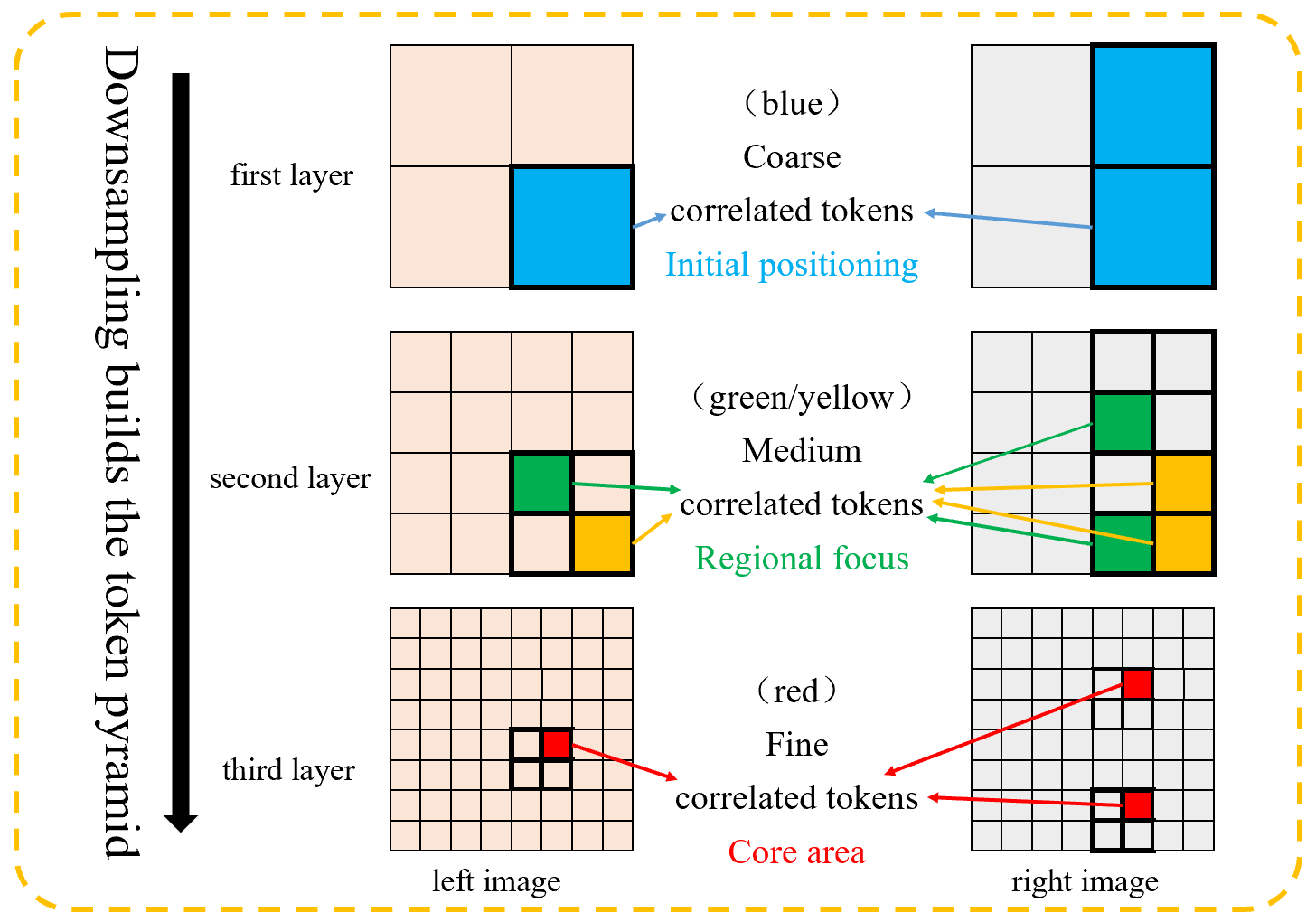}
    \caption{The network principle of the multi-scale attention Transformer module.}
    \label{fig3}
 \end{figure}

To solve the problem of insufficient correlation modeling in linear attention, this study designs a multi-scale attention Transformer module that aims to achieve global interaction and enhancement between local features and improve the discriminative power of feature maps. This study replaces the linear attention mechanism with quadtree attention \citep{tang2022quadtree} via the combination of a multi-scale token pyramid, hierarchical attention calculation, and weight information fusion. While achieving attention calculation with linear complexity, it can capture long-range dependent information, retain fine-grained features, and achieve scale-related collaborative optimization. Consequently, it can provide discriminative feature maps for subsequent aerial-ground image matching.

The core idea of quadtree attention is hierarchical attention calculation from coarse to fine and a weighted combination of information at different scales, as shown in Figure \ref{fig3}. It specifically includes three aspects. 

\begin{enumerate}[(1)]
    \item Taking as input the coarse-scale feature maps $\tilde{F}^{\text{A}}$, $\tilde{F}^{\text{B}}$ from the feature extraction module, hierarchical downsampling is performed to form a token pyramid at different scales, realizing hierarchical expression of multi-scale features for aerial-ground images. 
    \item At each scale level, the attention coefficient between it and the key feature token is calculated for each token of the query feature, and the top K tokens with the highest attention coefficients are selected as candidate regions. Attention calculation for the next scale is only performed within the candidate regions, skipping irrelevant regions to reduce information loss and accurately represent the correlation between features. 
    \item The attention information of different layers is fused using a weighted sum method, which not only avoids the excessive impact of coarse-scale features on the results of fine-scale features but also does not lose effective information in each feature layer. Besides, multi-scale locally enhanced positional encoding is added to ensure that attention calculation considers scale correlation, spatial position, and long-range dependence, and it finally outputs the enhanced feature sequences $\tilde{F}_{tr}^{A}(i)$ and $\tilde{F}_{tr}^{B}(i)$.
\end{enumerate}

This module enhances the feature correlation between multiple scales and highlights fine-grained details while retaining global context information, ultimately obtaining discriminative feature maps for subsequent matching. Considering the large-scale difference of aerial-ground images, the pyramid level of this module is set to 3 in this study, and the $K$ values for the coarse, medium, and fine scales are set to 32, 16, and 8, respectively, which effectively balances computational efficiency and feature representation capability.

\subsubsection{Coarse matching}
\label{sec3.1.3}
The purpose of the coarse matching module is to search for initial matches based on the enhanced feature maps by the multi-scale attention module, and select high-quality coarse matches as seed candidates for subsequent precise matching. The coarse matching module in this study is optimized based on the strategy proposed in \citep{sun2021loftr}. By adding a bias-based linear projection layer to the initial feature maps, learnable training adaptation is performed on the coarse features to enhance their discriminative ability, which is used for subsequent similarity calculation and matching decision-making, thereby more effectively selecting distinguishable features. The coarse matching consists of three steps.

(1) Dot product calculation. The coarse feature sequences $\tilde{F}_{tr}^{A}(i)$ and $\tilde{F}_{tr}^{B}(i)$ are linearly projected, and the similarity matrix $S_c$ of the transformed features is calculated using scaled dot product, as shown in Equation \ref{equ:1}. The temperature coefficient $\tau$ is set to 0.1 to control the dynamic range of similarity scores. Each element $S_c(i,j)$ of the similarity matrix indicates the similarity score between the $i$-th token in image A and the $j$-th token in image B.

\begin{equation}
    S_c(i, j) = \frac{1}{\tau} \cdot \langle \tilde{F}_{tr}^{A}(i), \tilde{F}_{tr}^{B}(j) \rangle
    \label{equ:1}
\end{equation}

(2) Normalization. To ensure that similar matches are symmetrical in both directions, this module converts the similarity matrix $S_c$ into a confidence matrix $P_c$ via bidirectional softmax operations, as represented in Equation \ref{equ:2}. This operation performs bidirectional normalization on the rows and columns of the similarity matrix $S_c$ simultaneously. This bidirectional relationship constraint ensures the correctness of coarse matching.

\begin{equation}
    P_c(i, j) = \text{softmax}(S_c(i, \cdot))_j \cdot \text{softmax}(S_c(\cdot, j))_i
    \label{equ:2}
\end{equation}

(3) Screening of coarse matches. All matches with confidence $P_c(i, j) \geq \theta_c$ are selected as the candidate of coarse matches, where $\theta_c=0.2$. Then, the mutual nearest neighbor criterion is applied to the candidate coarse matches to obtain the coarse matching set $M_c$, as shown in Equation \ref{equ:3}, which satisfies $j=\textit{argmax}_j P_c (i,j)$ and $i=\textit{argmax}_i P_c (i,j)$.

\begin{equation}
    M_c = \{ (\tilde{i}, \tilde{j}) \mid P_c(\tilde{i}, \tilde{j}) \geq \theta_c \cap \text{MNN}(\tilde{i}, \tilde{j}) \}
    \label{equ:3}
\end{equation}

\subsubsection{Bidirectional fine matching}
\label{sec3.1.4}

\begin{figure}[!t]
    \centering
    \includegraphics[width=0.8\linewidth]{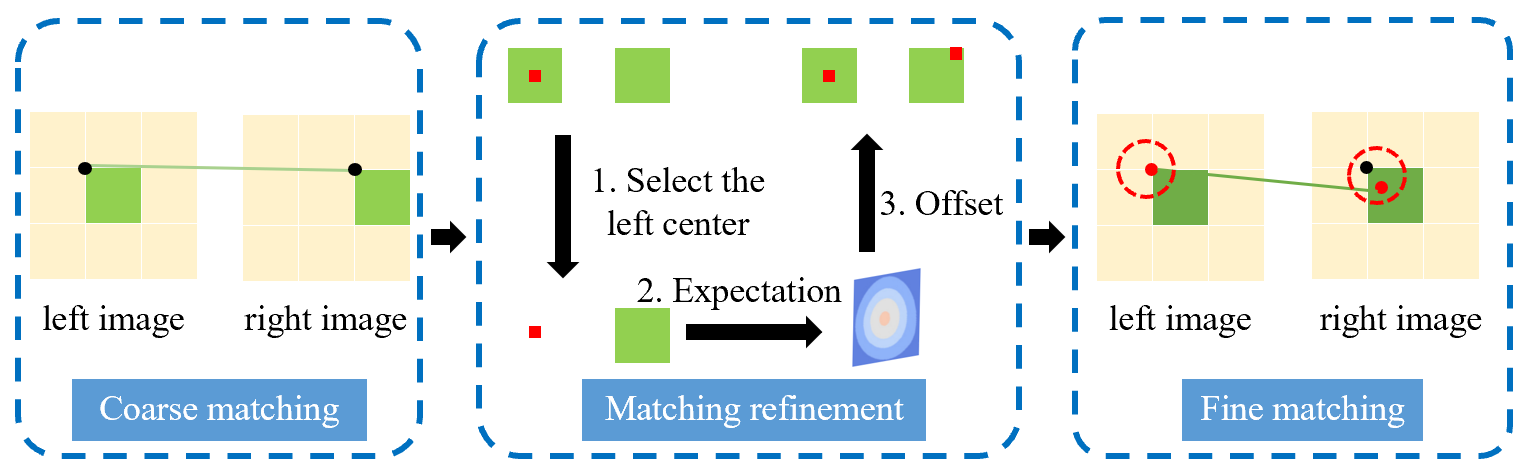}
    \caption{The main issue of one-way feature matching strategy in the refinement stage.}
    \label{fig4}
 \end{figure}

\begin{figure}[!t]
    \centering
    \includegraphics[width=0.8\linewidth]{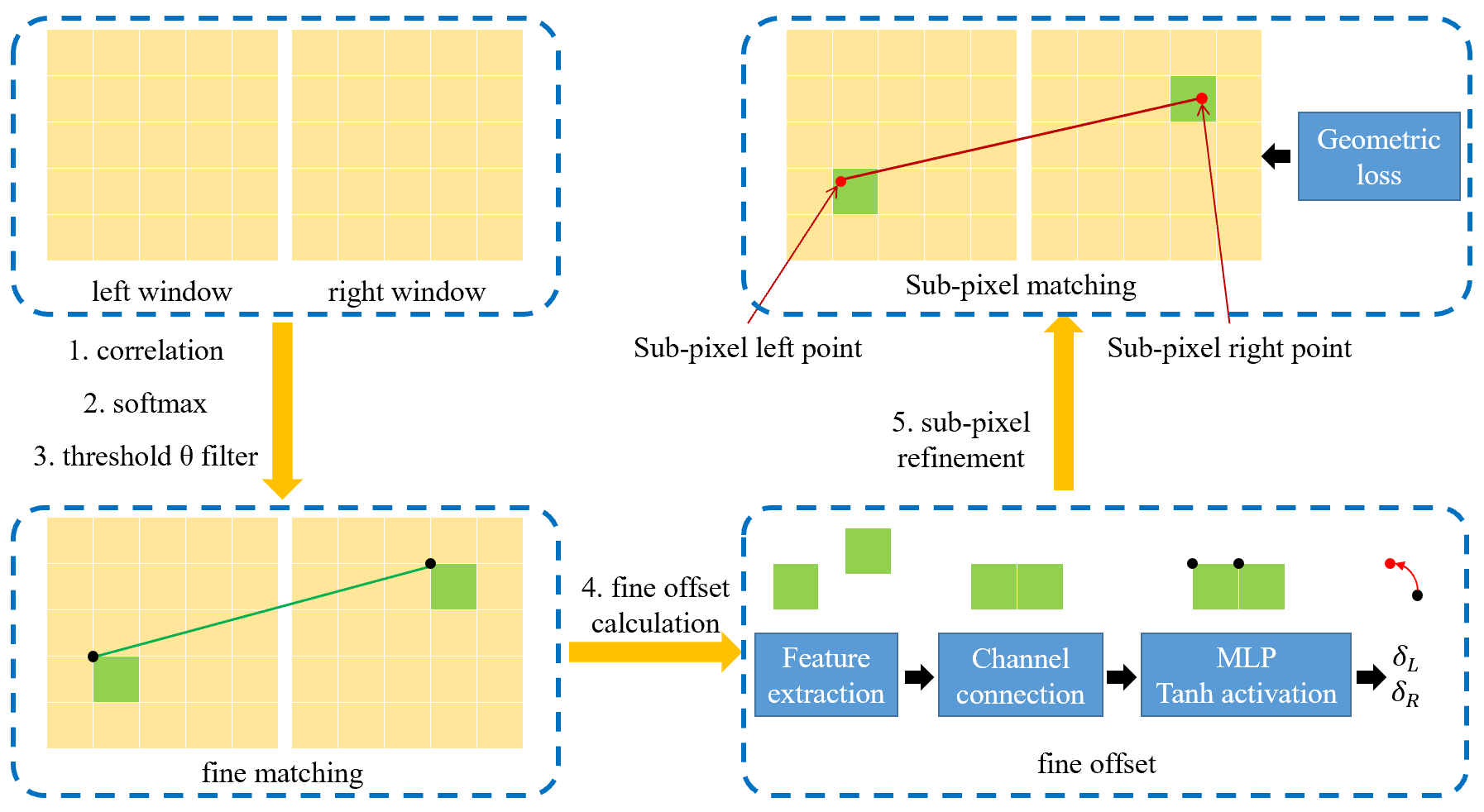}
    \caption{The principle of bidirectional precision matching module.}
    \label{fig5}
 \end{figure}

As shown in Figure \ref{fig4}, traditional coarse-to-fine matching suffers from low accuracy due to unidirectional dependency. Fine matching depends on fixed reference points from coarse matching, introducing ambiguity and localization error. Inspired by \cite{tuzcuoglu2024xoftr}, this study uses a bidirectional fine matching module for sub-pixel refinement, which directly optimizes correspondences symmetrically and improves feature point localization precision. The core function of this module is to refine the coarse matching results to sub-pixel accuracy, enhance the positioning accuracy of matching points in aerial-ground images, and avoid the problem of one-way optimization ambiguity in existing methods. As illustrated in Figure \ref{fig5}, the bidirectional precise matching module consists of four major steps.

(1) Local window cropping. For each coarse matching pair $(\tilde{i}, \tilde{j})$ obtained by the coarse matching module, a sub-pixel-sized fine local window centered on the matching token is cropped from the fine-scale feature maps $\widehat{F}^{\text{A}}, \widehat{F}^{\text{B}}$ that comes from the rotation-robust feature extraction module, focusing on the fine features of the matching points.

(2) Bidirectional attention feature interaction. The cropped fine local windows are all enhanced using a self-attention layer to improve the discriminative ability of features, and then a cross-attention layer is utilized for bidirectional attention feature interaction to fully explore the corresponding relationship between discriminative features, obtaining the enhanced fine discriminative features $\widehat{F}_{tr}^{\text{A}}(i)$ and $\widehat{F}_{tr}^{\text{B}}(j)$.

(3) Bidirectional similarity calculation and precise matching screening. Similar to coarse matching, the bidirectional similarity matrix $S_f(i, j)$ of the enhanced fine features $\widehat{F}_{tr}^{\text{A}}(i)$ and $\widehat{F}_{tr}^{\text{B}}(j)$ is calculated using the scaled dot product method with the temperature coefficient $\tau = 0.1$. Softmax operations are then performed on the rows and columns of the similarity matrix $S_f(i, j)$ to obtain the precise matching confidence matrix $P_f$. With the threshold $\theta_f=0.1$, candidate precise matches with $P_f(i,j) \geq\theta_f$ are selected, and the MNN condition is applied to optimize the bidirectional optimal precise matches, obtaining the precise matching set $M_f$ and the precise matching coordinates $(\widehat{x}_{\text{A}}, \widehat{y}_{\text{A}})$, $(\widehat{x}_{\text{B}}, \widehat{y}_{\text{B}})$.

(4) Sub-pixel bidirectional refinement. The feature vectors $\widehat{F}^{\text{A}}, \widehat{F}^{\text{B}}$ corresponding to the precise match $(\tilde{i}, \tilde{j})$ are concatenated, and the bidirectional sub-pixel offsets $(\widehat{x}_{\text{A}}, \widehat{y}_{\text{A}})$ and $(\widehat{x}_{\text{B}}, \widehat{y}_{\text{B}})$ are regressed using a MLP and the Tanh function. Then, the sub-pixel offsets are added to the coordinates of the precise match to obtain the sub-pixel match $M_{sub} = {((x_A^{sub}, y_A^{sub}), (x_B^{sub}, y_B^{sub}), \text{P}_f)}$ as calculated according to Equations \ref{equ:4}-\ref{equ:5}. According to the above steps, the relevant area is repositioned within the small local window, which significantly reduces the positioning error of matching points.

\begin{equation}
    \{\delta_{xA}, \delta_{yA}, \delta_{xB}, \delta_{yB}\} = \text{Tanh}\left( \text{MLP}\left( \tilde{f}^A(\hat{i}) \parallel \tilde{f}^B(\hat{j}) \right) \right)
    \label{equ:4}
\end{equation}

\begin{equation}
    \begin{cases} 
    (x_A^{sub}, y_A^{sub}) = (\hat{x}_A + \delta_{xA}, \hat{y}_A + \delta_{yA}) \\ 
    (x_B^{sub}, y_B^{sub}) = (\hat{x}_B + \delta_{xB}, \hat{y}_B + \delta_{yB}) 
    \end{cases}
    \label{equ:5}
\end{equation}

\subsubsection{Loss function}
\label{sec3.1.5}
Our final loss function $L$ consists of three components, i.e., coarse-level matching loss $L_c$, fine-level matching loss $L_f$, and sub-pixel refinement loss $L_{sub}$. For the coarse-level matching loss $L_c$ follows LoFTR, which uses camera poses and depth maps to compute the ground-truth labels for the confidence matrix $P_c$ during training. We define the ground-truth coarse matches $M_c^{gt}$ as the mutual nearest neighbors of the two sets of 1/8-resolution grids and minimize the negative log-likelihood loss over the grids in $M_c^{gt}$. The coarse-level loss is defined as Equation \ref{equ:6}.

\begin{equation}
    L_c = -\frac{1}{|M_c^{gt}|} \sum_{(\tilde{i}, \tilde{j}) \in M_c^{gt}} \log P_c(\tilde{i}, \tilde{j})
    \label{equ:6}
\end{equation}

For the fine-level matching loss, we select only one point as a match from fine-level windows and apply the Focal Loss (FL) to supervise all fine-level features correspondences in $P_f (\tilde{i}, \tilde{j})$. The fine-level loss is then defined according to Equation \ref{equ:7}.

\begin{equation}
    L_f = \frac{1}{|M_c^{gt}|} \sum_{(\tilde{i}, \tilde{j}) \in M_c^{gt}} FL\left( P_f(\tilde{i}, \tilde{j}), \widehat{P}_f(\tilde{i}, \tilde{j}) \right)
    \label{equ:7}
\end{equation}
where $\widehat{P}_f(\tilde{i}, \tilde{j})$ is the fine-level ground-truth matching matrix for a coarse-level match $(\tilde{i}, \tilde{j})$; $\widehat{P}_f(\tilde{i}, \tilde{j})$ is calculated at 1/2 scale similar to coarse-level ground-truth matching matrix, with the addition of a mutual nearest neighbor constraint allowing only one-to-one matches.

For the sub-pixel refinement loss, we follow XoFTR \citep{tuzcuoglu2024xoftr} to implement the symmetric epipolar distance function. This approach eliminates the need for explicit ground-truth matching pairs and enables us to supervise both matching coordinates jointly. Given an estimated matching coordinate pair $(\widehat{x}_A, \widehat{x}_B)$ in normalized image coordinates, the sub-pixel loss is defined as Equation \ref{equ:8}.

\begin{equation}
    L_{sub} = \frac{1}{|M_c^{gt}|} \sum_{(\widehat{x}_A, \widehat{x}_B)} \frac{\|\widehat{x}_A^T E \widehat{x}_B\|^2}{\|E^T \widehat{x}_A\|_{0:2}^2 + \|E \widehat{x}_B\|_{0:2}^2}
    \label{equ:8}
\end{equation}
where E is the ground-truth essential matrix obtained using camera poses. Thus, the final loss is formed as the sum of these three parts as presented in Equation \ref{equ:9}.

\begin{equation}
    L=\mu_c L_c + \mu_f L_f + \mu_{sub} L_{sub}
    \label{equ:9}
\end{equation}
where $\mu_c$, $\mu_f$ and $\mu_{sub}$ are hyperparameters and set as $1.0$, $1.0$ and $10^4$, respectively.

\subsection{Detector-free incremental SfM reconstruction framework}
\label{sec3.2}
\begin{figure}[!t]
    \centering
    \includegraphics[width=0.7\linewidth]{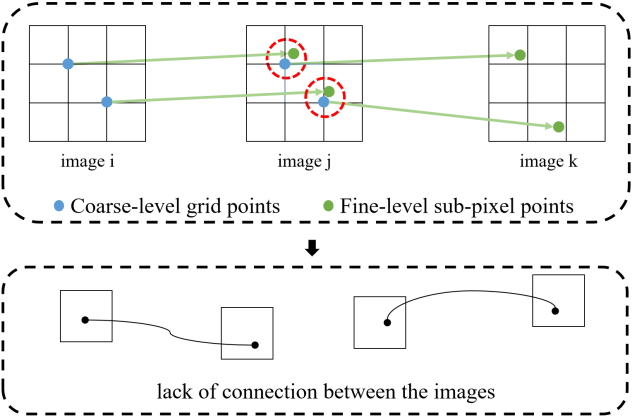}
    \caption{Multi-view geometric inconsistency.}
    \label{fig6}
 \end{figure}

Detector-free feature matching methods cannot be directly integrated into aerial-ground ISfM because of their poor feature repeatability across images, which causes the problem of fragmented feature tracks, as illustrated in Figure \ref{fig6}. In the literature, Detector-free SfM improves its adaptability with detector-free matching results by using grid quantization and iterative refinement. However, grid quantization used in coarse reconstruction reduces the positioning accuracy of matching points and affects the final ISfM reconstruction accuracy; iterative refinement, on the other hand, does not fully consider the large-scale difference and rotational characteristics between aerial and ground images, resulting in limited refinement effect on the ISfM reconstruction of aerial-ground images. To solve these problems, this study proposes a detector-free ISfM framework with two core components, i.e., fine feature tracks connection and coarse model refinement.

\subsubsection{Fine feature tracks connection}
\label{sec3.2.1}
Due to its inherent image pair processing characteristic, the matching points generated by detector-free feature matching methods have poor repeatability between different images. Existing methods mostly adopt grid quantization that forces all fine matching points to be transferred to the nearest grid corner points. The essence of this strategy is to reduce the matching positioning accuracy to improve the repeatability of feature points. This superficial improvement cannot fundamentally solve the problem of poor repeatability of matching points. Instead, it generates a large number of errors due to quantization operations, seriously affecting the success and precision of ISfM reconstruction.

\begin{figure}[!t]
    \centering
    \includegraphics[width=1.0\linewidth]{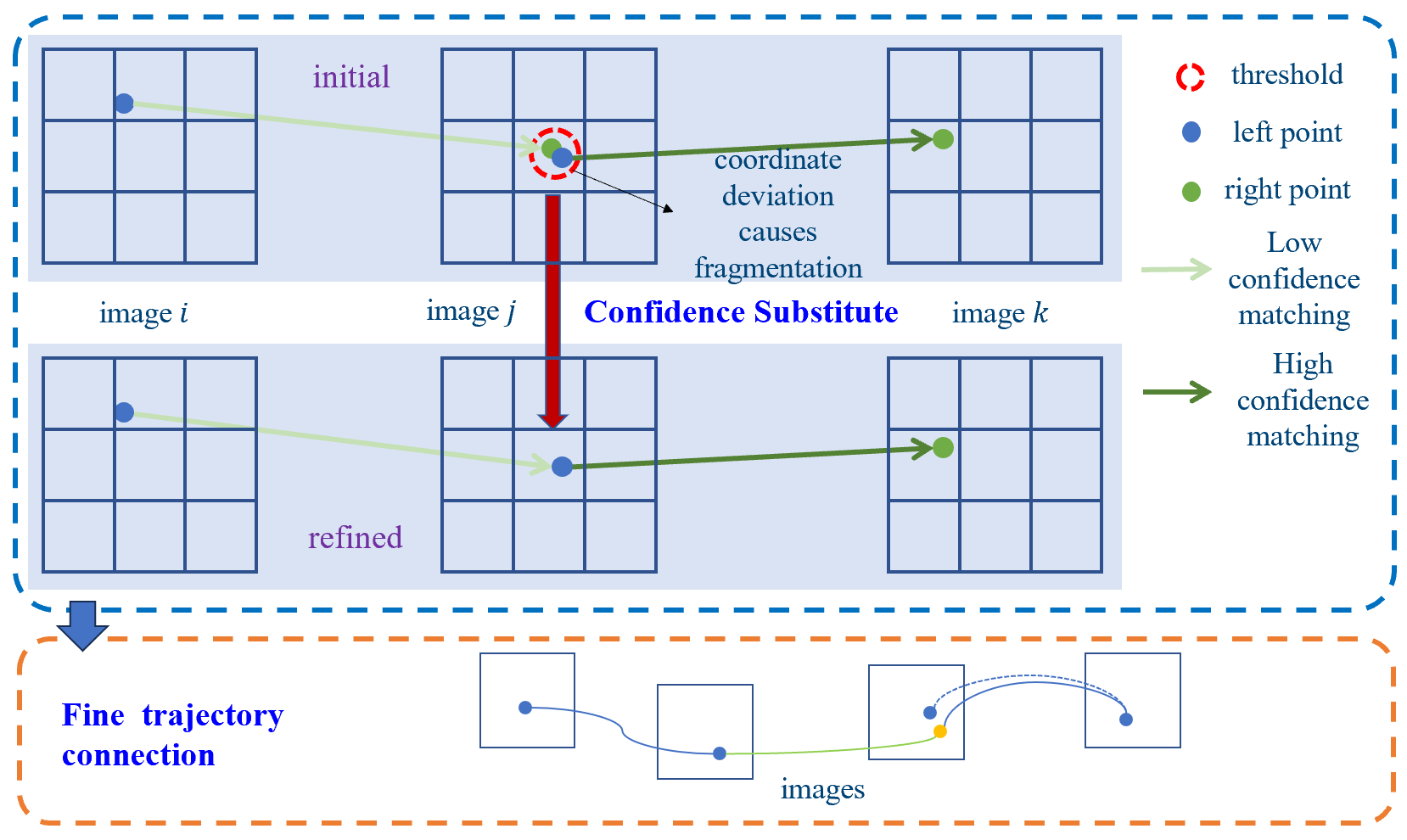}
    \caption{Fine feature tracks connection module.}
    \label{fig7}
 \end{figure}

To solve this issue, this study utilizes the core concepts of confidence substitution and track connection to combine the matching results between multiple associated image pairs. Its core it to select high-confidence feature points to connect fragmented feature tracks. This improves the overall repeatability of feature points while ensuring the accuracy of matching points, providing sufficient and accurate corresponding points for subsequent coarse point cloud construction and laying a solid foundation for subsequent fine reconstruction. The main idea of the fine feature tracks connection module is shown in Figure \ref{fig7}.

(1) Indexing of paired image matching results. For several associated image pairs such as $(i,j)$,$(j,k)$, $(i,k)$ in the aerial-ground dataset, all paired image matching results are first integrated to form an index structure of "image-feature point-confidence". Here, the feature point item includes two-dimensional pixel coordinates and its corresponding matching point index. The index structure should be able to identify the unique identifier of each feature point in its corresponding image and the matching relationship between different images. To compare the credibility of different feature matching results, the confidence value $P_f$ of the bidirectional softmax normalization in the matching stage is adopted, which can effectively reflect the reliability of the matching results.

(2) Collection of the spatial proximity of feature points. For all feature points in the intermediate associated image j, a spatial association connection threshold $d_{th}$ is set. With each feature point as the center, all other feature points within the Euclidean distance d less than $d_{th}$ ,i.e., $d \leq d_{th}$, are selected as the neighborhood set $N_j$. If there are two or more feature points in the neighborhood set $N_j$, it is considered that the current feature point may have a track connection possibility with other feature points, and the next step of confidence comparison is entered. Otherwise, the original feature points without connection are retained. As the detector-free method can achieve sub-pixel accuracy matching, the spatial connection threshold $d_{th}$ of feature points is set to 1.0 in this study.

(3) Confidence comparison and track connection. For the neighborhood feature point set $N_j$, the matching confidence $P_f$ of each feature point is first calculated, and the feature point with the maximum confidence is selected as the core point of the feature track. The matching relationships of other low-confidence feature points in the neighborhood set are assigned to the core point, which realizes the connection between high-confidence feature tracks and low-confidence feature tracks and obtains a complete feature track passing through multiple associated images. For example, the feature points in image j are spatially close with their $p_j^1 (P_f=0.85)$ and $p_j^2 (P_f=0.32)$. After updating $p_j^2$ to $p_j^1$, this step finally forms a complete track covering images $i-j-k$, improving the repeatability of feature points.

The optimized matching results will be used as the input of the ISfM reconstruction, which can output a complete and reliable initial coarse point cloud, providing high-quality input for the subsequent coarse point cloud optimization module.

\subsubsection{Coarse model refinement}
\label{sec3.2.2}
Coarse point clouds generated from initial ISfM reconstruction often suffer from lots of noises. This issue is then further compounded by large deviations driven by the severe scale and view point discrepancies between aerial-ground images, leaving the model inadequate for high-precision applications. To overcome these limitations, this study adapts the iterative refinement framework proposed by \cite{he2024detector}, tailoring it to the unique characteristics of aerial-ground image ISfM reconstruction. The framework employs a multi-stage iterative optimization workflow: feature track optimization leverages multi-view features to refine the of accuracy tracks; and geometric optimization utilizes bundle adjustment to optimize camera poses and 3D points. To optimaly balance geometric precision and computational efficiency, the pipeline executes exactly two optimization rounds with the reprojection error thresholds set sequentially to 1.5 and 1.0 pixels.

\section{Experimental results}
\label{sec4}

In this section, we conduct extensive experiments on four aerial-ground datasets and one rotated UAV dataset to validate the proposed solution. Evaluations mainly cover two aspects. First, the proposed rotation-robust detector-free matching network is compared with LoFTR on public datasets for pose estimation and image matching. Second, the ISfM framework is validated in terms of the unprocessed, quantized, and proposed track-connected matches, and the results are evaluated by using the metrics covering the completeness and precision derived from ISfM reconstruction.

All the experiments were conducted on a Linux platform equipped with a 3.0 GHz Intel Core i9-13900K CPU, 64 GB RAM, and a NVIDIA GeForce RTX 4090 with 24 GB memory. The ISfM implementation platform was the software COLMAP. All the matching results use the RANSAC for the removal of gross errors, with a threshold of 1.0 pixel. The image pairs of the five datasets were obtained from the effective matching pairs of the datasets processed by COLMAP's spatial matching method, and the numbers were 5,360, 4,120, 5,755, 5,452, and 3,715, respectively. In particular, before feature matching, the longest side of the images was uniformly scaled to 1,440 pixels to achieve maximum efficiency.

\subsection{Test sites and datasets}
\label{sec4.1}

\begin{figure}[!t]
    \centering
    \begin{minipage}[t]{0.45\linewidth}
    \centering
        \includegraphics[width=1\linewidth]{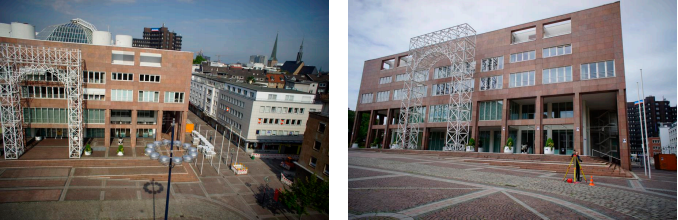} \\
        (a)  \\
    \end{minipage}
    \begin{minipage}[t]{0.45\linewidth}
    \centering
        \includegraphics[width=1\linewidth]{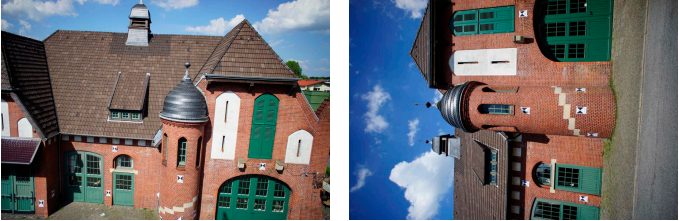} \\
        (b) \\
    \end{minipage}
    \begin{minipage}[t]{0.45\linewidth}
    \centering
        \includegraphics[width=1\linewidth]{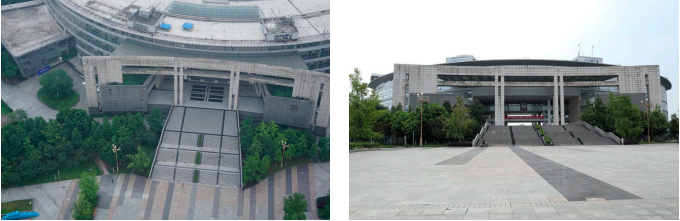} \\
        (c) \\
    \end{minipage}
    \begin{minipage}[t]{0.45\linewidth}
    \centering
        \includegraphics[width=1\linewidth]{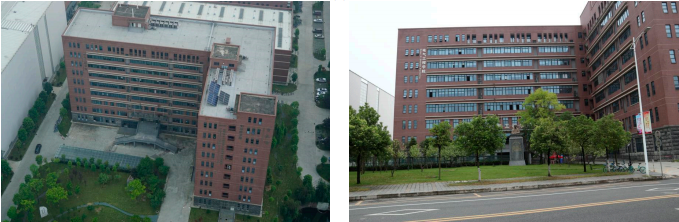} \\
        (d) \\
    \end{minipage}
    \begin{minipage}[t]{0.9\linewidth}
    \centering
        \includegraphics[width=1\linewidth]{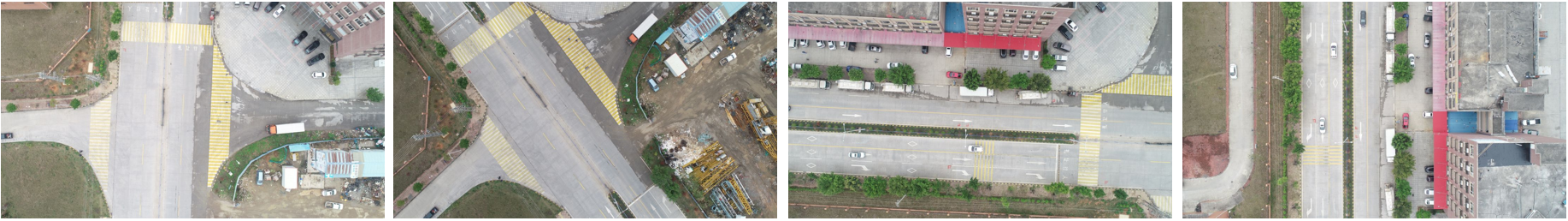} \\
        (e) \\
    \end{minipage}
    \caption{Representative images of the five datasets: (a) dataset 1: Center; (b) dataset 2: Zeche; (c) dataset 3: SWJTU-LIB; (d) dataset 4: SWJTU-BLD; (e) dataset 5: ShaoGuan.}
    \label{fig8}
 \end{figure}

The detailed information of the five datasets is presented in Table \ref{tab1}. The representative images of each dataset are illustrated in Figure \ref{fig8}. The description of each test site and the corresponding dataset is listed as follows.

\begin{table}[!t]
\centering
\caption{Detailed information on the five datasets.}
\label{tab1}
\small 
\makebox[1\linewidth]{
    \begin{tabular}{llllll}
    \toprule
    \textbf{Item Name} & \textbf{Dataset 1} & \textbf{Dataset 2} & \textbf{Dataset 3} & \textbf{Dataset 4} & \textbf{Dataset 5} \\
    \midrule
    Dataset name & Center & Zeche & SWJTU-LIB & SWJTU-BLD & ShaoGuan \\
    No. of images & \begin{tabular}[c]{@{}l@{}}ground: 203\\ aerial: 146\end{tabular} & \begin{tabular}[c]{@{}l@{}}ground: 172\\ aerial: 147\end{tabular} & \begin{tabular}[c]{@{}l@{}}ground: 78\\ aerial: 123\end{tabular} & \begin{tabular}[c]{@{}l@{}}ground: 88\\ aerial: 207\end{tabular} & 166 \\
    Camera mode & Sony NEX-7 & Sony NEX-7 & \begin{tabular}[c]{@{}l@{}}Canon EOS M6(A)\\ Sony ILCE-5100(T)\end{tabular} & \begin{tabular}[c]{@{}l@{}}Canon EOS M6(A)\\ Sony ILCE-5100(T)\end{tabular} & DJI FC6310R \\
    \begin{tabular}[c]{@{}l@{}}Focal length \\ (mm)\end{tabular} & \begin{tabular}[c]{@{}l@{}}ground: 16\\ aerial: 16\end{tabular} & \begin{tabular}[c]{@{}l@{}}ground: 16\\ aerial: 16\end{tabular} & \begin{tabular}[c]{@{}l@{}}ground: 19\\ aerial: 40\end{tabular} & \begin{tabular}[c]{@{}l@{}}ground: 18\\ aerial: 28/40\end{tabular} & 24 \\
    GSD(cm) & \begin{tabular}[c]{@{}l@{}}ground: 0.53\\ aerial: 1.10\end{tabular} & \begin{tabular}[c]{@{}l@{}}ground: 0.28\\ aerial: 0.56\end{tabular} & \begin{tabular}[c]{@{}l@{}}ground: 1.06\\ aerial: 1.69\end{tabular} & \begin{tabular}[c]{@{}l@{}}ground: 1.33\\ aerial: 1.93\end{tabular} & 2.1 \\
    \begin{tabular}[c]{@{}l@{}}Image size \\ (pixel)\end{tabular} & 6,000×4,000 & 6,000×4,000 & 6,000×4,000 & 6,000×4,000 & 5,472×3,078 \\
    \bottomrule
    \end{tabular}
}
\end{table}

\begin{itemize}
    \item Dataset 1 was collected on the buildings in the center of Dortmund, Germany, as shown in Figure \ref{fig8}(a). The ground images and aerial images recorded the appearance of the buildings from different angles, and the similarity of the buildings in front and back was very high. The overlapping part of the aerial-ground images includes the structures of the three facades of the building. As shown in Figure 8(a), there are significant differences between images taken from different perspectives.
    \item Dataset 2 was collected from a building in Zurich, Switzerland, as shown in Figure \ref{fig8}(b). The ground image captured the building facade, while the aerial image showed the building and its surrounding environment. The overlapping part of the air-ground images contains structural information of the building, ranging in length from short to long. The two perspectives of this dataset have the greatest difference, while the difference in lighting is relatively small.
    \item Dataset 3 was collected from a library building in Southwest Jiaotong University, as shown in Figure \ref{fig8}(c). The ground image represents the entrance area of the library, while the aerial image covers the top and side areas of the building. The overlapping part contains the information about the entrance. Thus, the two images have significant changes in illumination but small changes in perspective.
    \item Dataset 4 was collected for a research building in Southwest Jiaotong University, as shown in Figure \ref{fig8}(d). The ground image represents the facade of the entire building, while the aerial image shows the upper part and the surrounding area of the research building. The overlapping part is the part of the research building. Therefore, the illumination changes between the two images are significant, while the viewing angle changes are small.
    \item Dataset 5 is located on a suburban road surrounded by low buildings, as shown in Figure \ref{fig8}(e). The data collection platform used for this dataset is the DJI Phantom 4 RTK (Real-Time Dynamic Positioning) version. The onboard positioning sensor of this platform has been enhanced through the Continuous Operating Reference Stations system. It is equipped with a DJI FC6310R camera with 5,472×3,078 pixels, and 166 images were captured at a flight height of 70 m.
\end{itemize}

\subsection{Experimental metrics}
\label{sec4.2}

As listed in Table \ref{tab2}, the test targets two major components of the proposed solution, i.e., the detector-free matching network and the ISfM reconstruction framework. For the feature matching network, performance is validated across three distinct categories, including pose estimation, ablation study, and feature matching.

(1)	Pose estimation is quantified by using the AUC (Area Under the Curve) of camera pose errors at thresholds of 5$^\circ$, 10$^\circ$, and 20$^\circ$, alongside matching accuracy, which measures the proportion of correspondences with epipolar errors below 5e-4, as in \cite{sun2021loftr}.

(2)	Ablation study then uses the two identical metrics of pose estimation to rigorously verify the validation of each individual module.

(3)	Feature matching is finally evaluated by tracking the number of successfully matched image pairs, initial matches, and inliers.

For the ISfM reconstruction framework, the solution is then benchmarked across three dimensions covering feature matching, ISfM reconstruction, and geometric precision.

(1)	Track refinement is firstly determined by the number of inliers and fine matches, and the mean epipolar error.

(2)	Reconstruction quality is evaluated by using four indicators, i.e., mean track length, the number of registered images, generated 3D points, and the reprojection error.

(3)	Coarse model refinement is monitored by tracking changes in the total 3D point count and the reprojection error during iterative updates.

(4)	Fine model accuracy is assessed by using mean and Std.Dev., values of distance discrepencies to quantify the geometric errors between the reconstructed model and the reference point cloud.

\begin{table}[!t]
\centering
\caption{The details of the evaluation metrics.}
\label{tab2}
\small 
\makebox[1\linewidth]{
    \begin{tabular}{lll}
    \toprule
    \textbf{Module} & \textbf{Category} & \textbf{Metric} \\
    \midrule
    \multirow{3}{*}{\begin{tabular}[c]{@{}l@{}}Matching \\ Network\end{tabular}} & Pose estimation & Pose error AUC, matching accuracy \\
     & Ablation study & Pose error AUC, matching accuracy \\
     & Feature matching & Matching pairs, initial matches,   inliers \\
     \midrule
    \multirow{4}{*}{\begin{tabular}[c]{@{}l@{}}ISfM \\ reconstruction\end{tabular}} & Track refinement & Inliers, fine matches, mean epipolar   error. \\
     & Reconstruction quality & \begin{tabular}[c]{@{}l@{}}Mean track length, registered   images, \\ 3D points, mean reprojection error\end{tabular} \\
     & Coarse model refinement & 3D points, mean reprojection error \\
     & Fine model accuracy & Mean and Std.Dev. of distance   discrepencies \\
     \bottomrule
    \end{tabular}
}
\end{table}

\subsection{Evaluation of detector-free matching network}
\label{sec4.3}

In the evaluation, the proposed feature matching network is first trained from scrach by using the MegaDepth dataset and compared with the benchmark network LoFTR in terms of pose estimation and ablation study as well as feature matching for the aerial-ground datasets.

\subsubsection{Pose estimation}
\label{sec4.3.1}
The MegaDepth dataset is used to retrain our matching network and the original LoFTR network, keeping the key parameters such as the number of layers in the attention module and the filtering threshold for coarse matching consistent with the original implementation. Only the size of the input image is set to 480×360 pixels, thus adapting to the limitations of available memory resources. The experimental results are shown in Table \ref{tab3}.

1) \textbf{Pose error AUC}. The success probability of the pose estimation within a certain range of pose errors that the model can tolerate increases with the value of the pose error AUC. The experiment shows that the proposed method has a more obvious advantage over the LoFTR method at all pose error tolerance thresholds. For example, when the confidence threshold is 5, the AUC value of the proposed method is 0.4591, which is 93.9\% higher than that of LoFTR. This reflects that the proposed matching network can adapt to the distribution characteristics of image features with different scales and rotations.

2) \textbf{Matching accuracy}. The experimental results show that the matching accuracy of the proposed matching network is 0.9320, which is 6.9\% higher than that of LoFTR. This indicates that the model can obtain more high-quality matches that satisfy the epipolar line constraints. The matching network achieves a breakthrough in matching reliability by fully retaining the rich matching information introduced by the detector-free method.

\begin{table}[!t]
\centering
\caption{The relative orientation estimation results.}
\label{tab3}
\makebox[1\linewidth]{
    \begin{tabular}{lrrrr}
    \toprule
    \textbf{Method} & \textbf{AUC@5$^\circ$} & \textbf{AUC@10$^\circ$} & \textbf{AUC@20$^\circ$} & \textbf{Accuray@5e-04} \\
    \midrule
    LoFTR & 0.2366 & 0.3882 & 0.5259 & 0.8716 \\
    Ours & 0.4591 & 0.6229 & 0.7487 & 0.9320 \\
    \bottomrule
    \end{tabular}
}
\end{table}

\subsubsection{Ablation study}
\label{sec4.3.2}
To further verify the effectiveness of each module in the proposed network, this section designs four revised models to perform ablation tests. By gradually adding the core modules, the performance of each revised module is quantitatively analyzed. In this test, 1,500 pairs from the MegaDepth dataset are used, and the setting is consistent with the pose estimation. Besides, LoFTR is used as the baseline model. The results are shown in Table \ref{tab4}. Model 1 is added with the rotation-robust feature extraction module on the basis of LoFTR. Model2 is added with the multi-scale attention Transformer modules on the basis of Model1. The full version is the proposed network with all core modules.

1) \textbf{Rotation-robust feature extraction module}. By comparing LoFTR and model1, it is shown that Model1 has achieved significant improvements in all indicators. Especially for AUC@5$^\circ$, it has increased from 0.2366 to 0.3830. This indicates that the eight-direction feature selective scan effectively solves the problem of insufficient rotation feature extraction in existing CNNs, which can significantly enhance the model's adaptability to rotated images and provide effective features for the matching task.

2) \textbf{Multi-scale attention TransFormer Module}. Compared with the results of Model1 and Model2, the performance has been further improved. For AUC@5$^\circ$, it has increased to 0.3998, which is slightly higher than the original 0.3880. Moreover, the matching accuracy has also increased by 1.1\%, reaching 0.9492. This indicates that the quadtree attention is better than the linear attention, which can enhance the discriminative power of features and has a great effect on improving the matching accuracy.

3) \textbf{Bidirectional fine matching module}. By comparing the performance of Model2 and the full version model, the AUC of pose estimation continuously increased. For AUC@5$^\circ$, it increased from 0.3998 to 0.4591. This indicates that this module effectively solves the problem of the incorrect selection of base vectors required by traditional one-way matching. Besides, it improves the positioning accuracy of matches and greatly enhances the performance of pose estimation. However, the matching precision of the full network was slightly lower than that of Model2. The bidirectional fine matching module strictly screened most edge matching pairs and eliminated a proportion of matches, but the geometric correctness of the matching results was significantly improved.

\begin{table}[!t]
\centering
\caption{Results of the ablation study.}
\label{tab4}
\makebox[1\linewidth]{
    \begin{tabular}{lrrrr}
    \toprule
    \textbf{Method} & \textbf{AUC@5$^\circ$} & \textbf{AUC@10$^\circ$} & \textbf{AUC@20$^\circ$} & \textbf{Accuray@5e-04} \\
    \midrule
    LoFTR & 0.2366 & 0.3882 & 0.5259 & 0.8716 \\
    Model1 & 0.3830 & 0.5582 & 0.7019 & 0.9381 \\
    Model2 & 0.3998 & 0.5731 & 0.7142 & 0.9492 \\
    Ours & 0.4591 & 0.6229 & 0.7487 & 0.9320 \\
    \bottomrule
    \end{tabular}
}
\end{table}

The results of the ablation study verified the rationality and effectiveness of each core module. The rotation-robust feature extraction module solved the problem of insufficient rotation feature capture. The multi-scale attention transform module enhanced the ability of feature correlation modeling. The bidirectional fine matching module improved the matching and positioning accuracy. Their collaborative effect enables the full network to achieve performance improvement in both pose estimation and matching accuracy.

\subsubsection{Feature matching}
\label{sec4.3.3}
By using the retrained network, this study conducts a comparison for feature matching. The matching results are shown in Table \ref{tab5}.

1) \textbf{Matching pairs}. It shows that the number of matching pairs from the proposed matching network is obviously larger than that of LoFTR. This indicates that the matching ability of the proposed method increases significantly, especially for aerial-ground datasets with large perspective and scale differences. For dataset 5 with a large number of rotated images, the number of matching pairs of the proposed method still increases by 11.2\%. This verifies that the proposed network has strong resilience to viewpoint, scale, and rotation changes in aerial-ground and rotated images.

2) \textbf{Initial matches}. The proposed network shows an order-of-magnitude improvement over LoFTR in initial matches across all datasets, which demonstrates the superiority of its feature matching precision. For dataset 5, the number of matches of the proposed network is 4-5 times that of LoFTR. This shows that by using robust feature extraction and multi-scale attention updating, the proposed network can accurately detect the image features under various rotation scales, resulting in more unique feature matching points.

3) \textbf{Inliers}. The results demonstrate that the number of inliers in the proposed network is significantly higher than that of LoFTR, indicating that this network enhances both the quantity and accuracy of matches. The number of inliers for the five datasets is 4-5 times higher than LoFTR. This shows that the proposed network uses bidirectional exact matching to increase the positioning accuracy of the matches, and there are still many inliers even after the removal of outliers.

\begin{table}[!t]
\centering
\caption{Feature matching results for the five datasets.}
\label{tab5}
\makebox[1\linewidth]{
    \begin{tabular}{llrrr}
    \toprule
    \textbf{Dataset} & \textbf{Method} & \textbf{Matching pairs} & \textbf{Initial matches} & \textbf{Inliers} \\
    \midrule
    \multirow{2}{*}{1} & LoFTR & 4,248 & 4,314,199 & 3,301,478 \\
     & Ours & 5,209 & 21,456,898 & 14,892,311 \\
    \multirow{2}{*}{2} & LoFTR & 3,447 & 5,098,968 & 3,738,795 \\
     & Ours & 3,778 & 25,139,455 & 19,152,896 \\
    \multirow{2}{*}{3} & LoFTR & 5,020 & 2,701,094 & 1,875,710 \\
     & Ours & 5,747 & 14,690,390 & 9,347,289 \\
    \multirow{2}{*}{4} & LoFTR & 4,764 & 2,142,026 & 1,334,100 \\
     & Ours & 5,356 & 12,576,511 & 6,817,404 \\
    \multirow{2}{*}{5} & LoFTR & 3,233 & 2,835,124 & 2,296,121 \\
     & Ours & 3,596 & 12,372,724 & 9,917,067 \\
     \bottomrule
    \end{tabular}
}
\end{table}

\begin{figure}[!t]
    \centering
    \begin{minipage}[t]{0.45\linewidth}
    \centering
        \includegraphics[width=1\linewidth]{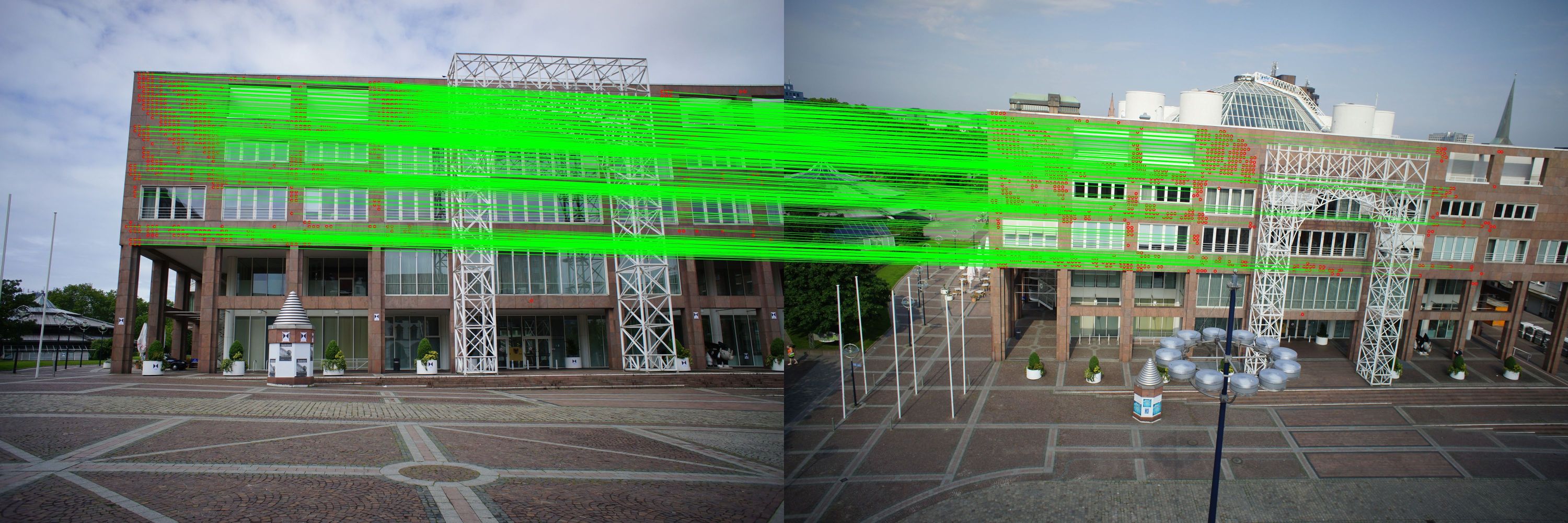} \\
        (a) LoFTR (575/742) 
    \end{minipage}
    \begin{minipage}[t]{0.45\linewidth}
    \centering
        \includegraphics[width=1\linewidth]{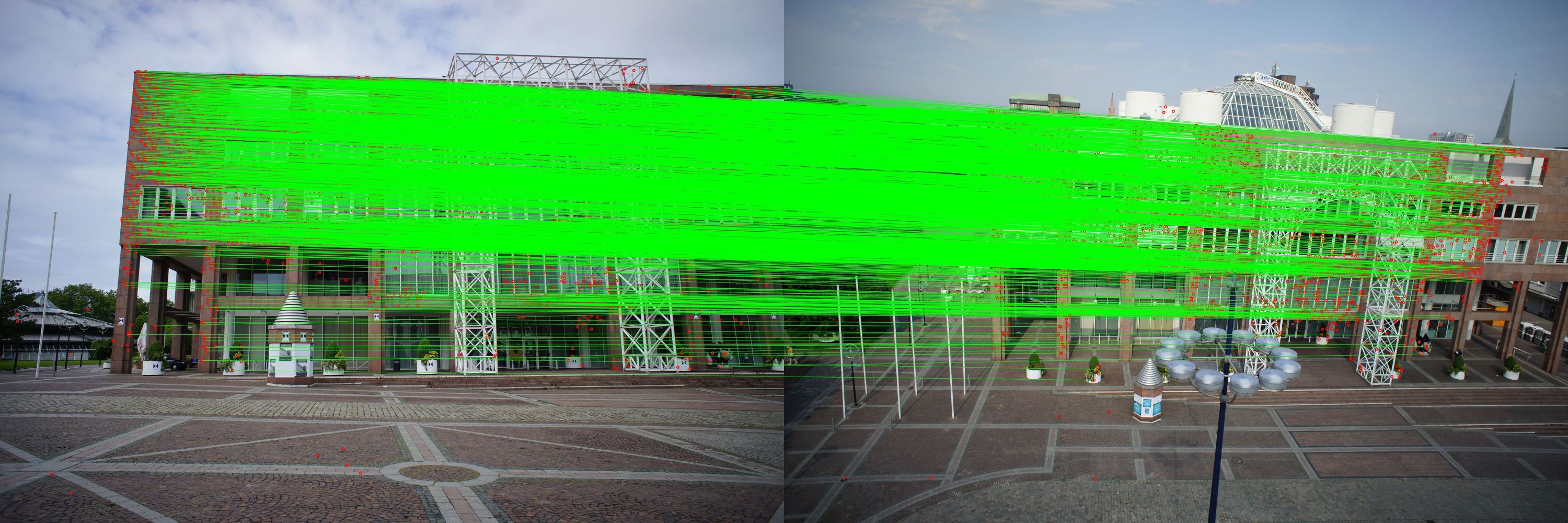} \\
        (b) Ours (1810/2,738)\\
    \end{minipage}
    \begin{minipage}[t]{0.45\linewidth}
    \centering
        \includegraphics[width=1\linewidth]{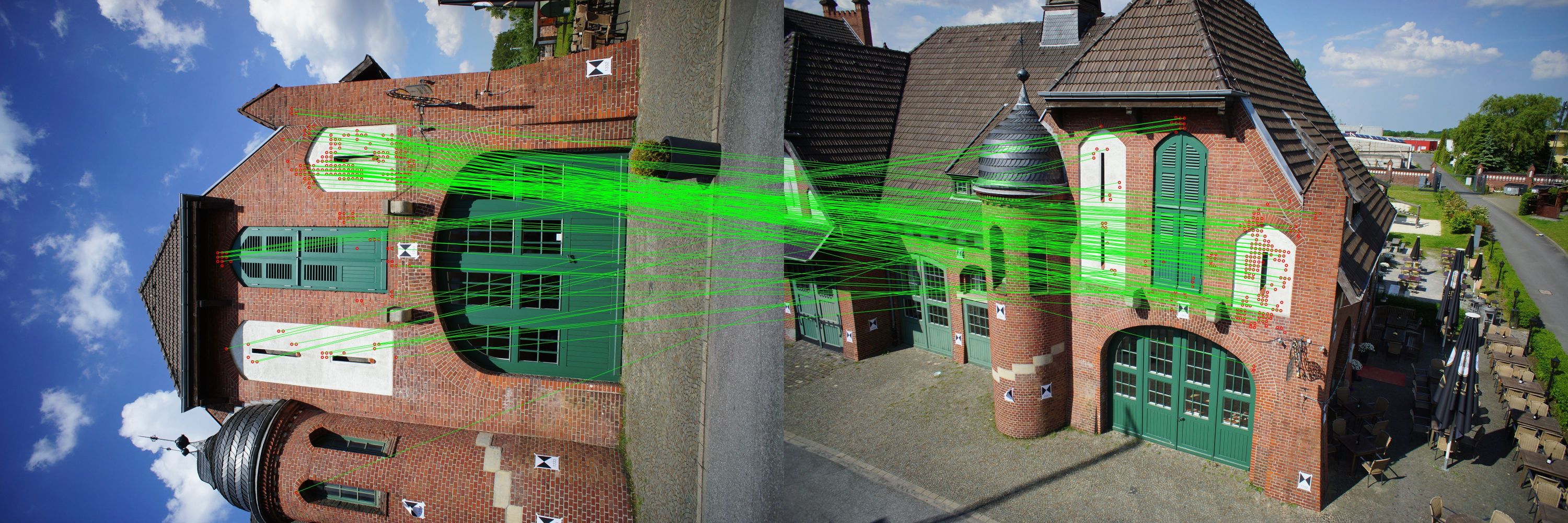} \\
        (c) LoFTR (130/200) 
    \end{minipage}
    \begin{minipage}[t]{0.45\linewidth}
    \centering
        \includegraphics[width=1\linewidth]{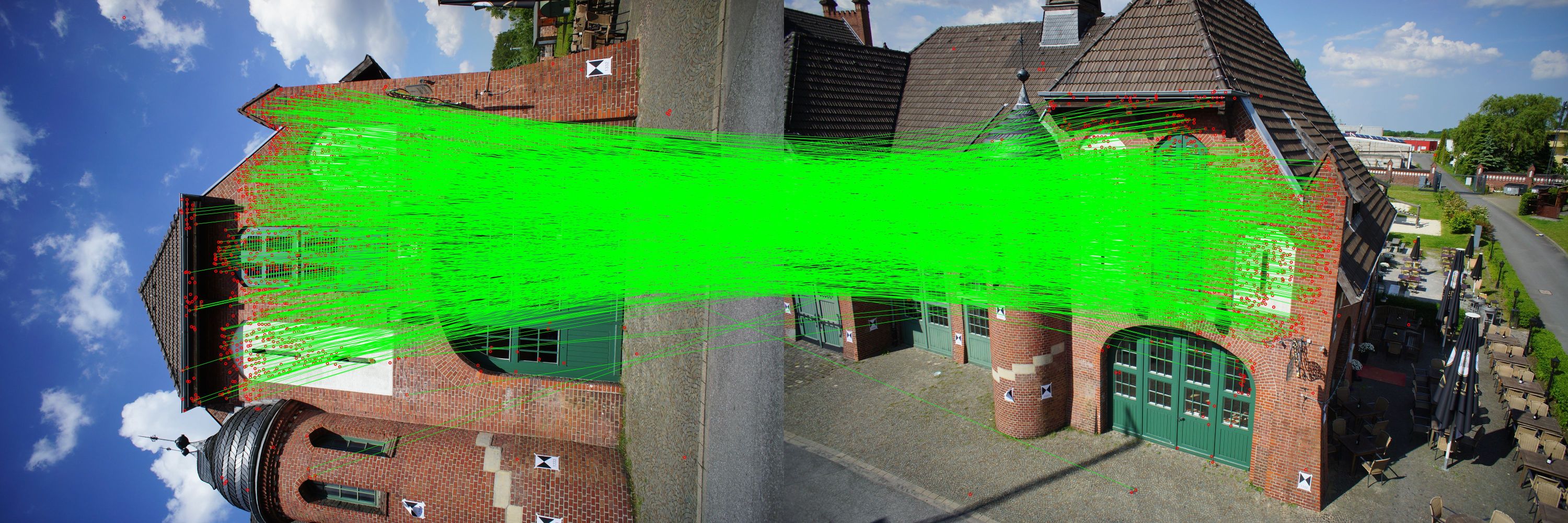} \\
        (d) Ours (1,211/1,783)\\
    \end{minipage}
    \begin{minipage}[t]{0.45\linewidth}
    \centering
        \includegraphics[width=1\linewidth]{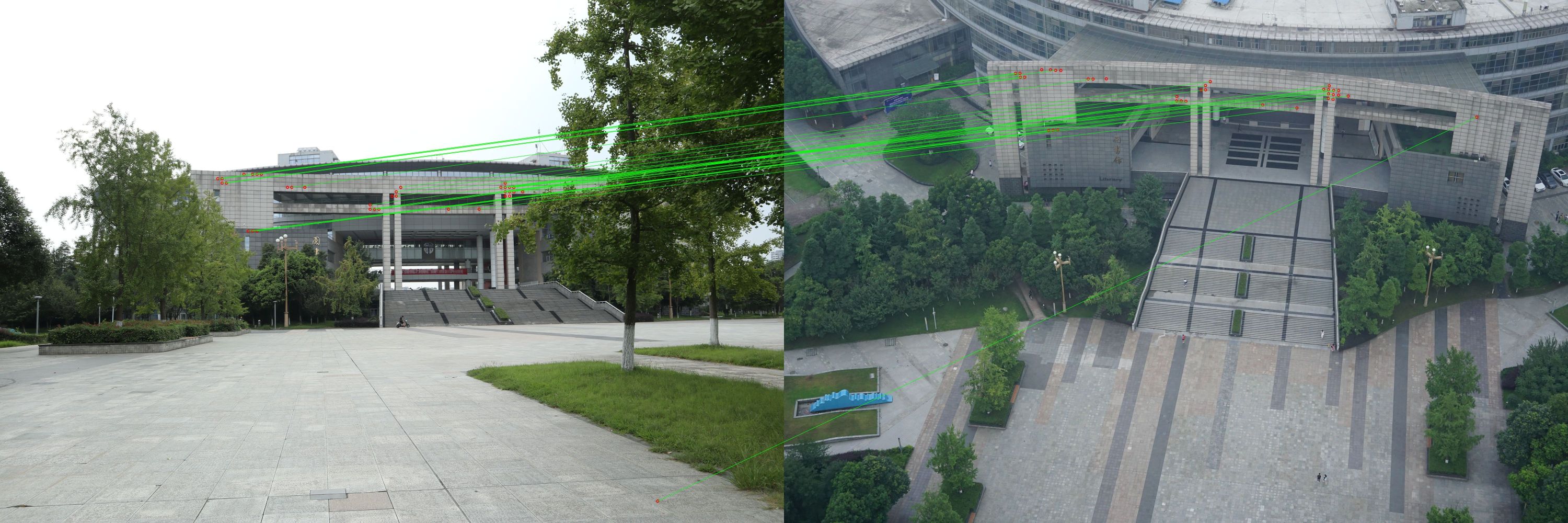} \\
        (e) LoFTR (27/36) 
    \end{minipage}
    \begin{minipage}[t]{0.45\linewidth}
    \centering
        \includegraphics[width=1\linewidth]{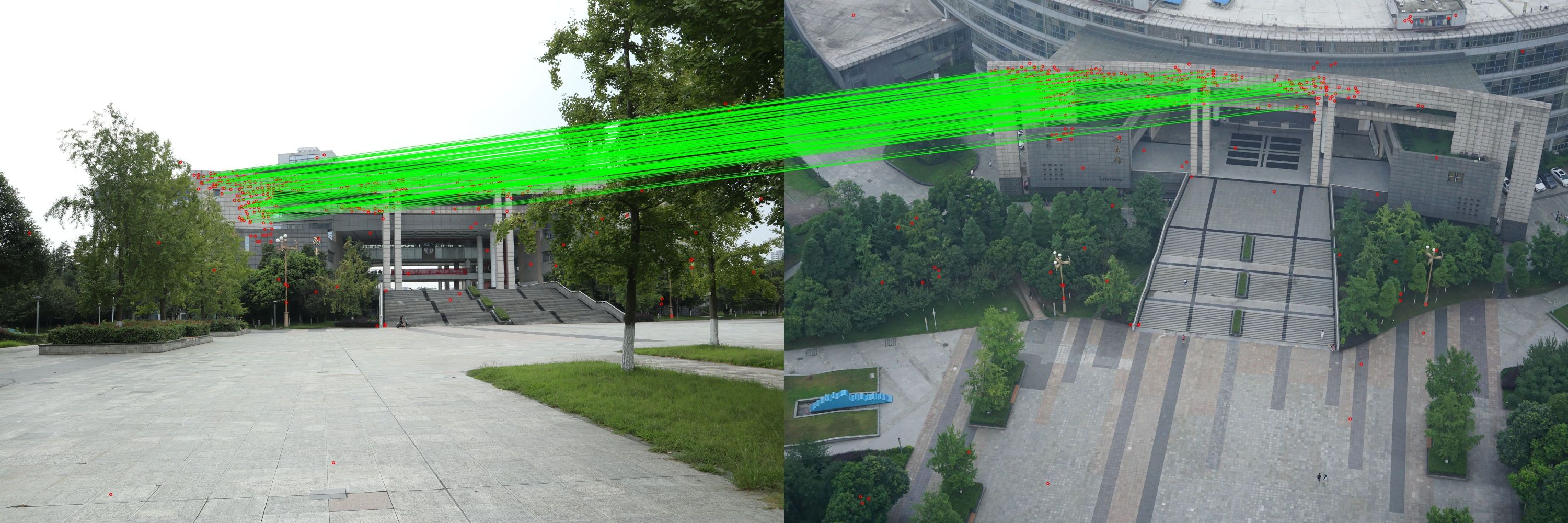} \\
        (f) Ours (215/366)\\
    \end{minipage}
    \begin{minipage}[t]{0.45\linewidth}
    \centering
        \includegraphics[width=1\linewidth]{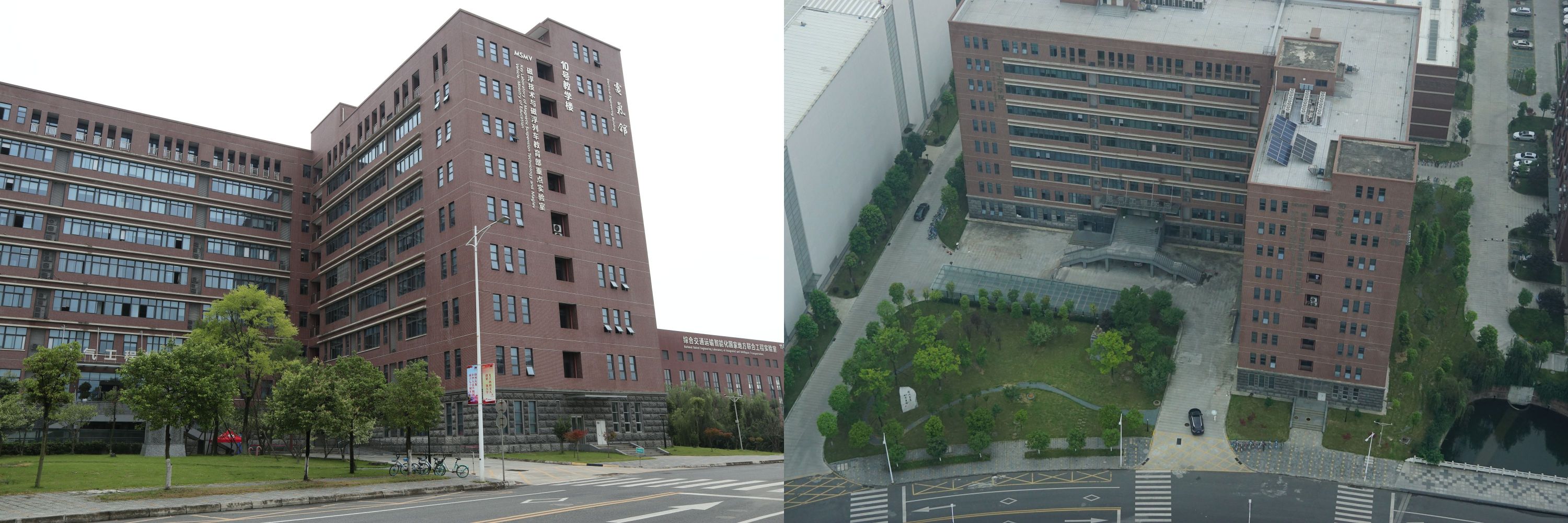} \\
        (g) LoFTR (0/0) 
    \end{minipage}
    \begin{minipage}[t]{0.45\linewidth}
    \centering
        \includegraphics[width=1\linewidth]{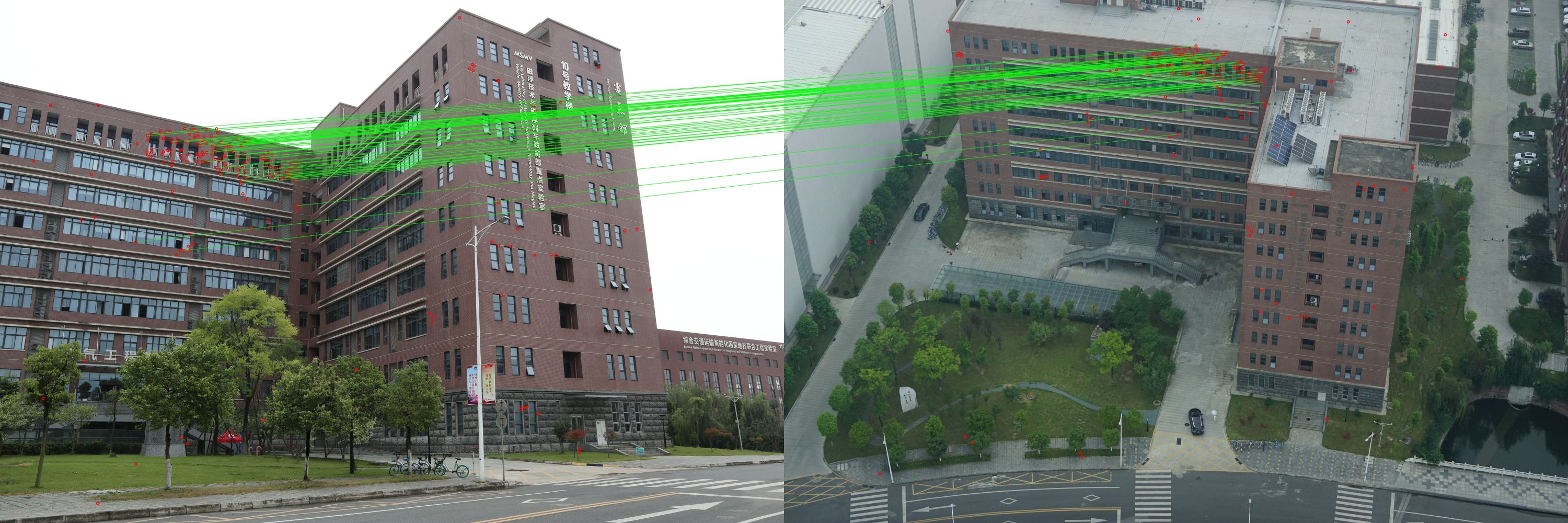} \\
        (h) Ours (60/197)\\
    \end{minipage}
    \caption{Matching results of the normal overlapped images from datasets 1, 2, 3 and 4. The value in the bracket indicates the number of initial matches and inliers, respectively.}
    \label{fig9}
\end{figure}

\begin{figure}[!t]
    \centering
    \begin{minipage}[t]{0.45\linewidth}
    \centering
        \includegraphics[width=1\linewidth]{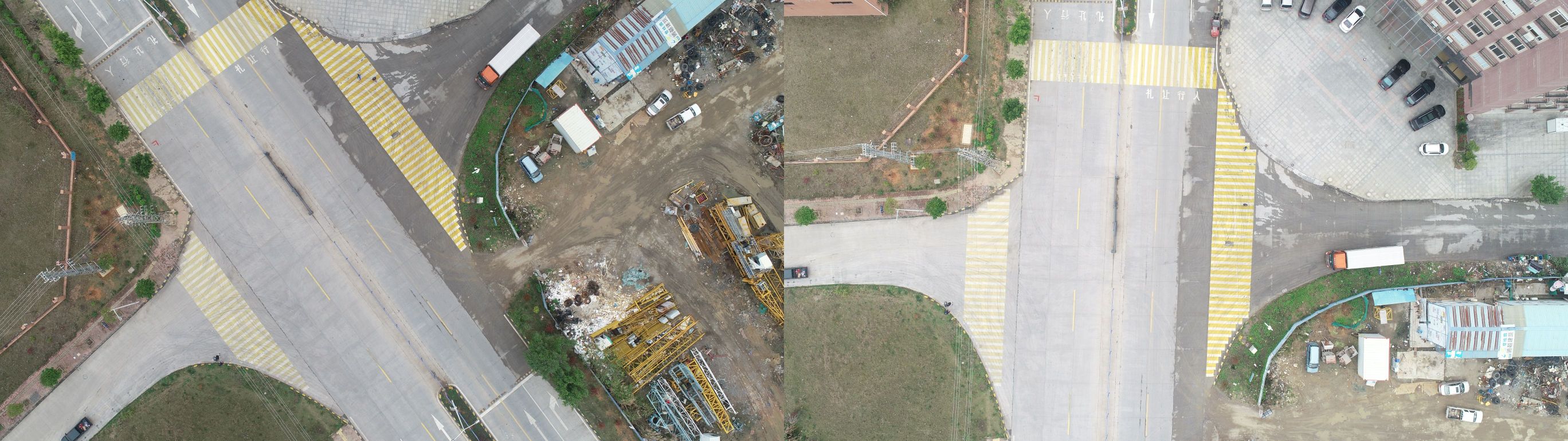} \\
        (a) LoFTR (0/0) 
    \end{minipage}
    \begin{minipage}[t]{0.45\linewidth}
    \centering
        \includegraphics[width=1\linewidth]{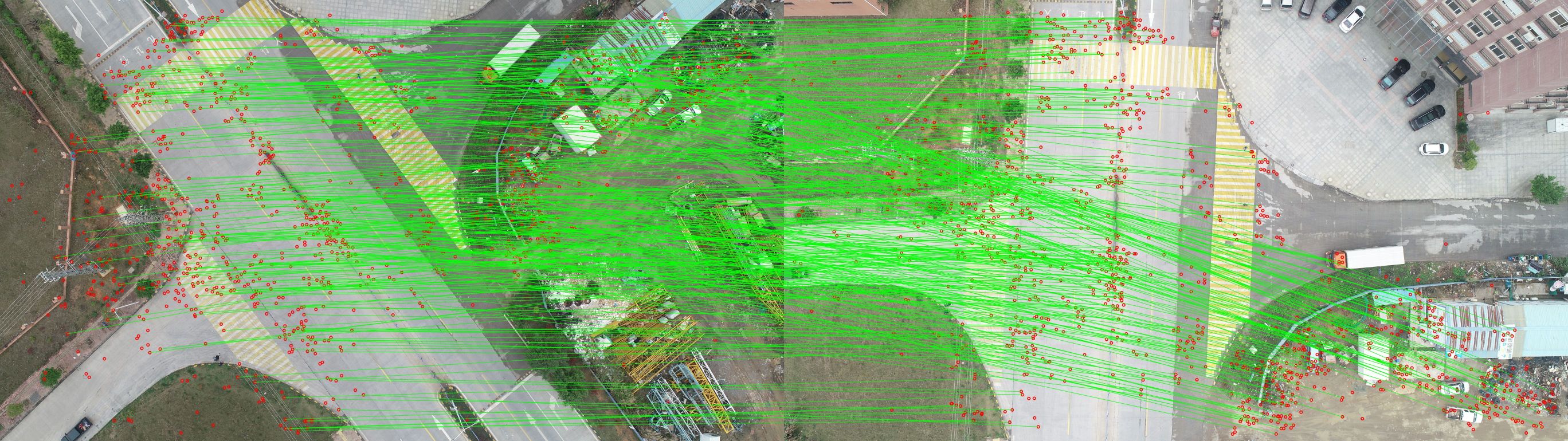} \\
        (b) Ours (392/1,166)\\
    \end{minipage}
    \begin{minipage}[t]{0.45\linewidth}
    \centering
        \includegraphics[width=1\linewidth]{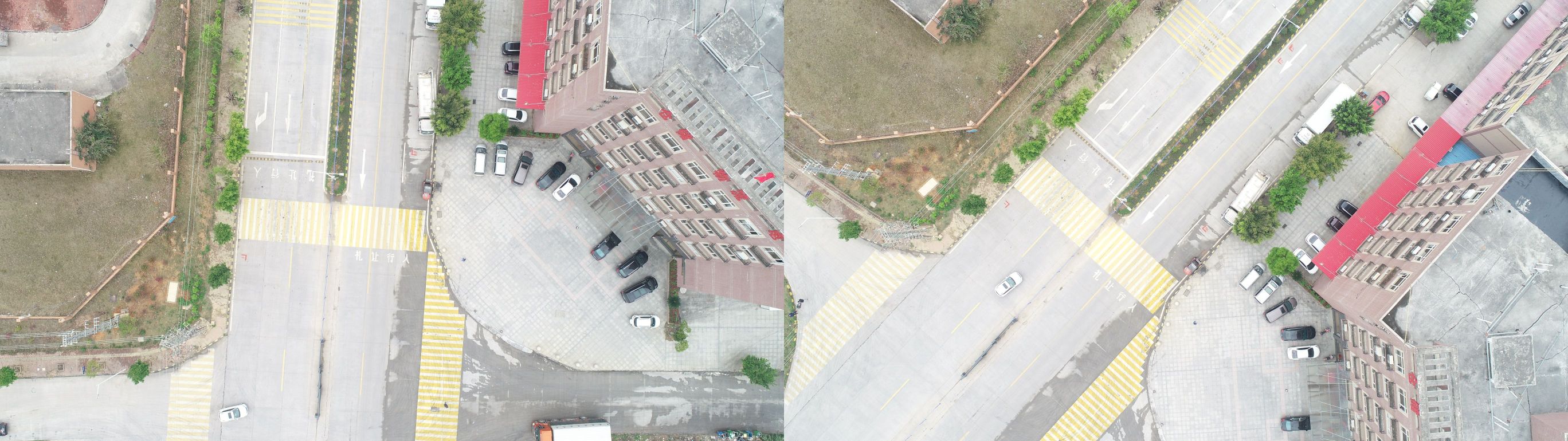} \\
        (c) LoFTR (0/0) 
    \end{minipage}
    \begin{minipage}[t]{0.45\linewidth}
    \centering
        \includegraphics[width=1\linewidth]{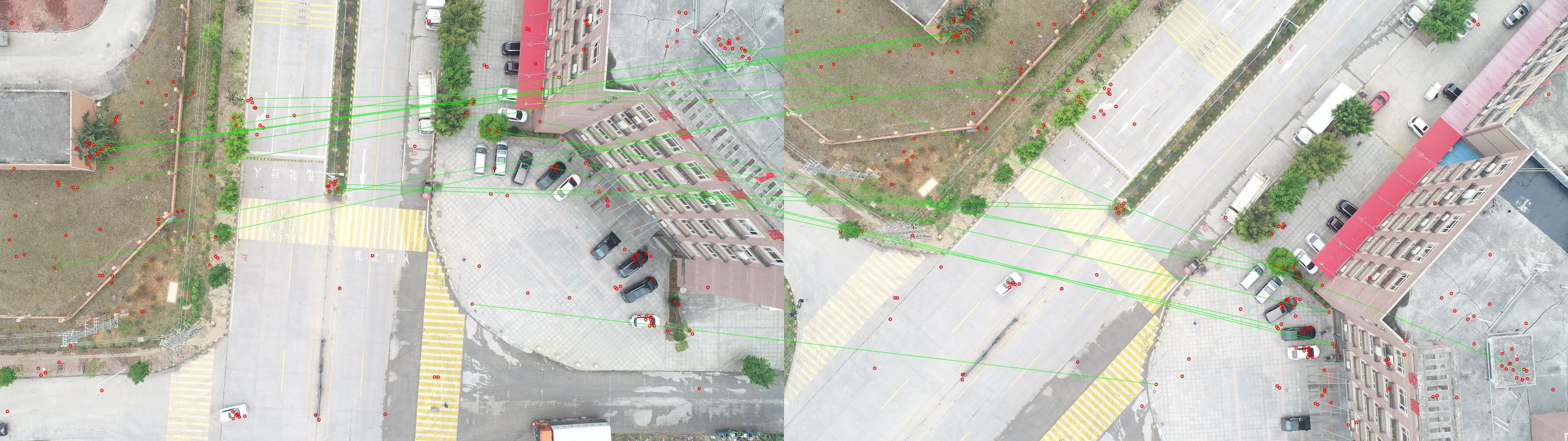} \\
        (d) Ours (22/180)\\
    \end{minipage}
    \caption{Matching results of the rotated images from dataset 5. The value in the bracket indicates the number of initial matches and inliers, respectively.}
    \label{fig10}
\end{figure}

To further verify the robustness of the precise matching ability of the proposed matching network for rotated images and normal overlapped images, this study first selects four pairs of normal overlapped images from datasets 1, 2, 3 and 4, and then selects one pair of overlapped images with obvious rotation from dataset 5. Based on the selected images, we further conduct analysis for feature matching. The visual results are shown in Figure \ref{fig9} and Figure \ref{fig10}. The results show that for the rotated image pairs, LoFTR cannot achieve effective matching, which demonstrates the deficiency of LoFTR in extracting features in rotation scenarios. On the contrary, the proposed network can achieve stable results on the two rotated image pairs. The main reason is that through the selective scanning of eight directions of features, the proposed network can accurately capture features under different rotation angles. It breaks the dependence of LoFTR on fixed feature directions and solves the core problem of detector-free methods in rotation scenarios.

\subsection{Evaluation of ISfM reconstruction}
\label{sec4.4}
This section evaluates the proposed ISfM reconstruction framework to verify individual module effectiveness and overall performance. In this test, the pre-trained models are used for detector-free matching methods. The evaluation pipeline proceeds sequentially through four steps. First, initial matches are generated using ASpanFormer and Efficient LoFTR for the four aerial-ground datasets. Second, initial matches are processed by using three track connection strategies, i.e., unprocessed, grid quantization, and the proposed approach. Third, the refined matches are then fed into the ISfM framework to generate coarse 3D point clouds, which are used to compare baseline reconstruction quality. Fourth, the iterative refinement framework is applied to these coarse models to generate fine point clouds, which are used to verify the efficacy of the refinement module. Finally, the refined point cloud is aligned with the LiDAR point cloud to evaluate their geometric precision.

In this evaluation, SIFT+MNN and SuperPoint+MNN serve as the baseline benchmarks. For brevity, the method naming rule is abbreviated as follows: original matching results are denoted as "ori", grid-quantized results as "gq" (with the quantization scale indicated in parentheses), and results optimized by our track connection strategy as "con". Additionally, efficient LoFTR is abbreviated as ELoFTR. Method variants are named using an underscore format combining the matching architecture and the corresponding processing strategy.

\subsubsection{Track refinement}
\label{sec4.4.1}
Detector-free feature matching methods usually cause fragmented tracks, mainly due to their low repeatability of feature points from varying images. Thus, this study proposes the track refinement module and compares its performance with other strategies. Table \ref{tab6} presents the results of feature matching after track refinement.

1) \textbf{Inliers and fine matches}. The number of inliers and fine matches of detector-free methods is significantly higher than that of handcrafted and detector-based methods. Among detector-free methods, grid quantization obviously reduces the number of matches, and it becomes more severe with the increase of the grid-quantized scales, which can be observed by comparing the methods with the suffix “gq(4)” and “gq(1)”. While the proposed strategy only slightly decreases the number of inliers, the number of fine matches remains above 90\% of the original inliers. It has been proven that the matching advantage of the detector-free method solves the problem of the grid quantization strategy of filtering inlier matches.

2) \textbf{Mean epipolar error}. SIFT has the best matching accuracy, with the mean epipolar error being the lowest in all datasets. This is due to its precise feature positioning accuracy. SuperPoint+MNN has the second-best accuracy. It is still not as good as handcrafted features, although it's superior to detector-free methods. For methods with the suffix “gq”, increasing the quantization scale also leads to an increase in matching error. On the contrary, the mean error of the proposed strategy is basically the same as the original result, such as 1.311 and 1.312, respectively, for ELoFTR\_ori and ELoFTR\_con in dataset 1. This indicates that the proposed strategy does not affect the accuracy of the final matches, achieving the dual goals of retaining the number of matches and maintaining high matching accuracy.

\begin{table}[!htp]
	\centering
	\caption{Track refinement for the four aerial-ground datasets (error unit in pixels).}
	\label{tab6}
    \footnotesize
	\makebox[0.5\linewidth]{
		\begin{tabular}{lrrrr}
        \toprule
        \textbf{Dataset} & \textbf{Method} & \textbf{Inliers} & \textbf{Fine matches} & \begin{tabular}[c]{@{}l@{}}\textbf{Epipolar} \\ \textbf{error}\end{tabular} \\
        \midrule
        \multirow{10}{*}{1} & SIFT+MNN & 2,620,550 & 2,572,012 & 1.087 \\
         & SuperPoint+MNN & 3,037,327 & 3,017,246 & 1.215 \\
         & ASpanFormer\_ori & 17,209,121 & 17,125,481 & 1.272 \\
         & ASpanFormer\_gq(1) & 13,138,946 & 13,087,678 & 1.290 \\
         & ASpanFormer\_gq(4) & 11,637,498 & 11,596,112 & 1.307 \\
         & ASpanFormer\_con & 17,177,934 & 17,093,964 & 1.273 \\
         & ELoFTR\_ori & 14,188,605 & 14,129,305 & 1.311 \\
         & ELoFTR\_gq(1) & 7,364,925 & 7,340,406 & 1.552 \\
         & ELoFTR\_gq(4) & 7,362,853 & 7,337,851 & 1.559 \\
         & ELoFTR\_con & 14,152,279 & 14,090,165 & 1.312 \\
         \midrule
        \multirow{10}{*}{2} & SIFT+MNN & 7,189,342 & 7,149,693 & 0.805 \\
         & SuperPoint+MNN & 6,870,792 & 6,847,560 & 0.837 \\
         & ASpanFormer\_ori & 20,755,415 & 20,709,363 & 1.049 \\
         & ASpanFormer\_gq(1) & 15,598,761 & 15,567,077 & 1.090 \\
         & ASpanFormer\_gq(4) & 7,942,398 & 7,915,668 & 0.959 \\
         & ASpanFormer\_con & 20,697,129 & 20,648,577 & 1.050 \\
         & ELoFTR\_ori & 19,087,211 & 19,032,169 & 1.181 \\
         & ELoFTR\_gq(1) & 8,615,654 & 8,597,773 & 1.463 \\
         & ELoFTR\_gq(4) & 6,810,338 & 6,785,568 & 0.917 \\
         & ELoFTR\_con & 19,067,038 & 19,011,514 & 1.184 \\
         \midrule
        \multirow{10}{*}{3} & SIFT+MNN & 2,506,495 & 2,495,785 & 0.581 \\
         & SuperPoint+MNN & 2,622,705 & 2,621,265 & 0.583 \\
         & ASpanFormer\_ori & 11,490,361 & 11,484,528 & 0.584 \\
         & ASpanFormer\_gq(1) & 9,128,400 & 9,125,046 & 0.631 \\
         & ASpanFormer\_gq(4) & 3,917,555 & 3,910,645 & 0.883 \\
         & ASpanFormer\_con & 11,472,762 & 11,466,477 & 0.584 \\
         & ELoFTR\_ori & 10,642,060 & 10,637,219 & 0.613 \\
         & ELoFTR\_gq(1) & 5,211,464 & 5,207,880 & 0.797 \\
         & ELoFTR\_gq(4) & 3,084,452 & 3,076,787 & 1.005 \\
         & ELoFTR\_con & 10,610,726 & 10,605,769 & 0.615 \\
         \midrule
        \multirow{10}{*}{4} & SIFT+MNN & 2,010,582 & 2,003,262 & 0.536 \\
         & SuperPoint+MNN & 2,258,134 & 2,253,917 & 0.576 \\
         & ASpanFormer\_ori & 8,666,316 & 8,654,811 & 0.626 \\
         & ASpanFormer\_gq(1) & 7,153,243 & 7,145,437 & 0.648 \\
         & ASpanFormer\_gq(4) & 3,960,427 & 3,953,873 & 0.612 \\
         & ASpanFormer\_con & 8,653,896 & 8,642,196 & 0.625 \\
         & ELoFTR\_ori & 7,909,540 & 7,899,929 & 0.640 \\
         & ELoFTR\_gq(1) & 4,243,864 & 4,239,220 & 0.739 \\
         & ELoFTR\_gq(4) & 3,034,698 & 3,029,535 & 0.667 \\
         & ELoFTR\_con & 7,898,592 & 7,886,275 & 0.644 \\
         \bottomrule
        \end{tabular}
	}
\end{table}

\begin{figure}[!t]
    \centering
    \begin{minipage}[t]{0.45\linewidth}
    \centering
        \includegraphics[height=0.13\linewidth,left]{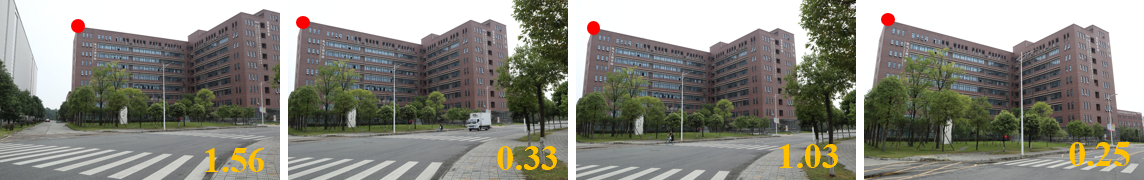} \\
        (a) SIFT+MNN 
    \end{minipage}
    \begin{minipage}[t]{0.45\linewidth}
    \centering
        \includegraphics[height=0.13\linewidth,left]{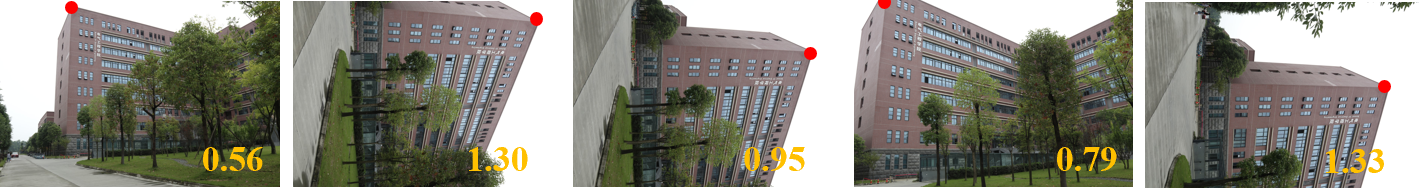} \\
        (b) SuperPoint+MNN\\
    \end{minipage}
    \begin{minipage}[t]{0.45\linewidth}
    \centering
        \includegraphics[height=0.13\linewidth,left]{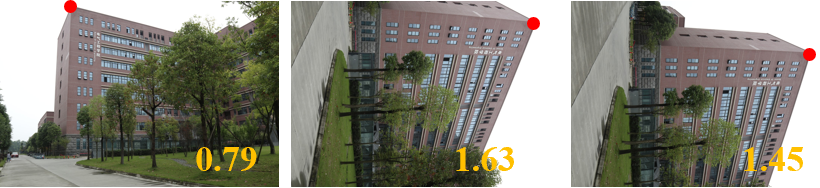} \\
        (c) ASpanFormer\_ori 
    \end{minipage}
    \begin{minipage}[t]{0.45\linewidth}
    \centering
        \includegraphics[height=0.13\linewidth,left]{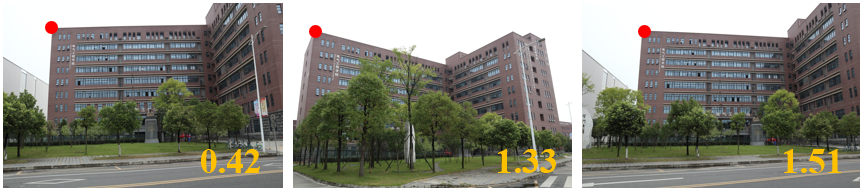} \\
        (g) ELoFTR\_ori \\
    \end{minipage}
    \begin{minipage}[t]{0.45\linewidth}
    \centering
        \includegraphics[height=0.13\linewidth,left]{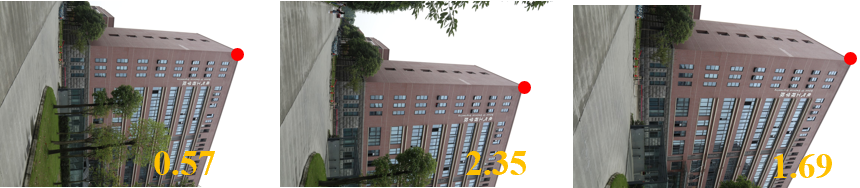} \\
        (d) ASpanFormer\_gq(1) 
    \end{minipage}
    \begin{minipage}[t]{0.45\linewidth}
    \centering
        \includegraphics[height=0.13\linewidth,left]{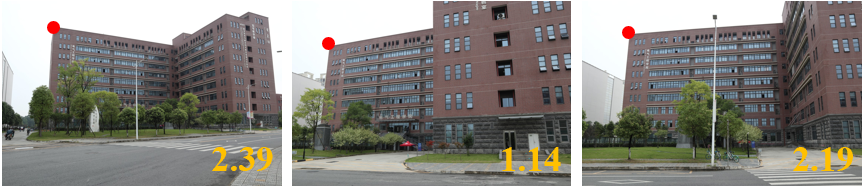} \\
        (h) ELoFTR\_gq(1) \\
    \end{minipage}
    \begin{minipage}[t]{0.45\linewidth}
    \centering
        \includegraphics[height=0.13\linewidth,left]{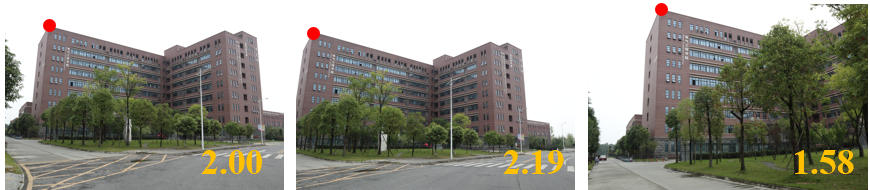} \\
        (e) ASpanFormer\_gq(4) 
    \end{minipage}
    \begin{minipage}[t]{0.45\linewidth}
    \centering
        \includegraphics[height=0.13\linewidth,left]{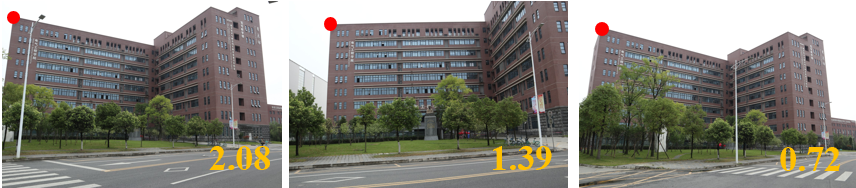} \\
        (i) ELoFTR\_gq(4) \\
    \end{minipage}
    \begin{minipage}[t]{0.45\linewidth}
    \centering
        \includegraphics[height=0.26\linewidth,left]{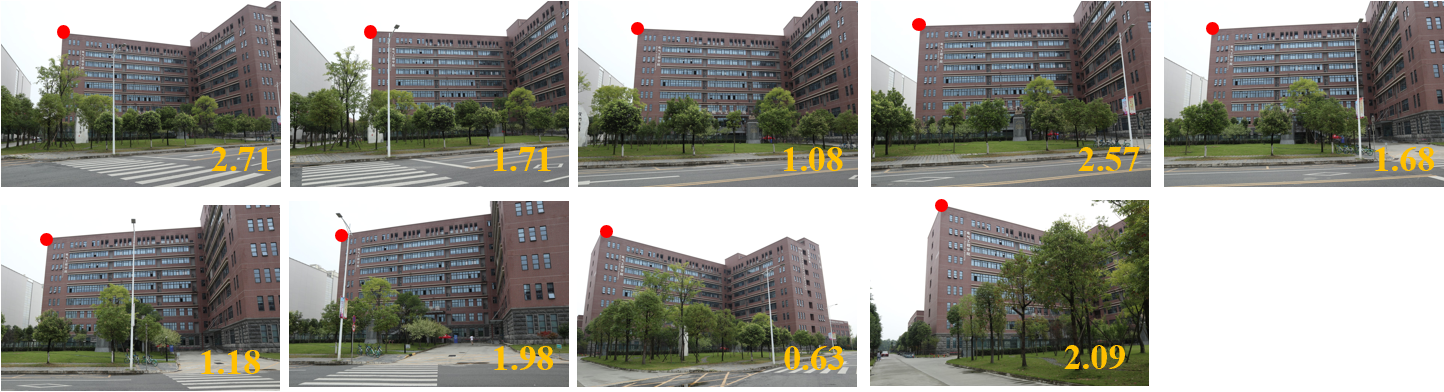} \\
        (f) ASpanFormer\_con
    \end{minipage}
    \begin{minipage}[t]{0.45\linewidth}
    \centering
        \includegraphics[height=0.26\linewidth,left]{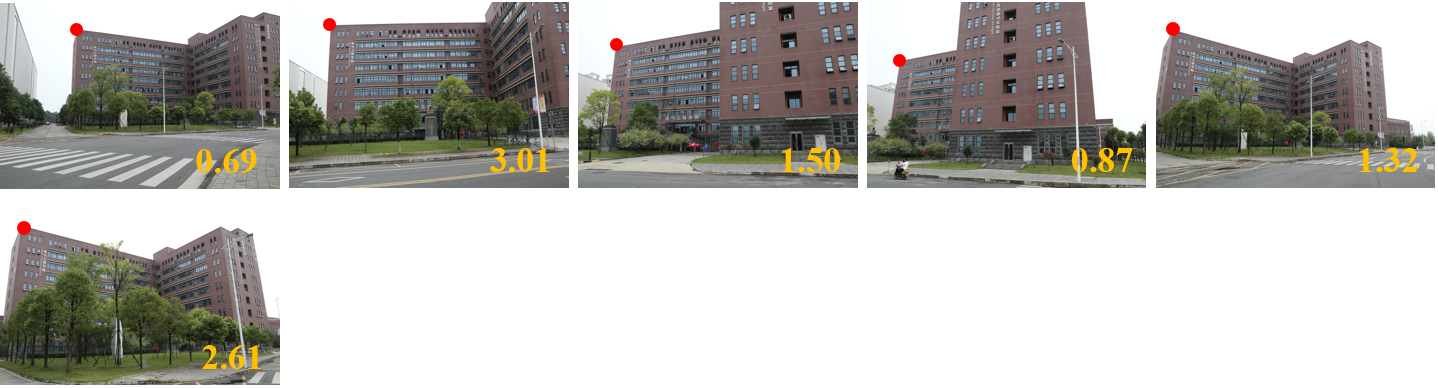} \\
        (j) ELoFTR\_con \\
    \end{minipage}
    \caption{The illustration of the track refinement module for one track point in dataset 4. The value in each sub-figure indicates the reprojection error of the track point on the image plane.}
    \label{fig11}
\end{figure}

For further visual analysis, Figure \ref{fig11} illustrates the qualitative performance of the track refinement module for a single track point in dataset 4, with yellow text indicating the reprojection error on each image plane. We can see that traditional handcrafted methods, e.g., SIFT+MNN and SuperPoint+MNN, track the feature across only a small subset of views with length ranging from 4 to 5 images. While detector-free architectures like ASpanFormer\_ori and ELoFTR\_ori can capture longer dependencies, their standard implementations suffer from severe track fragmentation, maintaining connections across only 3 images before failing. After using the grid quantization strategy, ASpanFormer\_gq and ELoFTR\_gq still degrade tracking performance, yielding high, erratic reprojection errors, e.g., up to 2.39 pixels, due to their spatial discretization errors. Conversely, our proposed track connection strategy successfully resolves this fragmentation bottleneck, as verified by ASpanFormer\_con and ELoFTR\_con. It bridges disjoint sub-tracks to continuously trace the feature point across a much larger sequence of views. The results demonstrate the module's capacity to substantially boost feature track repeatability and spatial consistency for the subsequent ISfM-based 3D reconstruction.

\subsubsection{Reconstruction quality}
\label{sec4.4.2}
\begin{figure}[!t]
    \centering
    \begin{minipage}[t]{0.3\linewidth}
    \centering
        \includegraphics[width=1\linewidth]{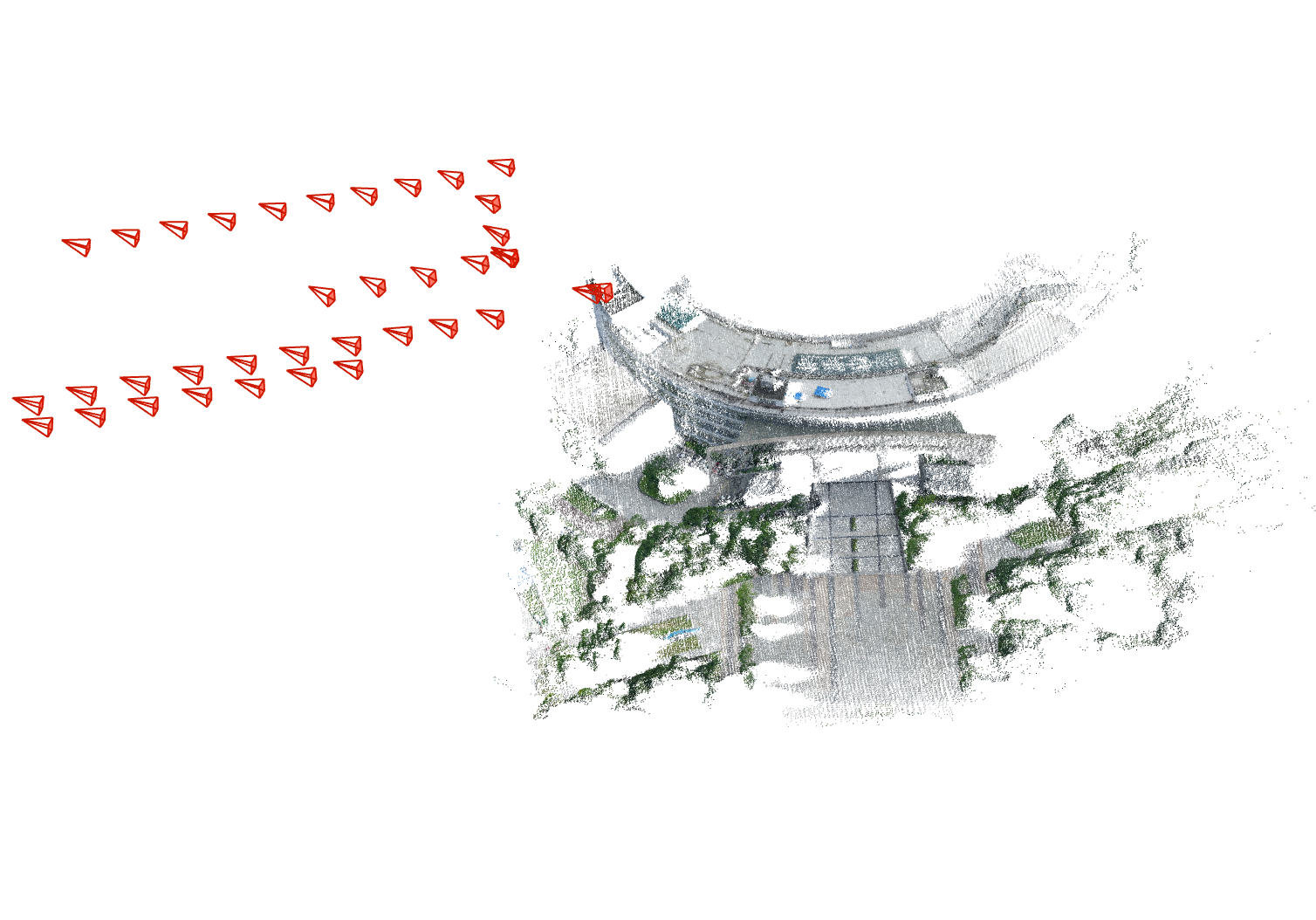} \\
        (a)
    \end{minipage}
    \begin{minipage}[t]{0.3\linewidth}
    \centering
        \includegraphics[width=1\linewidth]{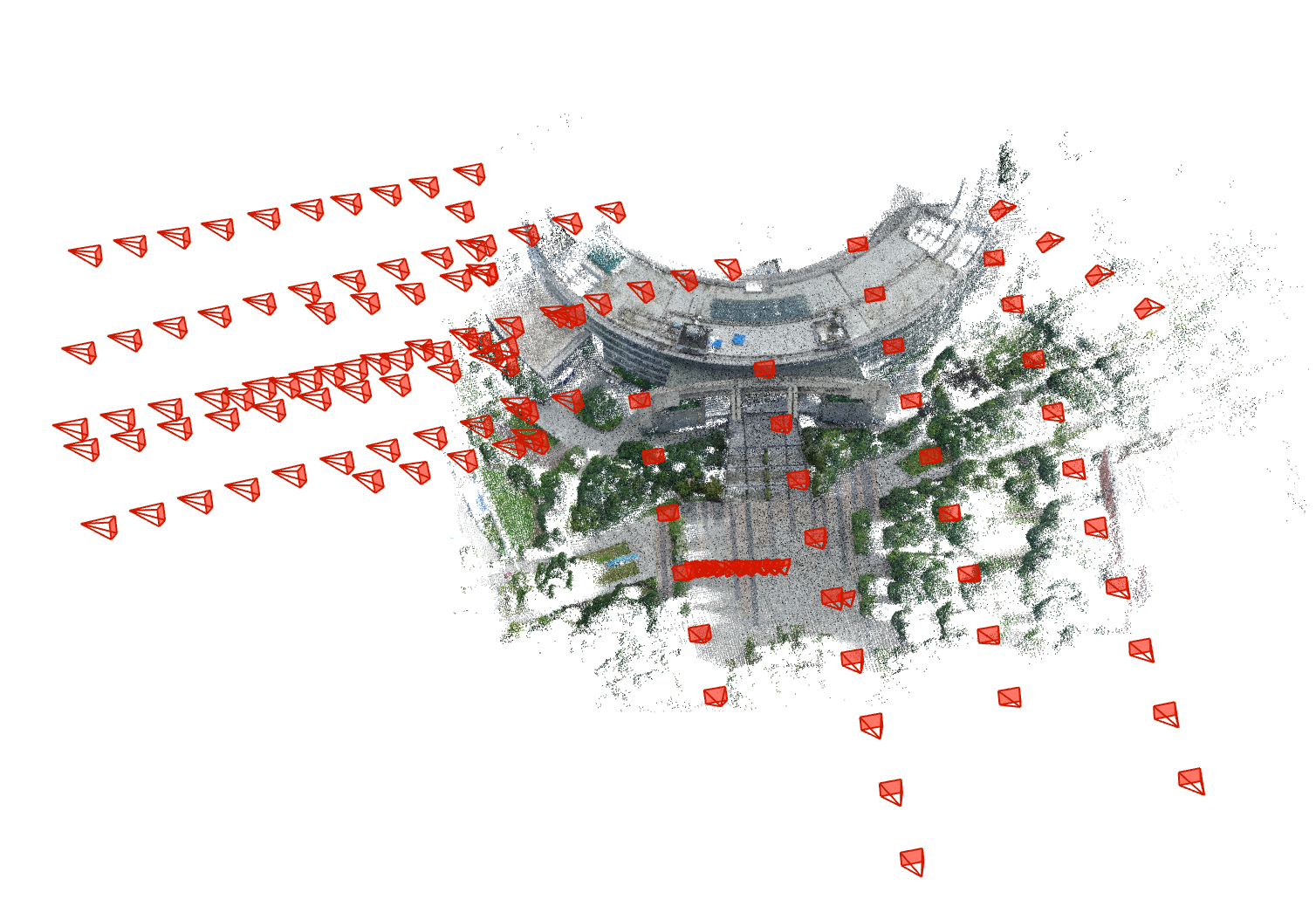} \\
        (b) 
    \end{minipage}
    \begin{minipage}[t]{0.3\linewidth}
    \centering
        \includegraphics[width=1\linewidth]{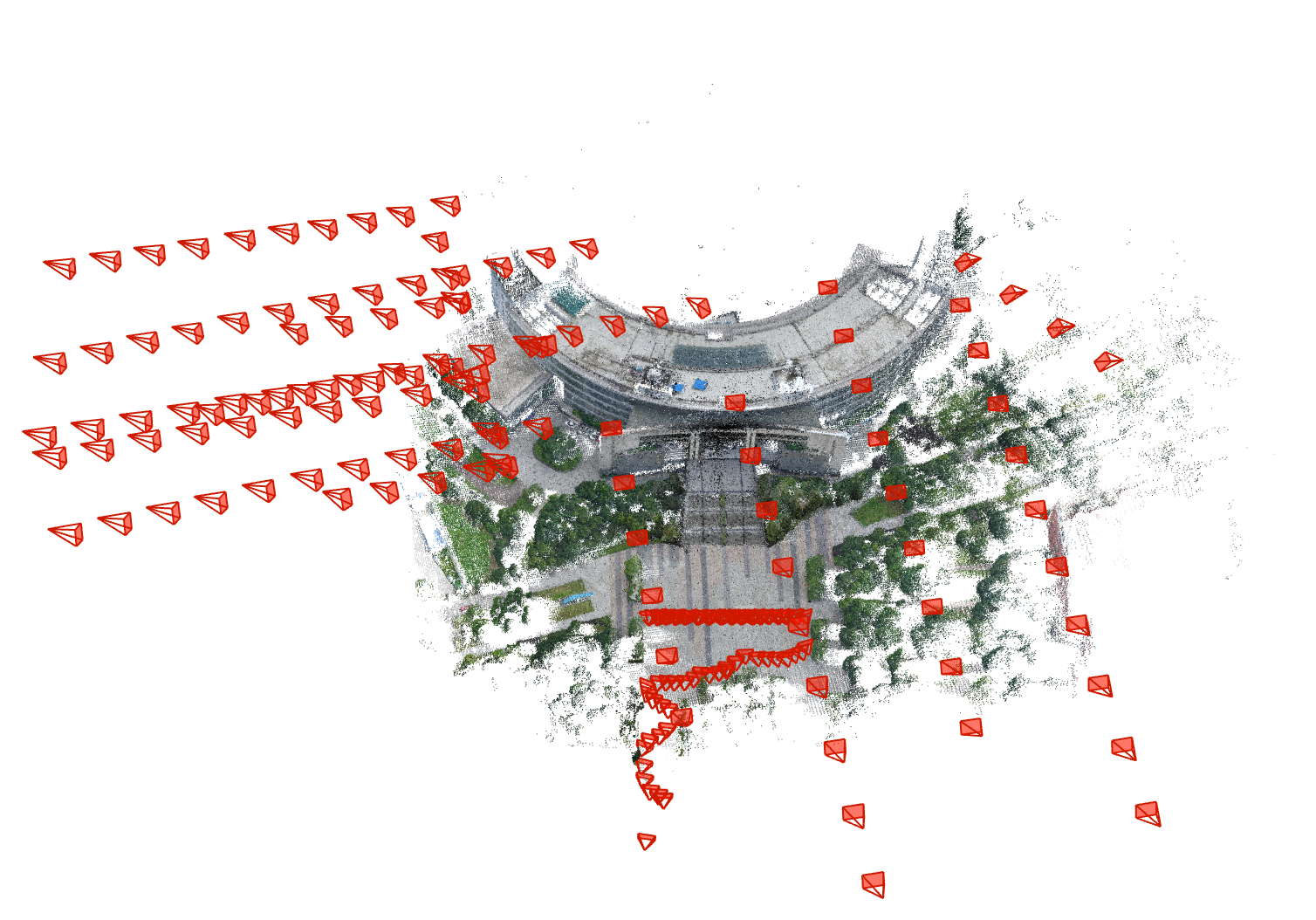} \\
        (c) 
    \end{minipage}
    \begin{minipage}[t]{0.3\linewidth}
    \centering
        \includegraphics[width=1\linewidth]{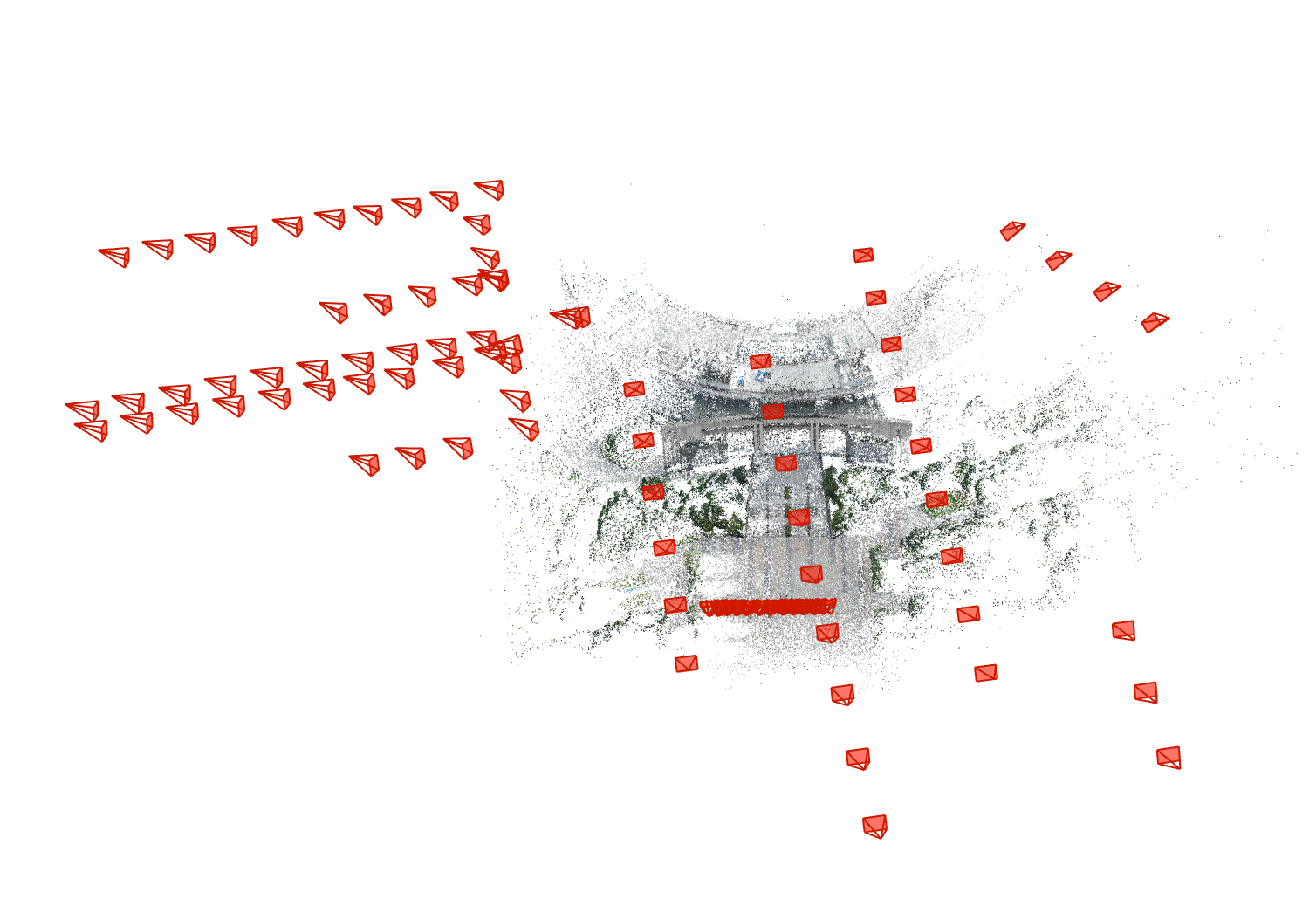} \\
        (d) 
    \end{minipage}
    \begin{minipage}[t]{0.3\linewidth}
    \centering
        \includegraphics[width=1\linewidth]{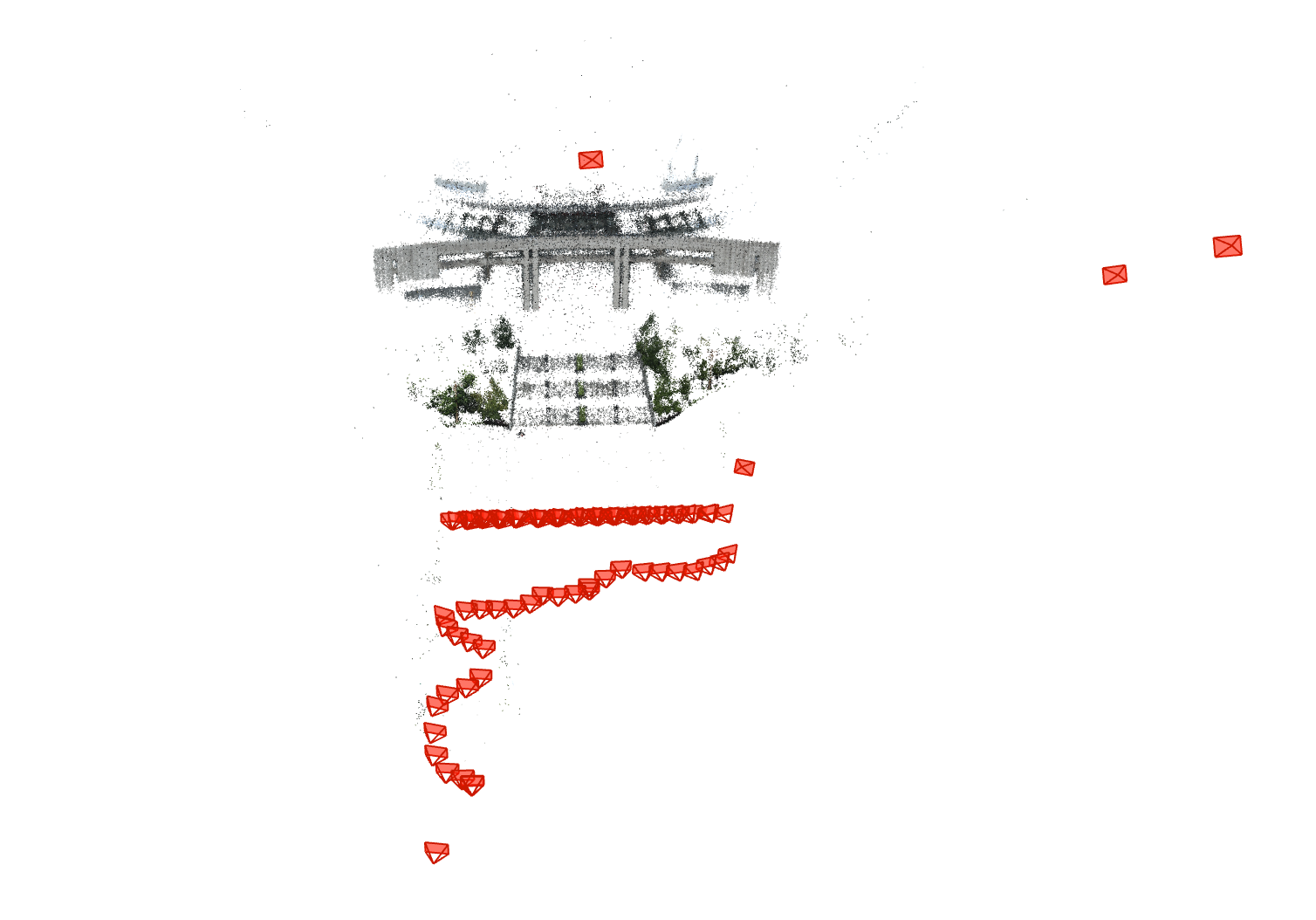} \\
        (e) 
    \end{minipage}
    \begin{minipage}[t]{0.3\linewidth}
    \centering
        \includegraphics[width=1\linewidth]{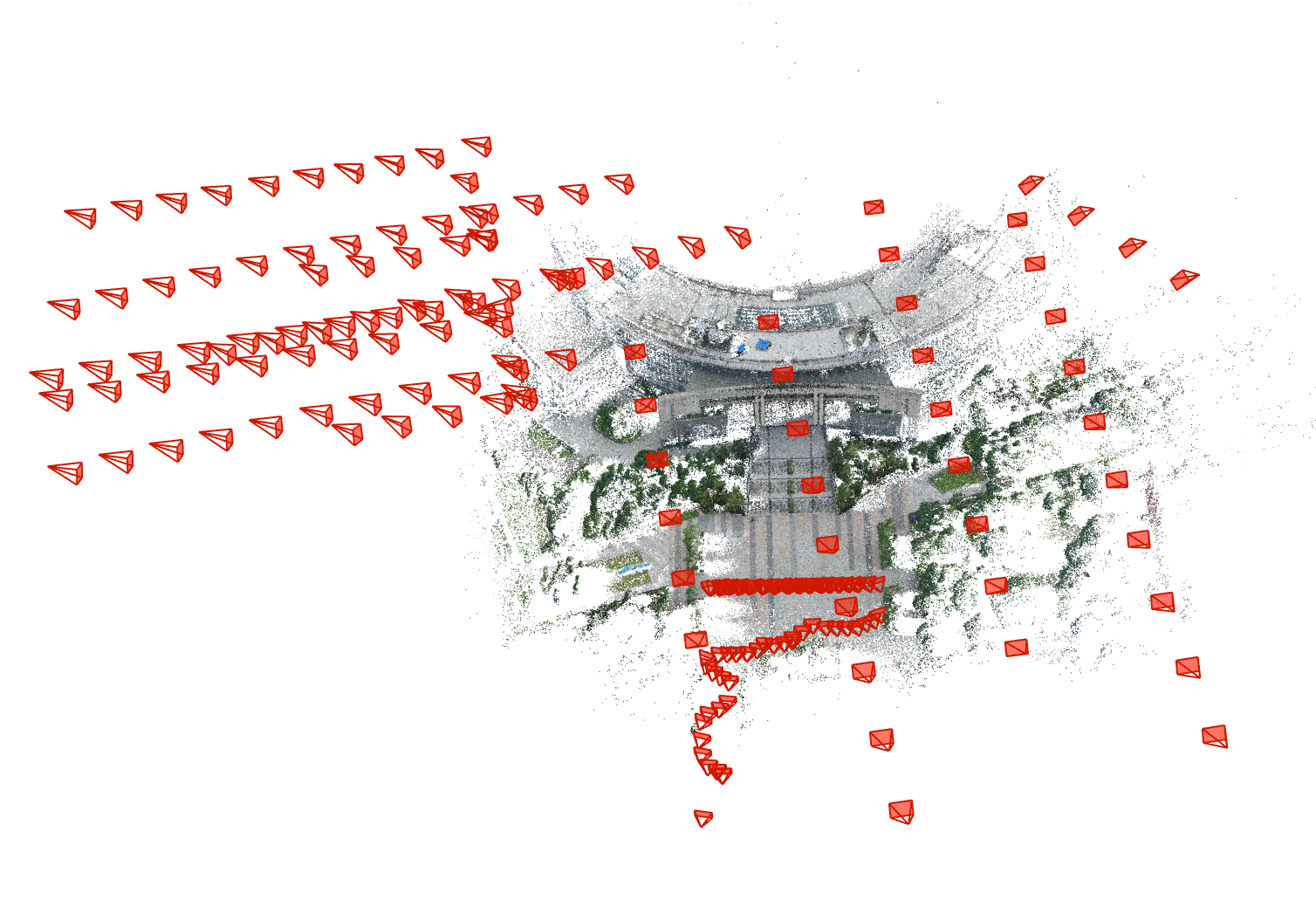} \\
        (f) 
    \end{minipage}
    \caption{ISfM reconstruction for dataset 3: (a) ASpanFormer\_ori; (b) ASpanFormer\_gq(1); (c) ASpanFormer\_con; (d) ELoFTR\_ori; (e) ELoFTR\_gq(1); (f) ELoFTR\_con.}
    \label{fig12}
\end{figure}

\begin{figure}[!t]
    \centering
    \begin{minipage}[t]{0.3\linewidth}
    \centering
        \includegraphics[width=1\linewidth]{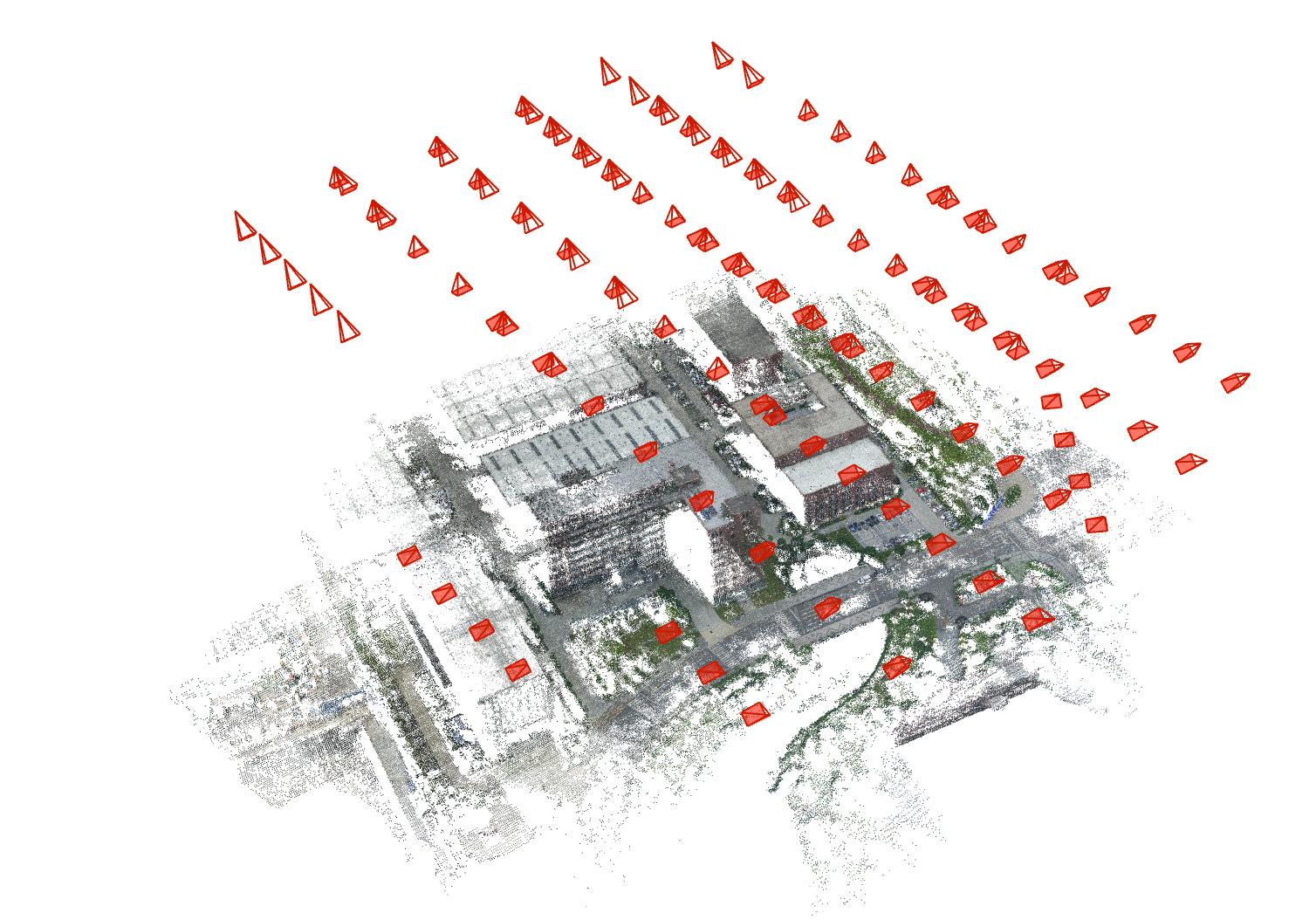} \\
        (a)
    \end{minipage}
    \begin{minipage}[t]{0.3\linewidth}
    \centering
        \includegraphics[width=1\linewidth]{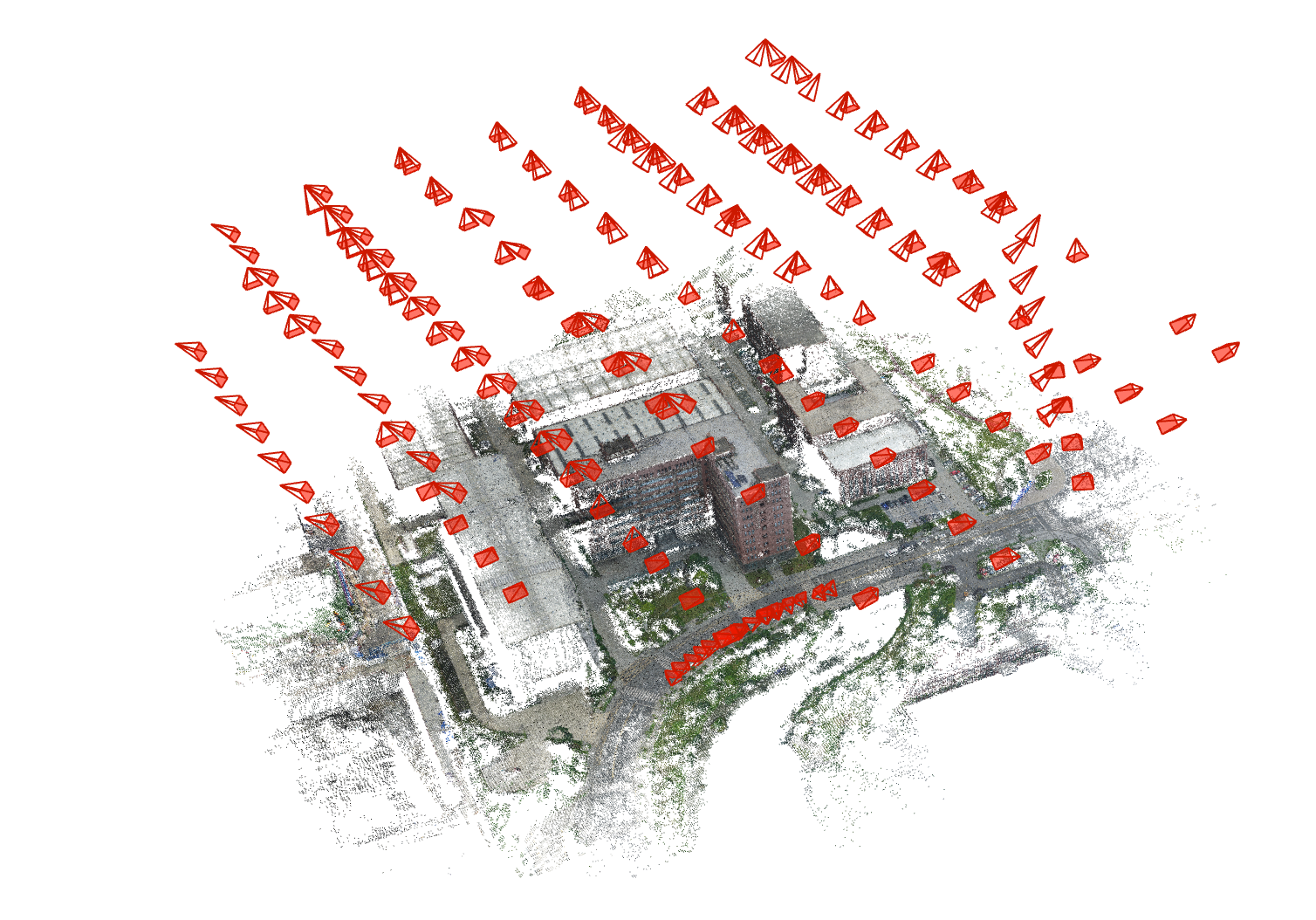} \\
        (b) 
    \end{minipage}
    \begin{minipage}[t]{0.3\linewidth}
    \centering
        \includegraphics[width=1\linewidth]{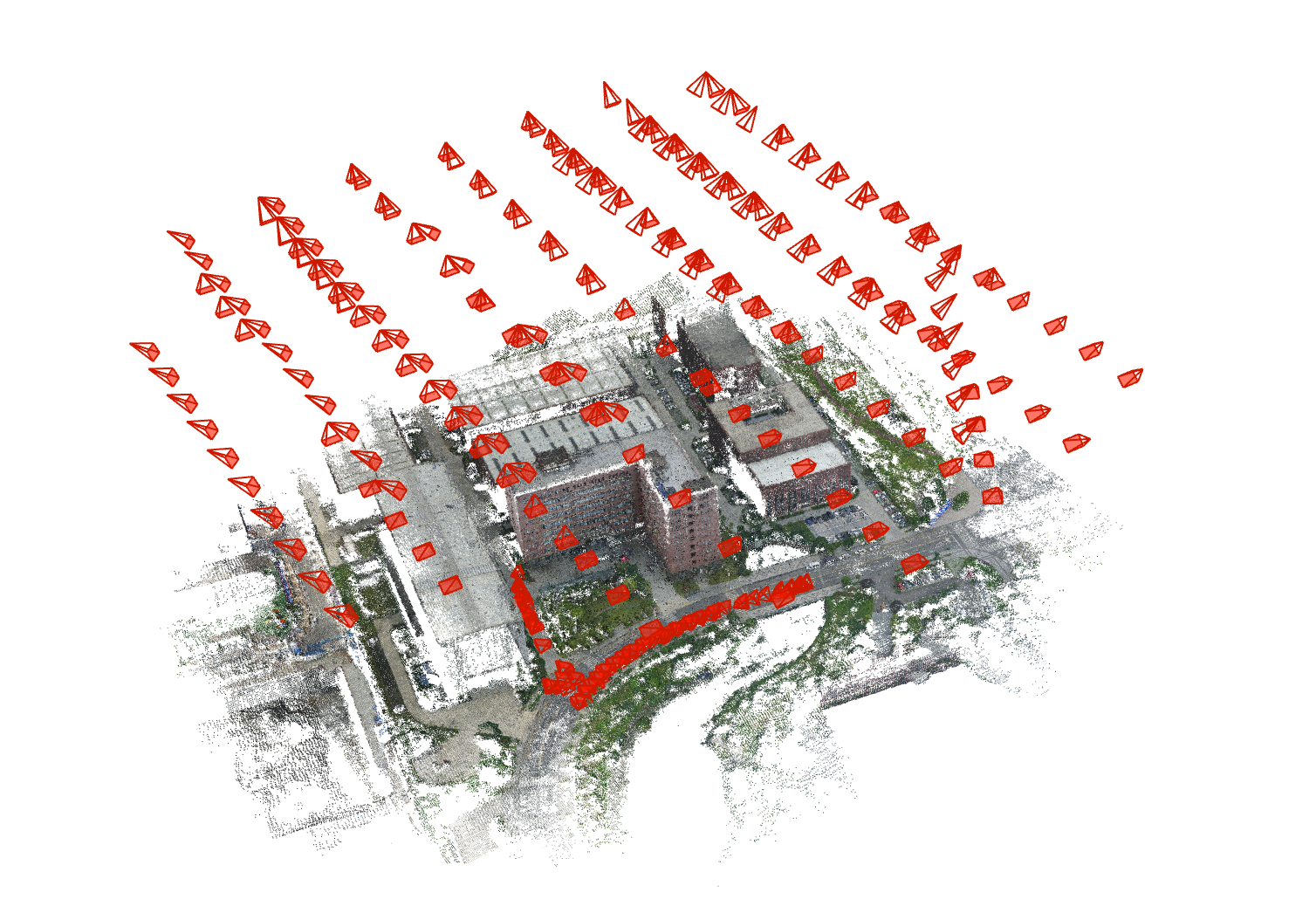} \\
        (c) 
    \end{minipage}
    \begin{minipage}[t]{0.3\linewidth}
    \centering
        \includegraphics[width=1\linewidth]{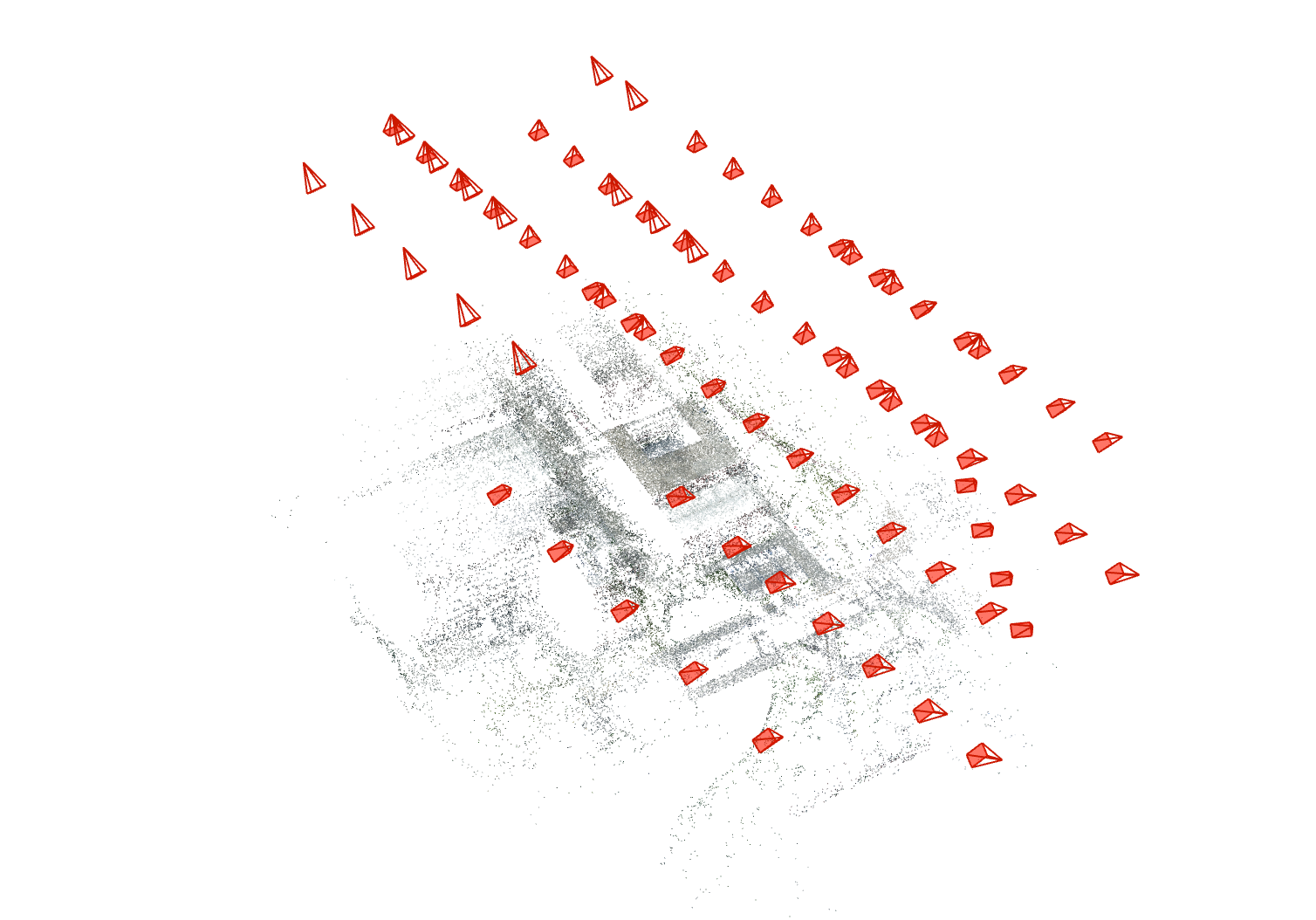} \\
        (d) 
    \end{minipage}
    \begin{minipage}[t]{0.3\linewidth}
    \centering
        \includegraphics[width=1\linewidth]{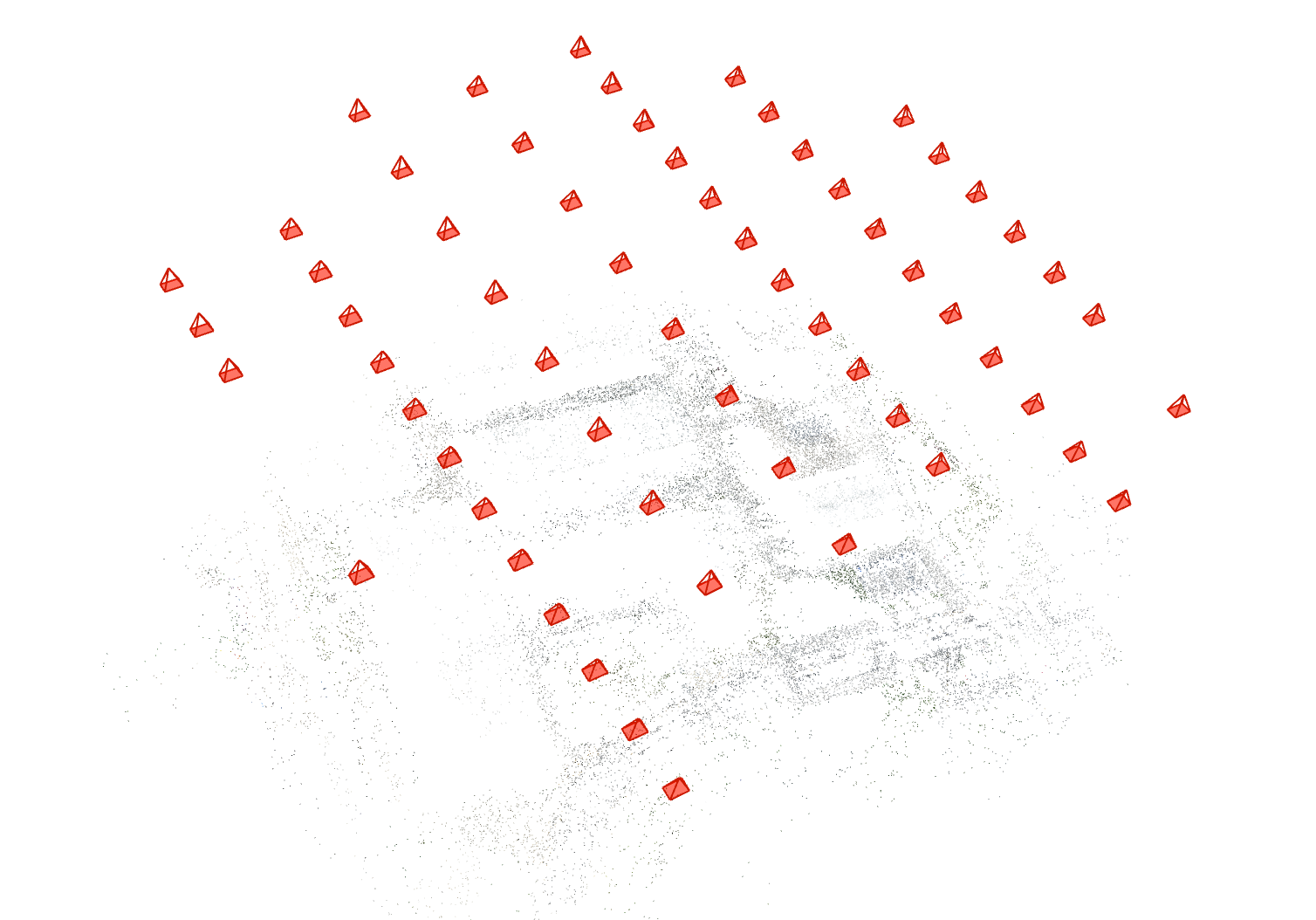} \\
        (e) 
    \end{minipage}
    \begin{minipage}[t]{0.3\linewidth}
    \centering
        \includegraphics[width=1\linewidth]{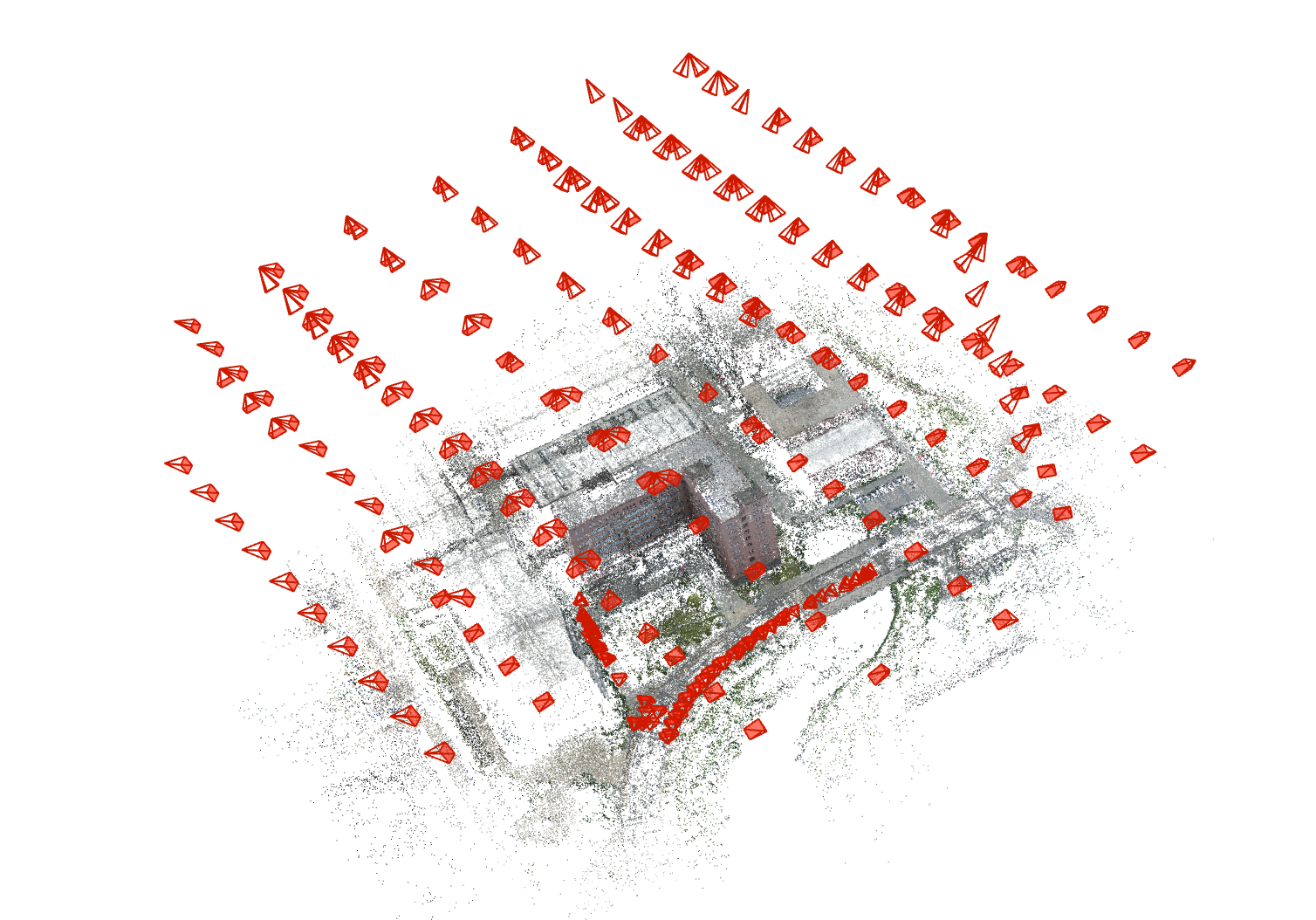} \\
        (f) 
    \end{minipage}
    \caption{ISfM reconstruction for dataset 4: (a) ASpanFormer\_ori; (b) ASpanFormer\_gq(1); (c) ASpanFormer\_con; (d) ELoFTR\_ori; (e) ELoFTR\_gq(1); (f) ELoFTR\_con.}
    \label{fig13}
\end{figure}

To verify the validation of track refinement, this test uses the matching results for ISfM reconstruction. Table \ref{tab7} presents the comparison of ISfM reconstruction for the four datasets. For methods that generate multiple sub-models, only the largest sub-model is presented in the results. The generated 3D models of datasets 3 and 4 under three strategies, i.e., “ori”, “gq(1)”, and “con”, are shown in Figure \ref{fig12} and Figure \ref{fig13}.

\begin{table}[!htp]
	\centering
	\caption{ISfM reconstruction for the four aerial-ground datasets (error unit in pixels).}
	\label{tab7}
    \footnotesize
	\makebox[\linewidth]{
		\begin{tabular}{lrrrrr}
        \toprule
        \textbf{Dataset} & \textbf{Method} & \begin{tabular}[c]{@{}l@{}}\textbf{Mean track} \\ \textbf{length}\end{tabular} & \begin{tabular}[c]{@{}l@{}}\textbf{Registered}   \\ \textbf{images}\end{tabular} & \begin{tabular}[c]{@{}l@{}}\textbf{Number of} \\ 3D \textbf{points}\end{tabular} & \begin{tabular}[c]{@{}l@{}}\textbf{Reprojection} \\ \textbf{error}\end{tabular} \\
        \midrule
        \multirow{10}{*}{1} & SIFT+MNN & 5.256 & 349 & 200,497 & 0.616 \\
         & SuperPoint+MNN & 5.992 & 349 & 205,764 & 0.867 \\
         & ASpanFormer\_ori & 5.140 & 71 & 763,953 & 0.802 \\
         & ASpanFormer\_gq(1) & 5.058 & 349 & 2,469,156 & 1.118 \\
         & ASpanFormer\_gq(4) & 4.842 & 349 & 2,288,564 & 1.114 \\
         & ASpanFormer\_con & 5.704 & 349 & 3,016,177 & 0.882 \\
         & ELoFTR\_ori & 3.063 & 71 & 196,742 & 0.771 \\
         & ELoFTR\_gq(1) & 3.036 & 349 & 990,838 & 1.132 \\
         & ELoFTR\_gq(4) & 3.035 & 349 & 993,320 & 1.132 \\
         & ELoFTR\_con & 3.204 & 349 & 1,640,601 & 0.802 \\
         \midrule
        \multirow{10}{*}{2} & SIFT+MNN & 8.091 & 319 & 253,271 & 0.610 \\
         & SuperPoint+MNN & 10.411 & 319 & 175,831 & 0.801 \\
         & ASpanFormer\_ori & 6.602 & 162 & 1,384,704 & 0.841 \\
         & ASpanFormer\_gq(1) & 5.808 & 319 & 2,788,806 & 1.147 \\
         & ASpanFormer\_gq(4) & 2.904 & 319 & 1,434,850 & 1.405 \\
         & ASpanFormer\_con & 6.666 & 319 & 3,363,817 & 0.876 \\
         & ELoFTR\_ori & 3.118 & 162 & 673,950 & 0.662 \\
         & ELoFTR\_gq(1) & 3.059 & 319 & 1,137,649 & 1.096 \\
         & ELoFTR\_gq(4) & 2.599 & 319 & 1,309,068 & 1.368 \\
         & ELoFTR\_con & 3.338 & 319 & 2,445,306 & 0.705 \\
         \midrule
        \multirow{10}{*}{3} & SIFT+MNN & 6.682 & 201 & 99,866 & 0.880 \\
         & SuperPoint+MNN & 8.224 & 201 & 96,124 & 0.960 \\
         & ASpanFormer\_ori & 4.676 & 37 & 340,139 & 0.832 \\
         & ASpanFormer\_gq(1) & 5.522 & 143 & 1,069,767 & 1.197 \\
         & ASpanFormer\_gq(4) & 3.392 & 78 & 184,596 & 1.499 \\
         & ASpanFormer\_con & 7.103 & 201 & 1,645,298 & 1.004 \\
         & ELoFTR\_ori & 3.097 & 113 & 306,765 & 0.785 \\
         & ELoFTR\_gq(1) & 3.129 & 78 & 381,869 & 1.168 \\
         & ELoFTR\_gq(4) & 2.737 & 78 & 224,180 & 1.396 \\
         & ELoFTR\_con & 3.465 & 201 & 1,489,976 & 0.844 \\
         \midrule
        \multirow{10}{*}{4} & SIFT+MNN & 5.658 & 207 & 104,747 & 0.752 \\
         & SuperPoint+MNN & 6.791 & 288 & 136,152 & 0.915 \\
         & ASpanFormer\_ori & 4.306 & 119 & 702,815 & 0.986 \\
         & ASpanFormer\_gq(1) & 3.834 & 212 & 1,146,886 & 1.121 \\
         & ASpanFormer\_gq(4) & 2.820 & 88 & 170,324 & 1.302 \\
         & ASpanFormer\_con & 4.766 & 295 & 1,807,415 & 1.024 \\
         & ELoFTR\_ori & 2.827 & 80 & 83,299 & 0.806 \\
         & ELoFTR\_gq(1) & 3.036 & 88 & 166,787 & 1.130 \\
         & ELoFTR\_gq(4) & 2.595 & 88 & 145,312 & 1.275 \\
         & ELoFTR\_con & 3.216 & 295 & 707,611 & 0.879 \\
         \bottomrule
        \end{tabular}
	}
\end{table}

1) \textbf{Mean track length}. It is shown that SIFT and SuperPoint maintain longer mean track lengths because of localized keypoint definitions, but generate far fewer overall 3D points. Among detector-free methods, standard models suffer from track fragmentation, leading to truncated track lengths and incomplete image registration. While grid quantization expands image coverage, it degrades track lengths further because discretization errors sever feature trajectories. Conversely, the proposed strategy consistently extends the mean track length across all datasets, such as reaching 5.70 pixels for ASpanFormer\_con, which can bridge broken sub-tracks to achieve stable feature connection for image registration.

2) \textbf{Registered images and 3D points}. The proposed method achieves an overwhelming advantage across all evaluated methods. While traditional handcrafted baselines achieve full registration in datasets 1, 2, and 3, they completely fail in the highly challenging dataset 4 and generate detail-deficient reconstructions. Standard detector-free methods perform the worst overall, e.g., registering 71 images in dataset 1, due to severe feature track fragmentation caused by low cross-view repeatability. Although grid quantization enforces full registration in simple scenes, it excessively filters out valid matches and leads to a decline in image registration for datasets 3 and 4. In contrast, our proposed track connection strategy achieves complete image registration across all datasets while yielding dense 3D point clouds, thereby maximizing reconstruction completeness, as shown in Figure \ref{fig12} and Figure \ref{fig13}.

3) \textbf{Reprojection error}. The results show that the proposed method achieves comparable reconstruction accuracy to that of the baseline methods. The mean reprojection error of SIFT+ MNN is the lowest, remaining between 0.60 and 0.93 pixels in all datasets, demonstrating the high geometric consistency of handcrafted feature extraction. The mean reprojection error of SuperPoint+MNN is slightly higher, ranging from 0.66 to 1.14 pixels. On the contrary, the mean reprojection error of grid quantization is the highest, mainly due to the reduction in feature point accuracy, which would degenerate the reconstruction accuracy.

\subsubsection{Coarse model refinement}
\label{sec4.4.3}
\begin{figure}[!t]
    \centering
    \begin{minipage}[t]{0.3\linewidth}
    \centering
        \includegraphics[width=1\linewidth]{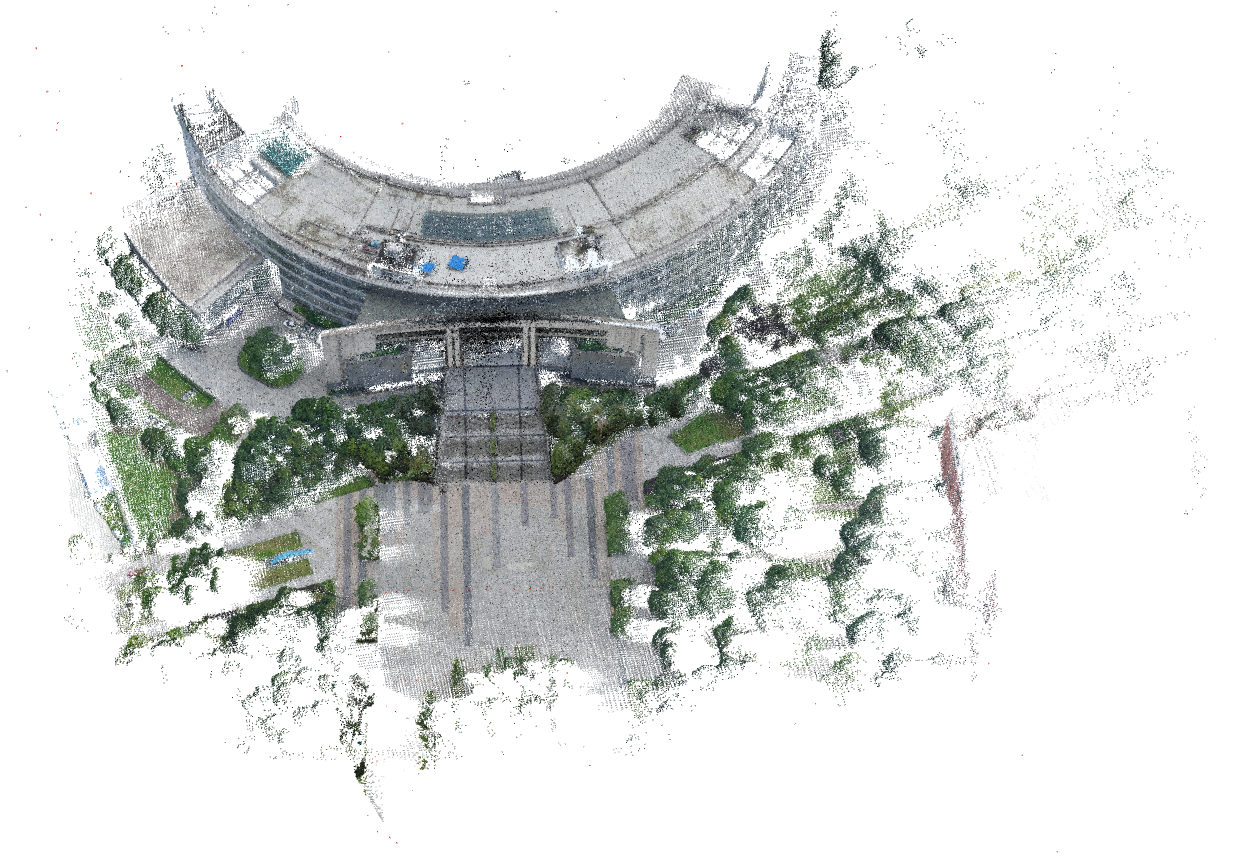} \\
        (a)
    \end{minipage}
    \begin{minipage}[t]{0.3\linewidth}
    \centering
        \includegraphics[width=1\linewidth]{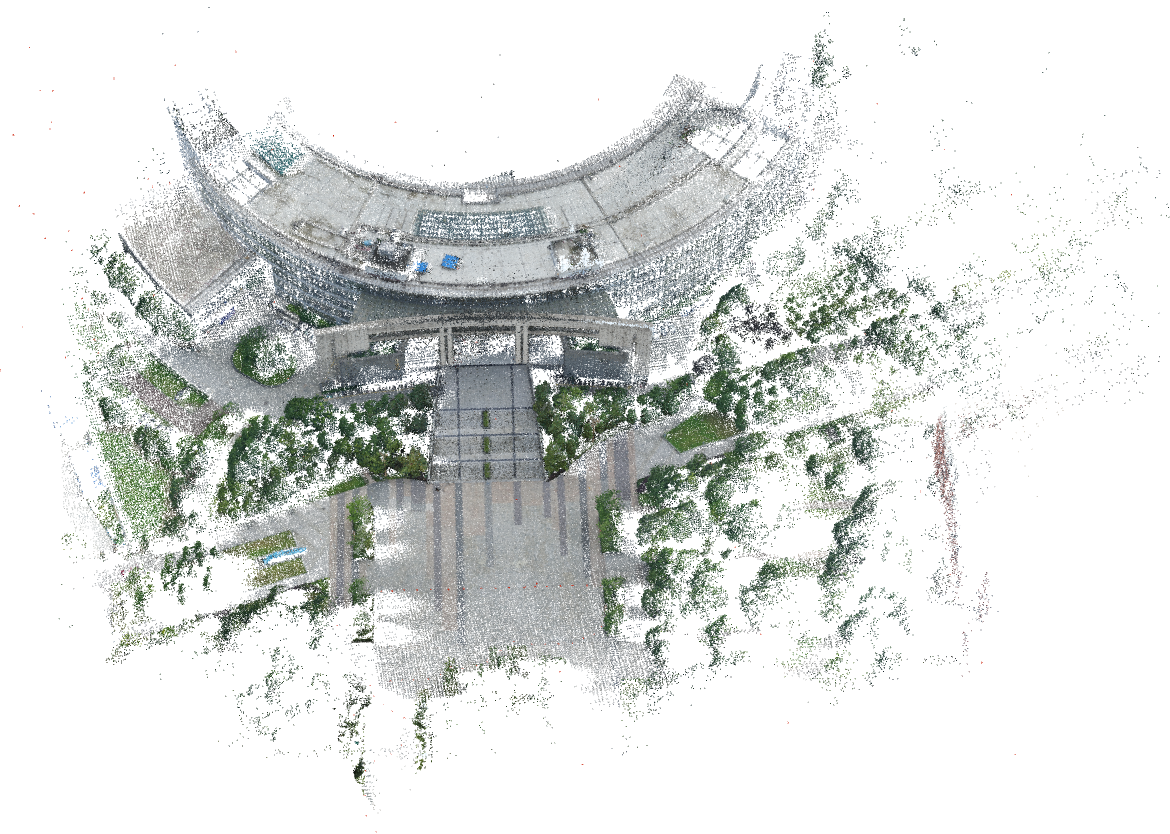} \\
        (b) 
    \end{minipage}
    \begin{minipage}[t]{0.3\linewidth}
    \centering
        \includegraphics[width=1\linewidth]{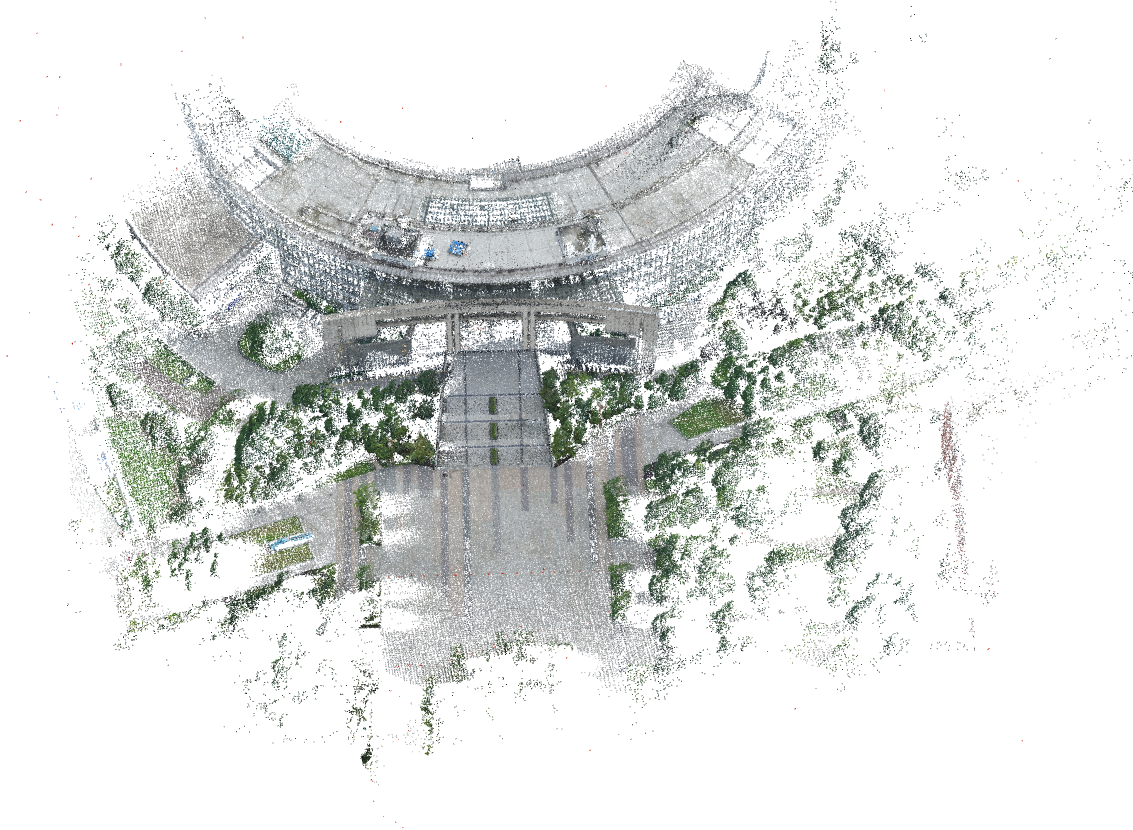} \\
        (c) 
    \end{minipage}
    \begin{minipage}[t]{0.3\linewidth}
    \centering
        \includegraphics[width=1\linewidth]{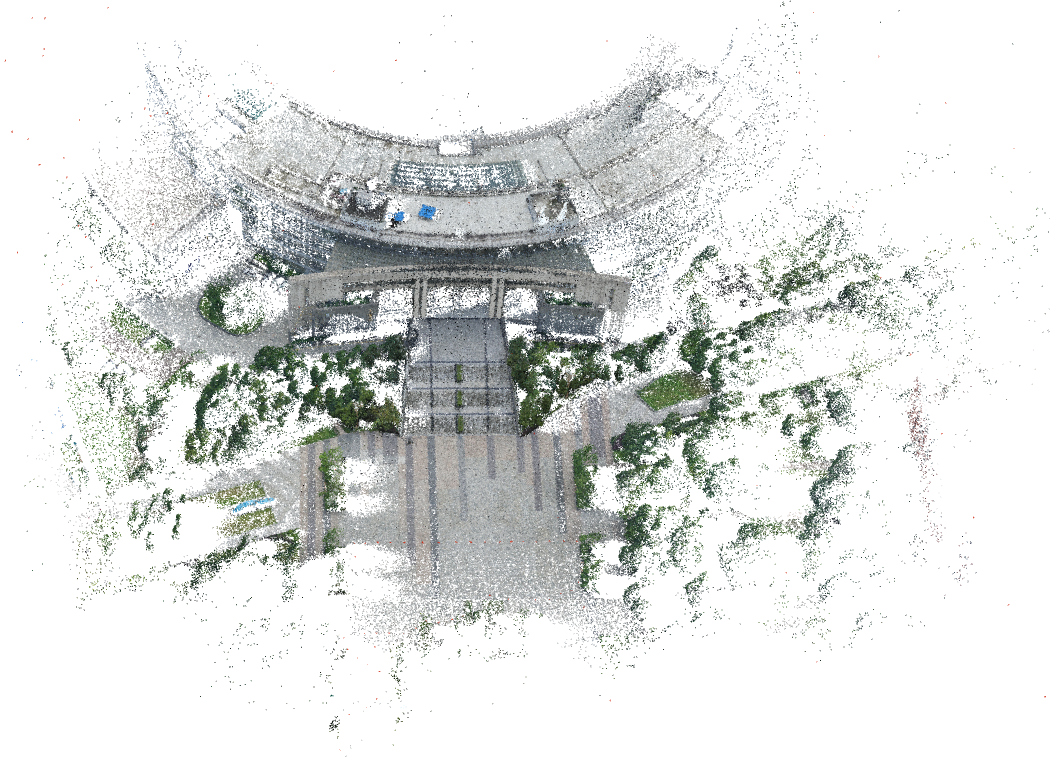} \\
        (d) 
    \end{minipage}
    \begin{minipage}[t]{0.3\linewidth}
    \centering
        \includegraphics[width=1\linewidth]{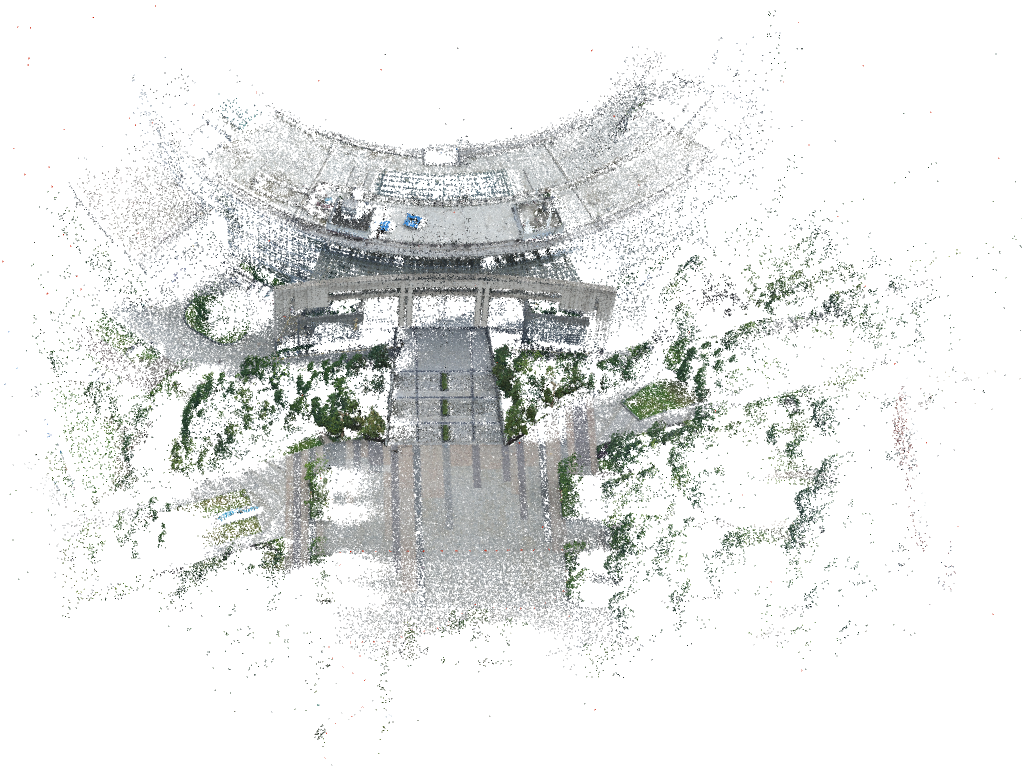} \\
        (e) 
    \end{minipage}
    \begin{minipage}[t]{0.3\linewidth}
    \centering
        \includegraphics[width=1\linewidth]{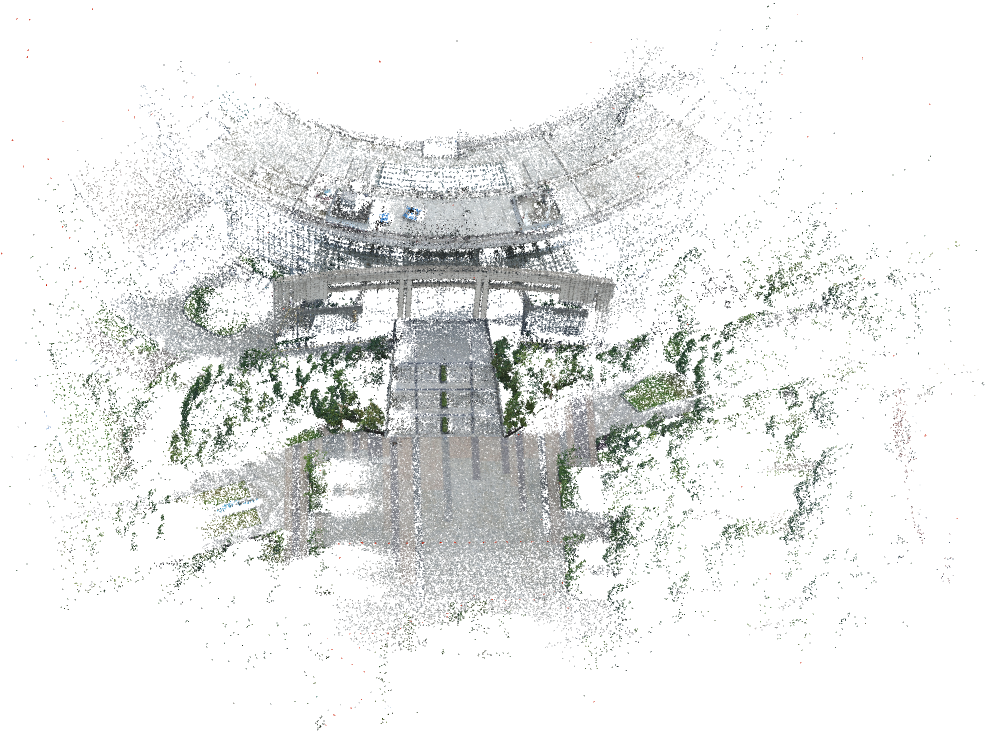} \\
        (f) 
    \end{minipage}
    \caption{The comparison of refined 3D models for dataset 3: (a) ASpanFormer\_con; (b) ASpanFormer\_con(1.5); (c) ASpanFormer\_con(1); (d) ELoFTR\_con; (e) ELoFTR\_con(1.5); (f) ELoFTR\_con(1).}
    \label{fig14}
\end{figure}

\begin{figure}[!t]
    \centering
    \begin{minipage}[t]{0.3\linewidth}
    \centering
        \includegraphics[width=1\linewidth]{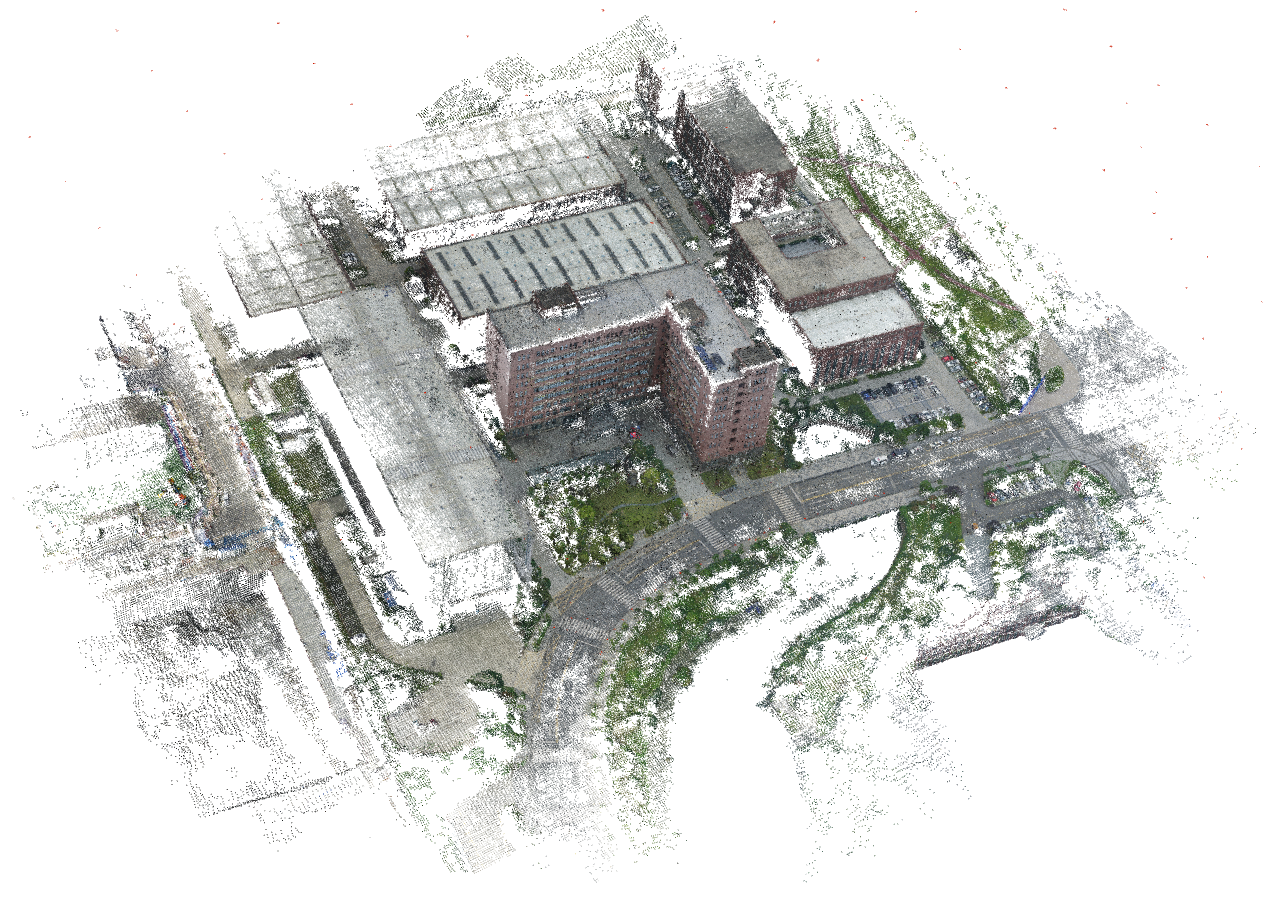} \\
        (a)
    \end{minipage}
    \begin{minipage}[t]{0.3\linewidth}
    \centering
        \includegraphics[width=1\linewidth]{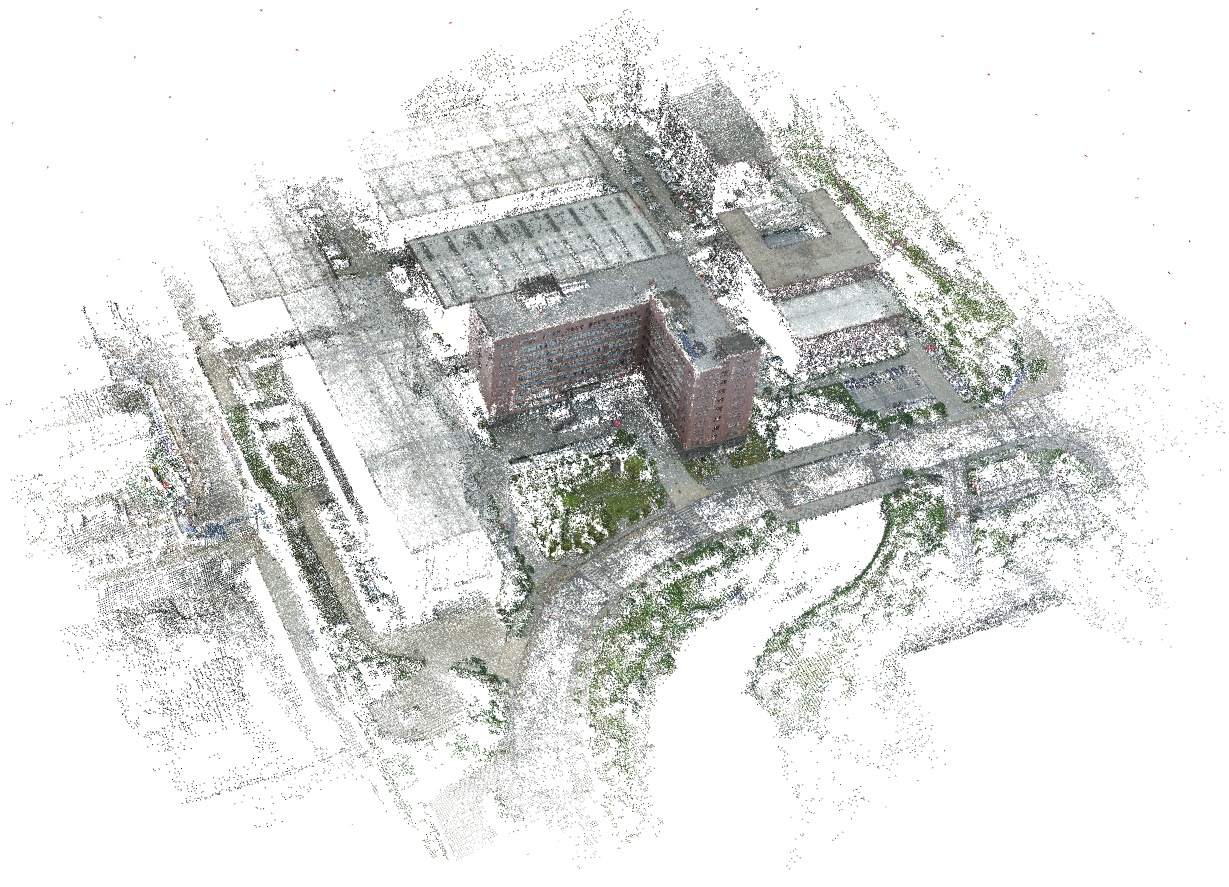} \\
        (b) 
    \end{minipage}
    \begin{minipage}[t]{0.3\linewidth}
    \centering
        \includegraphics[width=1\linewidth]{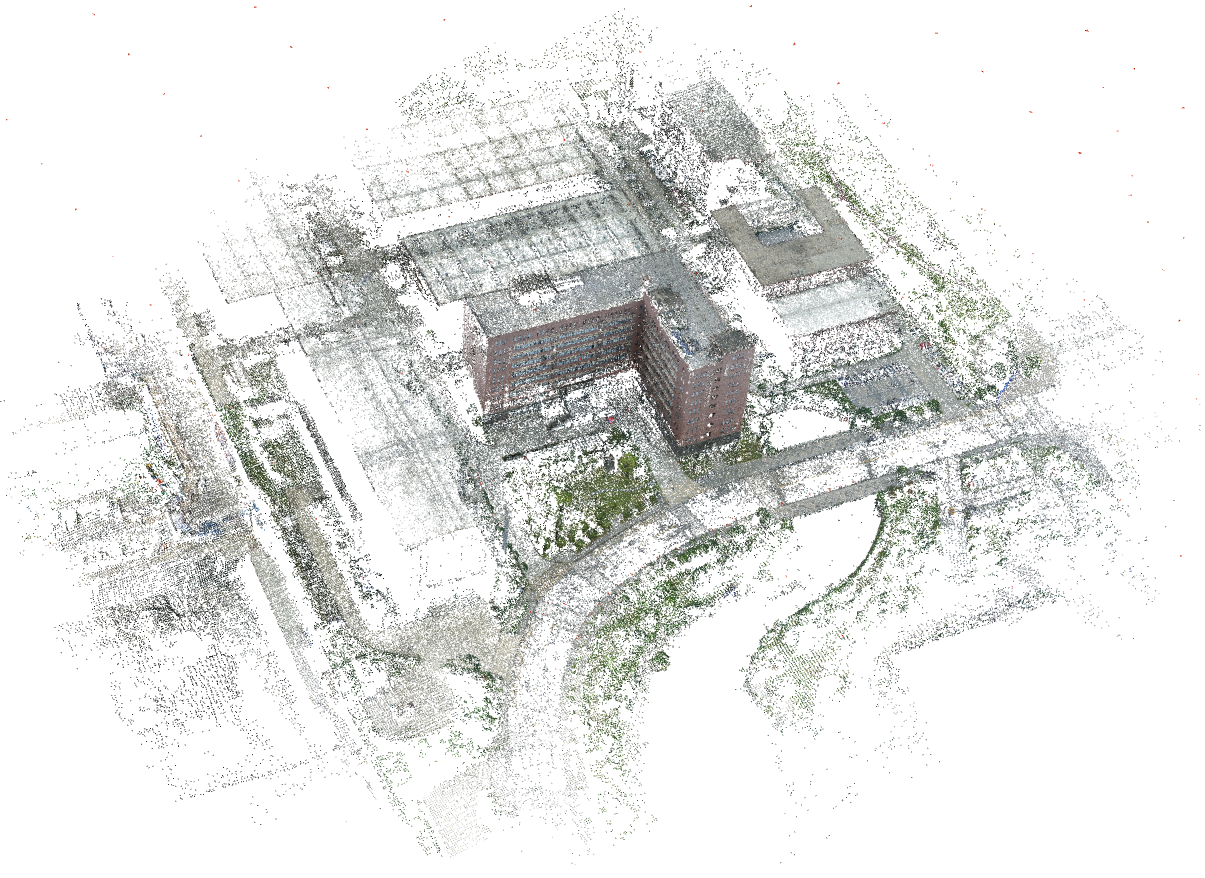} \\
        (c) 
    \end{minipage}
    \begin{minipage}[t]{0.3\linewidth}
    \centering
        \includegraphics[width=1\linewidth]{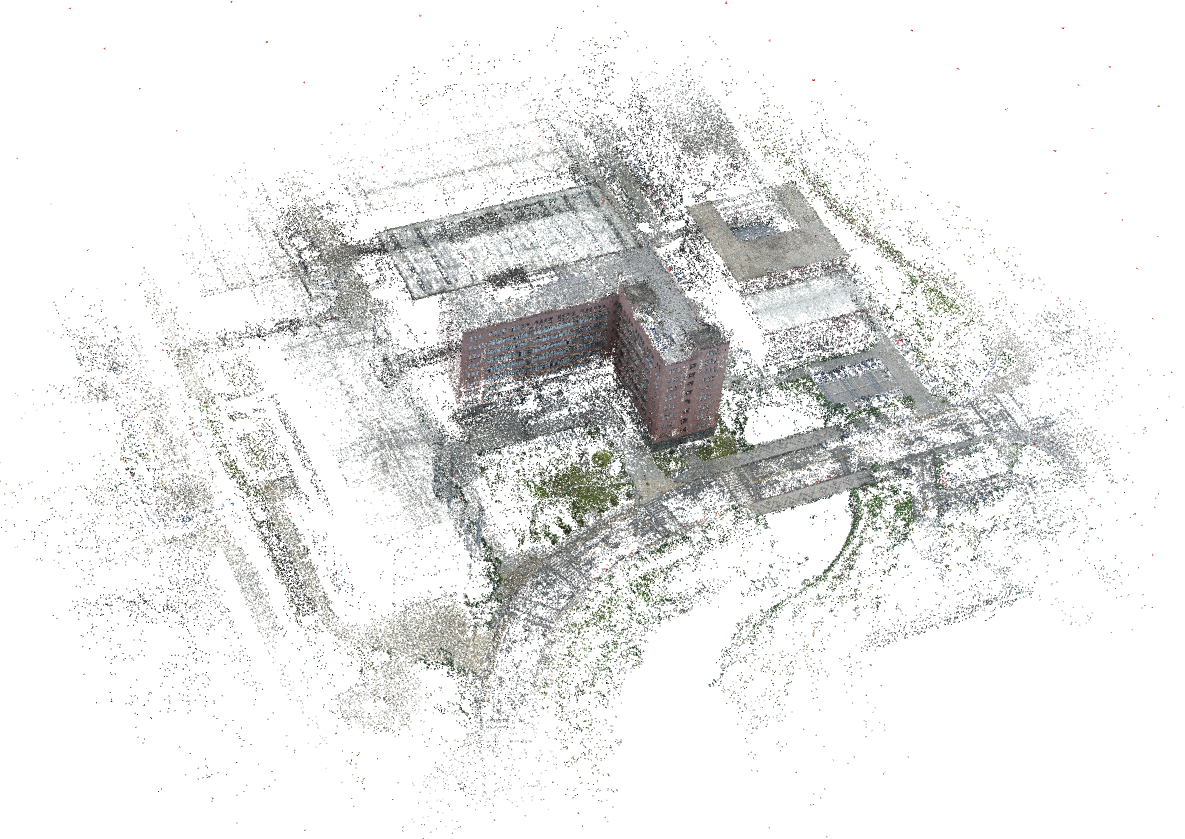} \\
        (d) 
    \end{minipage}
    \begin{minipage}[t]{0.3\linewidth}
    \centering
        \includegraphics[width=1\linewidth]{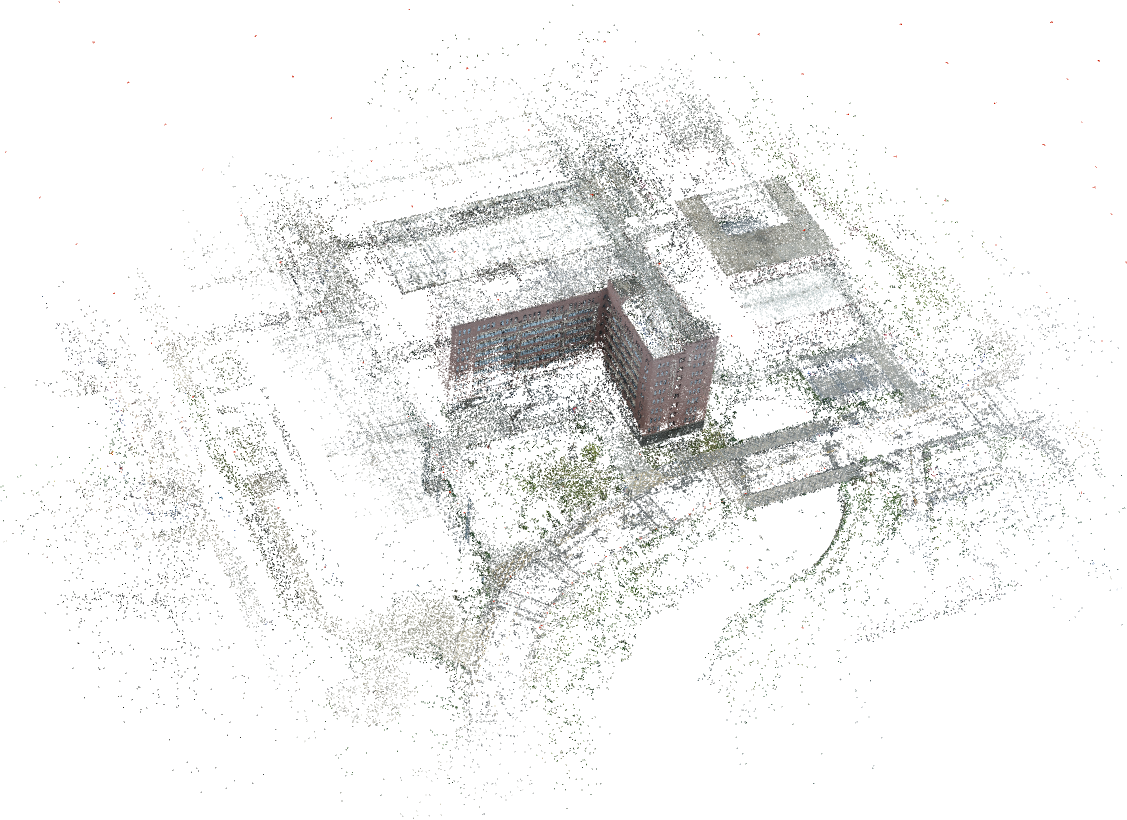} \\
        (e) 
    \end{minipage}
    \begin{minipage}[t]{0.3\linewidth}
    \centering
        \includegraphics[width=1\linewidth]{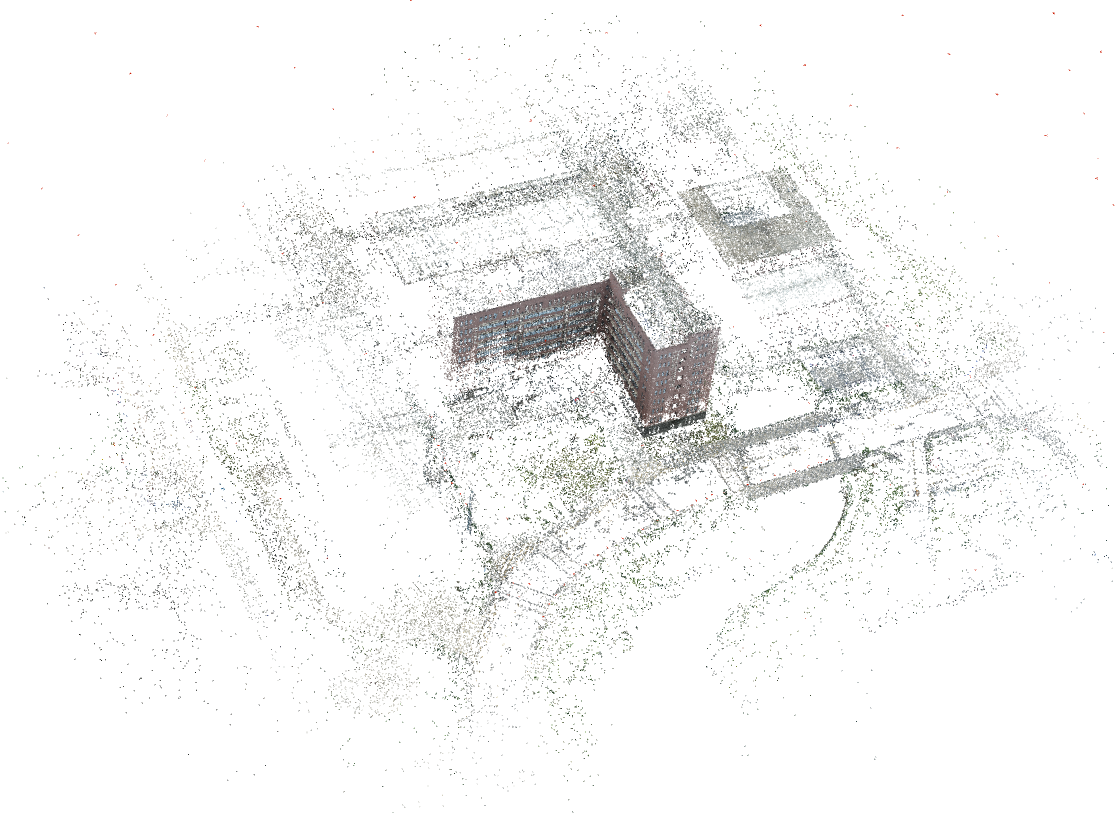} \\
        (f) 
    \end{minipage}
    \caption{The comparison of refined 3D models for dataset 4: (a) ASpanFormer\_con; (b) ASpanFormer\_con(1.5); (c) ASpanFormer\_con(1); (d) ELoFTR\_con; (e) ELoFTR\_con(1.5); (f) ELoFTR\_con(1).}
    \label{fig15}
\end{figure}

Coarse models obtained by the proposed method are used for iterative refinement. The quantitative results are shown in Table \ref{tab8}. Additionally, the visual results of datasets 3 and 4 are presented in Figure \ref{fig14} and Figure \ref{fig15}, respectively. In this test, two detector-free methods, i.e., ASpanFormer and ELoFTR, that are coupled with the proposed modules, are selected for tests. In addition, according to the used reconstruction error thresholds, the two iterative results are named by adding “(1.5)” and “(1)” to the original method, respectively.

\begin{table}[!ht]
	\centering
	\caption{Refined results of coarse models for the four datasets (error unit in pixels).}
	\label{tab8}
    \small
	\makebox[0.5\linewidth]{
		\begin{tabular}{lrrr}
        \toprule
        \textbf{Dataset} & \textbf{Method} & \begin{tabular}[c]{@{}l@{}}\textbf{Number of} \\ \textbf{3D points}\end{tabular} & \begin{tabular}[c]{@{}l@{}}\textbf{Reprojection} \\ \textbf{error}\end{tabular} \\
        \midrule
        \multirow{6}{*}{1} & ASpanFormer\_con & 3,016,177 & 0.882 \\
         & ASpanFormer\_con(1.5) & 2,247,052 & 0.548 \\
         & ASpanFormer\_con(1) & 1,996,602 & 0.389 \\
         & ELoFTR\_con & 1,640,601 & 0.802 \\
         & ELoFTR\_con(1.5) & 1,164,696 & 0.494 \\
         & ELoFTR\_con(1) & 1,021,031 & 0.351 \\
         \midrule
        \multirow{6}{*}{2} & ASpanFormer\_con & 3,363,817 & 0.876 \\
         & ASpanFormer\_con(1.5) & 2,686,209 & 0.486 \\
         & ASpanFormer\_con(1) & 2,531,879 & 0.356 \\
         & ELoFTR\_con & 2,445,306 & 0.705 \\
         & ELoFTR\_con(1.5) & 1,826,858 & 0.442 \\
         & ELoFTR\_con(1) & 1,706,773 & 0.311 \\
         \midrule
        \multirow{6}{*}{3} & ASpanFormer\_con & 1,645,298 & 1.004 \\
         & ASpanFormer\_con(1.5) & 1,166,769 & 0.572 \\
         & ASpanFormer\_con(1) & 1,025,444 & 0.394 \\
         & ELoFTR\_con & 1,489,976 & 0.844 \\
         & ELoFTR\_con(1.5) & 1,010,188 & 0.528 \\
         & ELoFTR\_con(1) & 891,721 & 0.357 \\
         \midrule
        \multirow{6}{*}{4} & ASpanFormer\_con & 1,807,415 & 1.024 \\
         & ASpanFormer\_con(1.5) & 1,090,557 & 0.545 \\
         & ASpanFormer\_con(1) & 888,323 & 0.371 \\
         & ELoFTR\_con & 707,611 & 0.879 \\
         & ELoFTR\_con(1.5) & 405,824 & 0.524 \\
         & ELoFTR\_con(1) & 328,573 & 0.352 \\
         \bottomrule
        \end{tabular}
	}
\end{table}

1) \textbf{3D points}. Observed from Figure \ref{fig14} and Figure \ref{fig15}, the results show that during the refinement process, the removed redundant and erroneous points are those with geometric inconsistencies, rather than core scene structures. The reduction in 3D points is positively correlated with the tightening of the error threshold. In the first round, approximately 25\% to 40\% of the 3D points are typically removed. In the second round, approximately 10\% to 15\% of the 3D points are removed. This hierarchical attenuation pattern avoids losing effective structures due to a strict single threshold, and the 3D points are still several times that of SIFT. This indicates that the refinement framework achieves high accuracy while successfully retaining rich scene details and can effectively adapt to the connection strategy.

2) \textbf{Reprojection error}. The results show that the refinement framework can achieve a significant stepwise reduction in error. The refinement effect shows high consistency and stability across all datasets. From the initial “con” to “con(1.5)” and then to “con(1)”, the error shows a two-level downward trend. For example, in dataset 1, the error of ELoFTR decreased from 0.802 to 0.494 pixels and finally reached 0.351 pixels. This fully proves that the refinement framework can effectively eliminate the geometric gross errors introduced by perspective distortion and scale differences in initial reconstruction. Although the initial errors of the two basic methods are different. After two rounds of refinement, their final errors tend to be consistent. In a word, the refinement framework can precisely remove these interfering points, ensuring the stability of subsequent reconstruction.

\subsubsection{Fine model accuracy}
\label{sec4.4.4}
\begin{figure}[!t]
    \centering
    \begin{minipage}[t]{0.8\linewidth}
    \centering
        \includegraphics[width=1\linewidth]{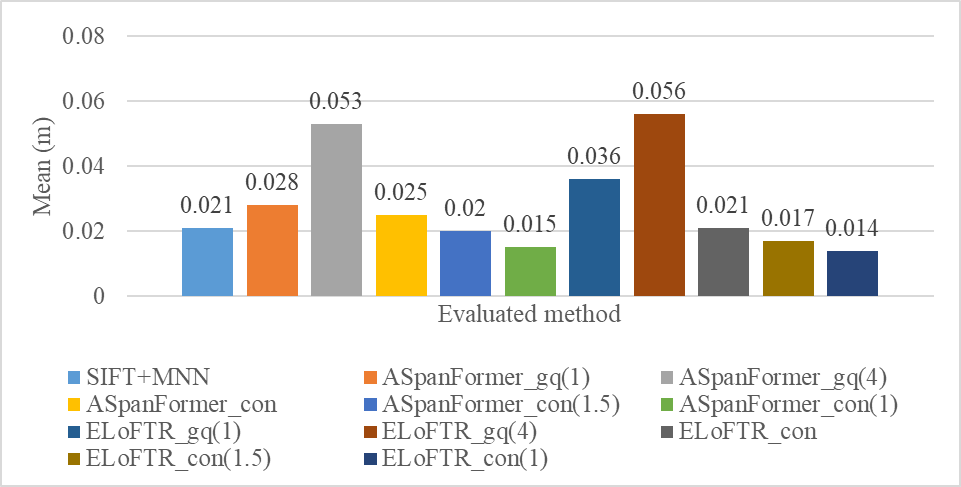} \\
        (a)
    \end{minipage}
    \begin{minipage}[t]{0.8\linewidth}
    \centering
        \includegraphics[width=1\linewidth]{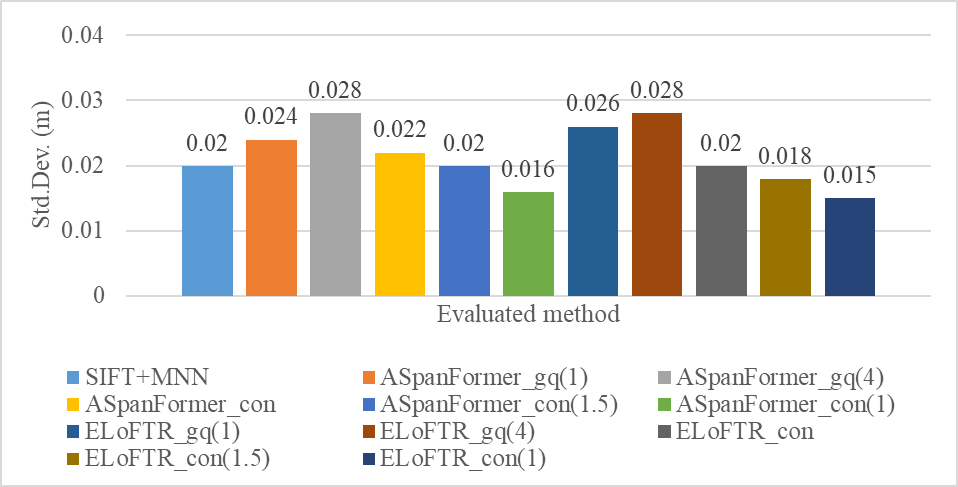} \\
        (b) 
    \end{minipage}
    \caption{The statistical results of fine model accuracy: (a) Mean distance; (b) Std.Dev. of distance.}
    \label{fig16}
\end{figure}

In this evaluation, dataset 2 is selected as it contains ground-truth LiDAR point clouds for geometric precision evaluation. The quantitative results are presented in Figure \ref{fig16}.

1) \textbf{Mean distance}. The refinement framework obviously reduces the mean distance and outperforms baseline methods. Grid quantization yields the lowest accuracy, which degrades further as the quantization scale expands. For example, ASpanFormer\_gq(4) is approximately 3.5 times higher than ASpanFormer\_con(1), indicating that excessive quantization disrupts geometric consistency and introduces global 3D point offsets. Conversely, our connection strategy successfully suppresses these systematic errors while retaining valid matches, with ELoFTR\_con achieving a 41.67\% reduction over ELoFTR\_gq(1). Notably, ELoFTR\_con(1) and ASpanFormer\_con(1) achieve optimal mean distances of 0.014 m and 0.015 m, surpassing the SIFT+MNN baseline by 33.33\% and 28.57\%, respectively.

2) \textbf{Std.Dev}. of distance Refinement also narrows error dispersion and filters out discrete structural outliers, concentrating the error distribution. The Std.Dev. of ASpanFormer\_con(1) (0.016 m) is dropped to only 57.14\% of ASpanFormer\_gq(4) (0.028 m). Thus, through multi-stage geometric refinement, the proposed detector-free ISfM framework overcomes the limitation of hand-crafted descriptors to deliver highly reliable 3D reconstruction.

\section{Conclusions}
\label{sec5}

This study presents a rotation-robust detector-free matching network coupled with a cross-view ISfM reconstruction framework to address the crucial bottlenecks of low rotation robustness and poor compatibility in aerial-ground ISfM. By combining a bidirectional refinement module for sub-pixel localization, quadtree attention for dependency preservation, and an omnidirectional state space block for rotation-invariant feature extraction, the proposed matching network prevents matching failures in rotated scenes and improves the 5° pose error AUC by 93.9\% when compared to LoFTR. The proposed ISfM framework adds a fine feature track connection method to the front-end network, which greatly improves cross-view feature repeatability without sacrificing localization accuracy. This solution can generate almost ten times as many 3D points as SIFT and improves final reconstruction accuracy, ranging from 27.6\% to 32.7\% while achieving complete image registration across all datasets by avoiding the precision loss inherent to grid quantization.

This work provides a highly effective solution for 3D reconstruction of aerial-ground images, and future research can be conducted from three aspects. First, in order to efficiently process ultra-high-resolution images, a lightweight improvement of the feature extraction backbone is required to reduce the GPU memory footprint. Second, the decoupling barriers between the network and the ISfM pipeline would be eliminated by creating an end-to-end architecture that simultaneously optimizes matching and reconstruction to increase overall efficiency. Furthermore, adding multi-modal priors, such as depth, semantic, and geometric information, will improve matching robustness in situations with intense lighting, strong occlusions, and textureless scenes. Finally, building a large-scale, cross-view aerial-ground matching benchmark will enable the investigation of few-shot and self-supervised learning techniques to significantly enhance open-world generalization.

\section*{Acknowledgments}
This research was funded by the National Natural Science Foundation of China (No. 42371442); the Shenzhen Science and Technology Program (No. JCYJ20250604181614019); the Shenzhen Key Laboratory Program (No. SYSPG20241211173845013); and the Shenzhen-Nanning Spatial Intelligence Joint Laboratory project(No. 25220019).

\appendix

\bibliographystyle{elsarticle-harv}
\bibliography{mybibfile}





\end{document}